\documentclass[10pt]{article}

\usepackage[preprint]{tmlr}

\usepackage{amsmath,amssymb,amsfonts,amsthm,mathtools,bm}
\usepackage{booktabs}
\usepackage{enumitem}
\usepackage{graphicx}
\usepackage{microtype}
\usepackage{tikz}
\usetikzlibrary{arrows.meta,positioning,calc,fit}
\usepackage{hyperref}
\usepackage{url}
\usepackage[nameinlink,capitalise]{cleveref}
\hypersetup{
    colorlinks=true,
    linkcolor=blue,
    citecolor=blue,
    urlcolor=blue
}

\setlist{nosep,leftmargin=1.5em}

\providecommand{\E}{\mathbb{E}}
\providecommand{\KL}{\mathrm{KL}}
\providecommand{\MI}{\mathrm{I}}
\newcommand{\sg}{\operatorname{sg}}
\newcommand{\cS}{\mathcal{S}}
\newcommand{\cX}{\mathcal{X}}

\newcommand{\DeltaK}{\Delta^{K-1}}
\newcommand{\ours}{\textsc{MCJEPA}}

\newtheorem{proposition}{Proposition}
\newtheorem{theorem}{Theorem}

\theoremstyle{definition}

\theoremstyle{remark}

\title{
Your Probabilistic JEPA Is Secretly a Hidden Markov Model\\[0.4em]
{\large \textit{A State-Space Interpretation of Joint-Embedding Predictive Learning}}
}

\author{
\name Yongchao Huang
\email{yongchao.huang@abdn.ac.uk} \\
\addr Department of Computing Science \\
University of Aberdeen
}

\def\month{08}
\def\year{2026}

\def\openreview{}

\begin{document}

\maketitle

\begin{abstract}
A hidden Markov model (HMM) combines three roles: inference of a hidden-state belief from observations, propagation through a Markov transition, and emission back to observation space. We show that full, time-indexed Predictive Information Bottleneck VJEPA (PIB-VJEPA) exposes the same computational structure: a stochastic context encoder plays the role of an amortized filtering distribution, a probabilistic predictor defines latent-state dynamics, and a decoder, inverse target encoder, or induced implicit conditional supplies the emission direction. We distinguish 4 progressively stronger levels of correspondence and give sufficient conditions for exact sequence-level HMM equivalence. To make the connection concrete, we introduce Markov-Chain JEPA (MCJEPA), which replaces the latent predictor by a learned transition matrix; in the finite time-homogeneous case, matrix powers guarantee exact multi-horizon Chapman--Kolmogorov consistency. Conditioned discrete-state transitions, continuous-state Markov kernels, and continuous-time dynamics extend this construction, while deterministic temporal JEPA appears as a degenerate Dirac-kernel special case. We further interpret predictive information-bottleneck learning as seeking a compact predictive state: compression promotes \textit{minimality}, while residual predictability tests \textit{sufficiency}. Controlled experiments support transition composition, the filtering interpretation, predictive Markovization in a known synthetic process, and the distinction between JEPA latent prediction and HMM-style sequence learning. Together, these results give temporal JEPA a principled state-space interpretation.
\end{abstract}

\tableofcontents
\newpage

\section{Introduction}

Joint-Embedding Predictive Architectures (JEPAs) learn by predicting a target representation from an observed context rather than reconstructing the target observation itself \citep{lecun2022path,assran2023ijepa,bardes2024vjepa}. Variational JEPA (VJEPA) makes this prediction probabilistic, replacing a point predictor with a conditional distribution over future latent states \citep{huang2026vjepa}. A full Predictive Information Bottleneck (PIB) extension additionally makes the current representation stochastic and explicitly controls how much information it retains about the observed history \citep{huang2026ibvjepa}.

This progression creates a natural question: \emph{what familiar probabilistic model is hidden inside a fully stochastic temporal JEPA?} The basic state-space analogy is immediate. An HMM follows
\begin{equation}
X_{\leq t}
\longrightarrow
p(S_t\mid X_{\leq t})
\longrightarrow
p(S_{t+1}\mid S_t)
\longrightarrow
p(X_{t+1}\mid S_{t+1}),
\label{eq:hmm-pipeline-intro}
\end{equation}
whereas full PIB-VJEPA, when equipped with an explicit observation model, has the corresponding pipeline
\begin{equation}
X_{\leq t}
\longrightarrow
q_\theta(Z_t\mid X_{\leq t})
\longrightarrow
p_\phi(Z_{t+1}\mid Z_t,\xi_t)
\longrightarrow
p_\psi(X_{t+1}\mid Z_{t+1}).
\label{eq:jepa-pipeline-intro}
\end{equation}
Here, $X_t$ denotes observation-level data such as pixels, video frames, or time-series measurements; $Z_t$ denotes a stochastic predictive representation; and $\xi_t$ contains side information such as an action, elapsed time, target position, or exogenous covariates \citep{huang2026vjepa}.

The observation-space path in \cref{eq:jepa-pipeline-intro} is optional for the core JEPA objective. It may be implemented by an explicit probabilistic decoder $p_\psi(X_{t+1}\mid Z_{t+1})$. Alternatively, if the target encoder $f_{\bar\theta}$ is invertible on the modeled data domain, its inverse provides a deterministic state-to-observation map,
\begin{equation}
\widehat X_{t+1}
=
f_{\bar\theta}^{-1}(\widehat Z_{t+1}).
\label{eq:inverse-target-emission}
\end{equation}
A reliable approximate inverse may similarly support reconstruction or forecasting, although it does not by itself define the normalized emission likelihood required for exact HMM-style likelihood training. Under these interpretations, observation-level data correspond to HMM observations, the stochastic context encoder plays the hidden-state inference role, the predictor propagates the latent state, and the decoder or inverse target encoder realizes the state-to-observation direction.

\begin{figure}[t]
\centering
\resizebox{0.98\linewidth}{!}{%
\begin{tikzpicture}[
x=1cm,
y=1cm,
latent/.style={
    draw,
    circle,
    minimum size=10mm,
    inner sep=0pt,
    font=\small
},
obs/.style={
    draw,
    circle,
    minimum size=10mm,
    inner sep=0pt,
    font=\small
},
pipebox/.style={
    draw,
    rounded corners,
    minimum width=2.8cm,
    minimum height=0.82cm,
    align=center,
    font=\small
},
arr/.style={
    -{Latex[length=2mm]},
    thick
},
darr/.style={
    -{Latex[length=2mm]},
    thick,
    dashed
},
pipearr/.style={
    -{Latex[length=1.8mm]},
    thick
},
lbl/.style={
    font=\small,
    align=left
},
smalllbl/.style={
    font=\scriptsize,
    align=center
},
rowtitle/.style={
    font=\bfseries
}
]


\node[rowtitle] at (0,5.25) {Hidden Markov model};

\node[obs] (x1) at (-4.0,4.1) {$X_1$};
\node[obs] (x2) at ( 0.0,4.1) {$X_2$};
\node[obs] (x3) at ( 4.0,4.1) {$X_3$};

\node[latent] (s1) at (-4.0,2.3) {$S_1$};
\node[latent] (s2) at ( 0.0,2.3) {$S_2$};
\node[latent] (s3) at ( 4.0,2.3) {$S_3$};

\draw[arr] (-6.4,2.3) -- (-4.65,2.3);
\draw[arr]
    (s1)
    --
    node[above,smalllbl] {$p(S_2\mid S_1)$}
    (s2);
\draw[arr]
    (s2)
    --
    node[above,smalllbl] {$p(S_3\mid S_2)$}
    (s3);
\draw[arr] (4.65,2.3) -- (6.4,2.3);

\draw[arr]
    (s1)
    --
    node[left,smalllbl] {$p(X_1\mid S_1)$}
    (x1);
\draw[arr]
    (s2)
    --
    node[left,smalllbl] {$p(X_2\mid S_2)$}
    (x2);
\draw[arr]
    (s3)
    --
    node[left,smalllbl] {$p(X_3\mid S_3)$}
    (x3);

\node[lbl,anchor=east] at (-6.8,4.2) {observations};
\node[lbl,anchor=east,align=left] at (-6.5,2.2) {hidden Markov\\chain};

\node[pipebox] (hobs) at (-5.1,0.65)
{$X_{\leq t}$};

\node[pipebox] (hinf) at (-1.7,0.65)
{$p(S_t\mid X_{\leq t})$};

\node[pipebox] (htrans) at (1.7,0.65)
{$p(S_{t+1}\mid S_t)$};

\node[pipebox] (hemit) at (5.1,0.65)
{$p(X_{t+1}\mid S_{t+1})$};

\draw[pipearr] (hobs) -- (hinf);
\draw[pipearr] (hinf) -- (htrans);
\draw[pipearr] (htrans) -- (hemit);

\node[smalllbl,above=1mm of hobs] {observation history};
\node[smalllbl,above=1mm of hinf] {state inference};
\node[smalllbl,above=1mm of htrans] {state transition};
\node[smalllbl,above=1mm of hemit] {emission};


\node[rowtitle] at (0,-1.15) {Full PIB-VJEPA};

\node[obs] (y1) at (-4.0,-2.3) {$X_1$};
\node[obs] (y2) at ( 0.0,-2.3) {$X_2$};
\node[obs] (y3) at ( 4.0,-2.3) {$X_3$};

\node[latent] (z1) at (-4.0,-4.1) {$Z_1$};
\node[latent] (z2) at ( 0.0,-4.1) {$Z_2$};
\node[latent] (z3) at ( 4.0,-4.1) {$Z_3$};

\draw[arr] (-6.4,-4.1) -- (-4.65,-4.1);
\draw[arr]
    (z1)
    --
    node[above,smalllbl]
    {$p_\phi(Z_2\mid Z_1,\xi_1)$}
    (z2);
\draw[arr]
    (z2)
    --
    node[above,smalllbl]
    {$p_\phi(Z_3\mid Z_2,\xi_2)$}
    (z3);
\draw[arr] (4.65,-4.1) -- (6.4,-4.1);

\draw[arr]
    (y1)
    --
    node[left,smalllbl] {$q_\theta$}
    (z1);
\draw[arr]
    (y2)
    --
    node[left,smalllbl] {$q_{\bar\theta}$}
    (z2);
\draw[arr]
    (y3)
    --
    node[left,smalllbl] {$q_{\bar\theta}$}
    (z3);

\draw[darr]
    ([xshift=2.7mm]z1.north)
    --
    node[right,smalllbl]
    {$p_\psi$ or $f_{\bar\theta}^{-1}$}
    ([xshift=2.7mm]y1.south);

\draw[darr]
    ([xshift=2.7mm]z2.north)
    --
    node[right,smalllbl]
    {$p_\psi$ or $f_{\bar\theta}^{-1}$}
    ([xshift=2.7mm]y2.south);

\draw[darr]
    ([xshift=2.7mm]z3.north)
    --
    node[right,smalllbl]
    {$p_\psi$ or $f_{\bar\theta}^{-1}$}
    ([xshift=2.7mm]y3.south);

\node[lbl,anchor=east] at (-6.6,-2.3) {observation-level data};
\node[lbl,anchor=east,align=left] at (-6.5,-4.25) {predictive latent\\state process};

\node[pipebox] (jobs) at (-5.1,-5.8)
{$X_{\leq t}$};

\node[pipebox] (jinf) at (-1.7,-5.8)
{$q_\theta(Z_t\mid X_{\leq t})$};

\node[pipebox] (jtrans) at (1.7,-5.8)
{$p_\phi(Z_{t+1}\mid Z_t,\xi_t)$};

\node[pipebox] (jemit) at (5.1,-5.8)
{\shortstack{
$p_\psi(X_{t+1}\mid Z_{t+1})$\\
or $f_{\bar\theta}^{-1}$
}};

\draw[pipearr] (jobs) -- (jinf);
\draw[pipearr] (jinf) -- (jtrans);
\draw[pipearr] (jtrans) -- (jemit);

\node[smalllbl,above=1mm of jobs] {observation history};
\node[smalllbl,above=1mm of jinf] {stochastic encoder};
\node[smalllbl,above=1mm of jtrans] {latent predictor};
\node[smalllbl,above=1mm of jemit] {observation map};

\end{tikzpicture}%
}
\caption{
Unified graphical and computational comparison of an HMM and full PIB-VJEPA.
\textbf{Top:} the classical HMM view, in which an unobserved Markov chain emits observations.
\textbf{Bottom:} the corresponding PIB-VJEPA view, in which stochastic encoders infer predictive latent states from observation-level data and a probabilistic predictor propagates those states.
Solid downward arrows in the PIB-VJEPA diagram denote observation-to-state encoding, while dashed upward arrows denote the optional state-to-observation map implemented by an explicit decoder or an inverse target encoder.
The boxed rows reproduce the aligned computational pipelines in \cref{eq:hmm-pipeline-intro,eq:jepa-pipeline-intro}.
}
\label{fig:hmm-jepa-graphical-intro}
\end{figure}
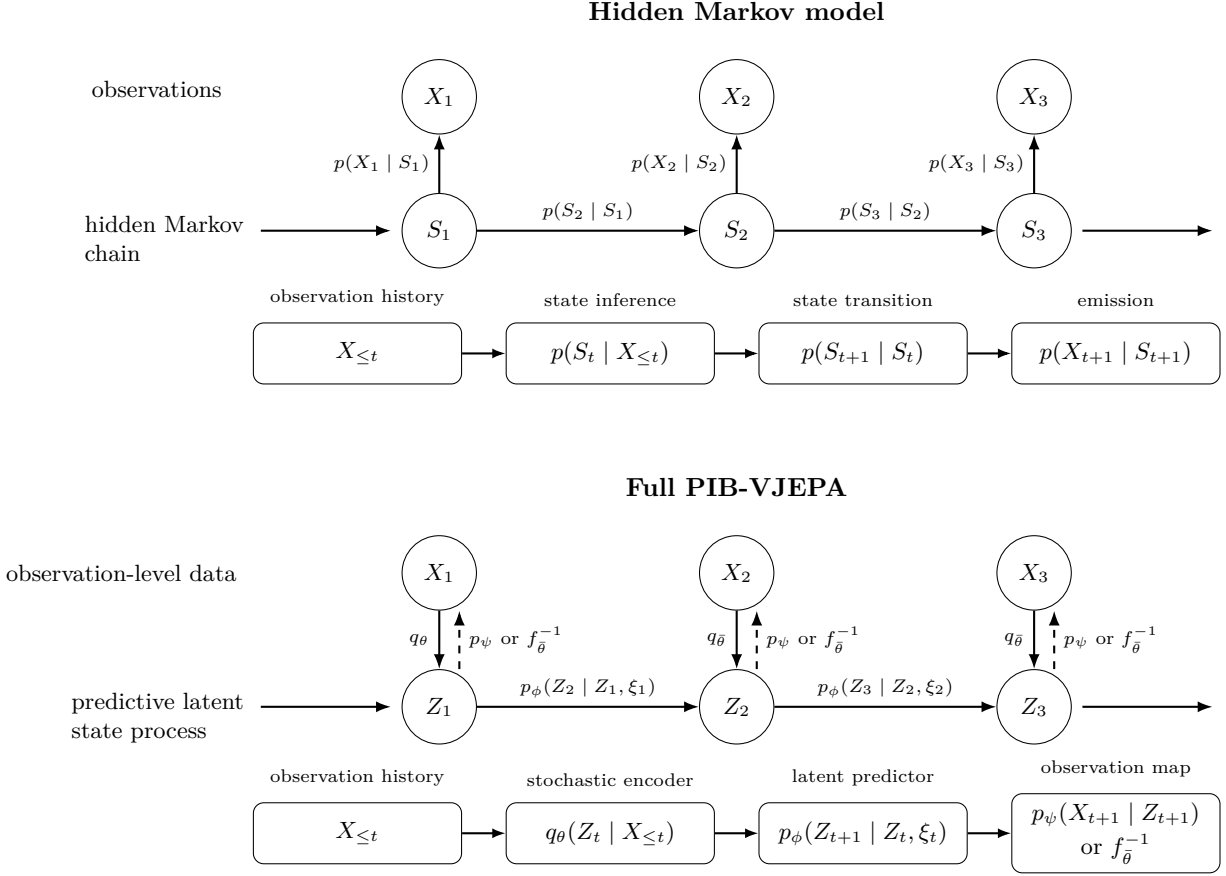

\Cref{fig:hmm-jepa-graphical-intro} combines two complementary views of this analogy, and \cref{tab:hmm-jepa-analogy-intro} summarizes the correspondence component by component. The figure separates two directions that are often conflated. In an HMM, the emission distribution maps a hidden state to an observation, whereas filtering\footnote{Here, \emph{filtering} is used in the state-space sense: it denotes inference of the current hidden-state belief $p(S_t\mid X_{\leq t})$ from the observations available up to time $t$. This belief is obtained recursively by combining the transition-based prediction from the previous state with the evidence provided by the current observation. It should not be interpreted only as noise removal, although a learned JEPA encoder may also suppress observation-level noise or other prediction-irrelevant variation.} maps observations to an inferred hidden state. Likewise, a PIB-VJEPA encoder defines a recognition or state-inference distribution rather than an emission model. Under the filtering-consistency conditions developed later, the history-dependent context encoder coincides with the corresponding HMM filtering distribution. The direct emission analogue is instead the decoder $p_\psi(X_t\mid Z_t)$ or, when available, the inverse target encoder $f_{\bar\theta}^{-1}$. The target encoder itself maps observations to latent states; only its inverse has the state-to-observation direction of an HMM emission.

\begin{table}[t]
\centering
\small
\renewcommand{\arraystretch}{1.08}
\caption{Component-wise analogy between an HMM and a full PIB-VJEPA.}
\label{tab:hmm-jepa-analogy-intro}
\begin{tabular}{
    p{0.22\linewidth}
    p{0.32\linewidth}
    p{0.38\linewidth}
}
\toprule
Component & Hidden Markov model & PIB-VJEPA \\
\midrule

Observation
&
$X_t$
&
Observation-level input $X_t$
\\

Hidden state
&
$S_t$
&
Predictive latent state $Z_t$
\\

State inference
&
$p(S_t\mid X_{\leq t})$
&
Online encoder $q_\theta(Z_t\mid X_{\leq t})$
\\

Future-state training target
&
Posterior or inferred belief over $S_{t+1}$
&
EMA target encoder $q_{\bar\theta}(Z_{t+1}\mid X_{t+1})$
\\

State transition
&
$p(S_{t+1}\mid S_t)$
&
Predictor $p_\phi(Z_{t+1}\mid Z_t,\xi_t)$
\\

Emission or observation map
&
$p(X_t\mid S_t)$
&
Decoder $p_\psi(X_t\mid Z_t)$, inverse target encoder $f_{\bar\theta}^{-1}$, or induced implicit emission
\\

\bottomrule
\end{tabular}
\end{table}

The subtlety is therefore not the high-level architecture, but its precise probabilistic and training interpretation. Standard JEPA training matches predicted future representations to stop-gradient target-encoder representations and need not optimize an observation likelihood. An explicit decoder or inverse target encoder can complete the observation-space prediction path, but the existence of such a map does not by itself make the JEPA objective identical to HMM maximum likelihood. Conversely, when neither explicit observation map is present, a local stochastic encoder can still induce an implicit emission distribution, although that induced conditional need not be tractable for generation or likelihood evaluation.

\paragraph{Scope and terminology.}
JEPA denotes the general joint-embedding predictive framework, while VJEPA denotes its probabilistic latent-prediction formulation \citep{huang2026vjepa}. Our main object of study is the \emph{full, time-indexed PIB-VJEPA} \citep{huang2026ibvjepa}, in which the current representation, future target representation, and latent transition are all probabilistic. This formulation makes the state-space structure most explicit and therefore provides the cleanest setting in which to develop the HMM correspondence. We use \emph{PIB-VJEPA} throughout the main probabilistic development, referring to JEPA or VJEPA when discussing the broader architectural family or relevant special cases. The latent Markov perspective is not restricted to PIB-VJEPA: other probabilistic temporal JEPA formulations inherit the same encode--transition structure when their context and target representations are connected by a probabilistic latent predictor, while classical deterministic temporal JEPA is recovered as the degenerate case in which the relevant latent distributions collapse to point masses and the predictor becomes a Dirac transition kernel. The sufficient-condition result developed later is stated for probabilistic temporal JEPA more generally, with full PIB-VJEPA providing the principal concrete realization. Exact sequence-level HMM equivalence nevertheless requires the additional emission and consistency conditions developed later.

Our main claim is therefore:
\begin{quote}
Full, time-indexed PIB-VJEPA exposes the encode--transition--emit structure of a hidden Markov model. Its stochastic context encoder plays the hidden-state inference or filtering role, its probabilistic predictor defines the latent-state transition, and an explicit decoder, an invertible target encoder, or an induced implicit conditional supplies the emission direction. This structural correspondence does not by itself imply that PIB-VJEPA defines the same sequence distribution or is trained by the same objective as an HMM. We therefore distinguish four progressively stronger levels: computational correspondence, emission-complete latent-state representation, sequence-level HMM equivalence, and model-and-objective equivalence. Sequence-level HMM equivalence additionally requires Markov, emission, marginal-consistency, and filtering-consistency conditions, while model-and-objective equivalence further requires the corresponding sequence-level probabilistic objective to participate in training.
\end{quote}

We develop this claim in four progressive steps\footnote{The main text presents the construction and central claims directly; detailed derivations and proofs are deferred to appendices.}:
\begin{enumerate}
    \item We introduce \emph{Markov-Chain JEPA} (\ours), in which a learned transition matrix replaces the usual latent-space predictor, and show that its matrix powers guarantee exact Chapman--Kolmogorov consistency between direct and composed multi-step predictions\footnote{The acronym MC-JEPA has previously been used for Motion-and-Content JEPA \citep{bardes2023mcjepa}. We use MCJEPA here as shorthand for \emph{Markov-Chain JEPA}, an unhyphenated, general-purpose, task-independent JEPA variant.}.
    \item We generalize the transition matrix to conditioned discrete transitions, continuous-state Markov kernels, and continuous-time dynamics, with deterministic temporal JEPA recovered as a degenerate Dirac-kernel special case.
    \item We formalize the HMM correspondence through explicit-decoder, inverse-target-encoder, and implicit-emission constructions; distinguish progressively stronger levels of correspondence; and give sufficient conditions for an exact sequence-level HMM representation.
    \item We interpret predictive information-bottleneck learning as Markov-state construction, separating \emph{minimality} through predictive compression from \emph{sufficiency} through residual predictability, and examine how JEPA latent prediction, hybrid JEPA--HMM learning, and HMM-style sequence learning impose different probabilistic semantics on the same latent-state architecture.
\end{enumerate}

We evaluate these claims in 4 controlled experiments (Section.\ref{sec:experiments}) that respectively examine finite-state transition recovery and path consistency, the filtering interpretation of the context encoder, predictive compression and Markovization, and the objective-level distinction between JEPA latent prediction and HMM-style sequence learning.

\section{From JEPA to Markov-Chain JEPA}

\subsection{Latent Markov dynamics}
\label{sec:markov-taxonomy}

Before constructing Markov-Chain JEPA, it is useful to locate it within the broader family of Markov models. Markov dynamics can be organized along two independent axes: whether time is discrete or continuous, and whether the state space is discrete or continuous. These choices give four common cases:
\begin{enumerate}
    \item \emph{Discrete time and discrete state:} the transition law is represented by a row-stochastic transition matrix. This is the setting adopted by the basic MCJEPA construction developed below.
    \item \emph{Discrete time and continuous state:} the transition law is represented by a conditional probability density or, more generally, a Markov kernel over continuous latent representations.
    \item \emph{Continuous time and discrete state:} the dynamics form a continuous-time Markov chain specified by a transition-rate generator.
    \item \emph{Continuous time and continuous state:} the dynamics may be represented by a stochastic differential equation or another continuous-time Markov process.
\end{enumerate}

We use \emph{latent Markov dynamics} as an umbrella term for these 4 cases. More specifically, \emph{Markov chain} commonly refers to a discrete-state process, whereas \emph{Markov process} or \emph{Markov kernel} also covers continuous-state models. Our main development focuses on discrete-time prediction because temporal JEPA training is typically organized around frames, tokens, or measurements indexed by discrete steps. We begin with the discrete-time, discrete-state case because it yields the most transparent connection to a classical HMM transition matrix. We then relax the fixed-matrix and discrete-state assumptions using conditioned transition matrices and continuous-state kernels. Continuous-time variants are included later to situate the framework more broadly and to accommodate irregularly sampled systems.

\subsection{The HMM--PIB-VJEPA analogy at a glance}

The construction is easiest to understand through the common encode--transition--emit pipeline summarized in \cref{tab:hmm-jepa-dictionary}. In both an HMM and full PIB-VJEPA, observations are used to infer a belief over a latent state, the state is advanced by a transition model, and the predicted state may be mapped back to observation space. Compared with a classical HMM, PIB-VJEPA primarily changes how these roles are parameterized and trained: state inference is amortized by an encoder, future-state supervision is supplied by a target encoder, and the principal predictive objective is imposed in representation space rather than through an observation-sequence likelihood.

\begin{table}[t]
\centering
\small
\renewcommand{\arraystretch}{1.08}
\caption{Direct correspondence between an HMM and full PIB-VJEPA. Observation-level inputs correspond to HMM observations, the stochastic context encoder plays the hidden-state inference role, the predictor defines the latent transition, and an explicit decoder, inverse target encoder, or induced implicit conditional can supply the emission direction.}
\label{tab:hmm-jepa-dictionary}
\begin{tabular}{@{}p{0.17\linewidth}p{0.30\linewidth}p{0.43\linewidth}@{}}
\toprule
Role & Hidden Markov model & Full PIB-VJEPA \\
\midrule

Observed variable
&
Observation $X_t$ generated by an emission model
&
Observation-level input $X_t$, such as a frame, pixel array, or time-series measurement
\\

State inference
&
Filtering distribution $p(S_t\mid X_{\leq t})$
&
Stochastic context encoder $q_\theta(Z_t\mid X_{\leq t})$
\\

Hidden state
&
Latent state $S_t$
&
Predictive latent representation $Z_t$
\\

Future-state reference
&
Posterior or inferred belief over $S_{t+1}$
&
EMA target encoder $q_{\bar\theta}(Z_{t+1}\mid X_{t+1})$
\\

State transition
&
$p(S_{t+1}\mid S_t)$ or transition matrix $A$
&
Probabilistic predictor $p_\phi(Z_{t+1}\mid Z_t,\xi_t)$
\\

Emission direction
&
$p(X_{t+1}\mid S_{t+1})$ or emission matrix $B$
&
Decoder $p_\psi(X_{t+1}\mid Z_{t+1})$, inverse target encoder $f_{\bar\theta}^{-1}$, or induced implicit emission
\\

Training signal
&
Observation likelihood or sequence ELBO
&
Latent target matching plus information-bottleneck regularization; observation decoding is optional
\\

\bottomrule
\end{tabular}
\end{table}

We now instantiate this correspondence for full PIB-VJEPA \citep{huang2026ibvjepa}. Its stochastic context encoder
\begin{equation*}
q_\theta(Z_t\mid X_{\leq t})
\end{equation*}
represents the current latent-state belief, while its probabilistic predictor
\begin{equation*}
p_\phi(Z_{t+1}\mid Z_t,\xi_t)
\end{equation*}
defines the latent-state transition. The target encoder
\begin{equation*}
q_{\bar\theta}(Z_{t+1}\mid X_{t+1})
\end{equation*}
provides the future latent distribution against which the prediction is trained. The identification of the history-dependent context encoder with an HMM filtering distribution becomes exact only under the filtering-consistency conditions developed later.

To complete the encode--transition--emit path explicitly, PIB-VJEPA may additionally use a probabilistic observation model
\begin{equation*}
p_\psi(X_{t+1}\mid Z_{t+1}).
\end{equation*}
The resulting one-step predictive observation distribution is
\begin{equation}
\begin{aligned}
p_\Theta(X_{t+1}\mid X_{\leq t})
=
\int\!\!\int
&p_\psi(X_{t+1}\mid Z_{t+1})
\,p_\phi(Z_{t+1}\mid Z_t,\xi_t)
\\[-1mm]
&\times q_\theta(Z_t\mid X_{\leq t})
\,dZ_t\,dZ_{t+1},
\end{aligned}
\label{eq:predictive-observation}
\end{equation}
where $\Theta=(\theta,\phi,\psi)$. This conditional implements the same operational sequence as HMM prediction: infer a current latent-state belief from the observation history, propagate it through the transition model, and map the predicted state to the next observation. When the context encoder coincides with the corresponding Bayesian filter, this becomes the usual HMM predictive construction.

As an alternative to introducing a separate decoder, PIB-VJEPA may use the inverse target-encoder construction introduced in \cref{eq:inverse-target-emission}. If
\begin{equation*}
f_{\bar\theta}:\cX\rightarrow\cS
\end{equation*}
is bijective on the modeled data domain, then
\begin{equation*}
\widehat X_{t+1}
=
f_{\bar\theta}^{-1}(\widehat Z_{t+1})
\end{equation*}
provides a deterministic state-to-observation map without requiring a separate decoder. A normalized emission likelihood additionally requires a tractable density, for example through a change-of-variables model or an explicit observation-noise distribution. Compatible dimensions and invertibility are strong conditions that ordinary compressed JEPA encoders generally do not satisfy, so we treat inverse target encoding as an alternative realization of the emission direction rather than a universal requirement of PIB-VJEPA.

\subsection{Temporal JEPA}

Let $X_{\leq t}$ denote observations up to time $t$, and let $X_{t+h}$ denote a future observation or target segment. A temporal JEPA uses an online encoder, an EMA target encoder, and a latent predictor \citep{huang2026vjepa}:
\begin{equation*}
Z_t=f_\theta(X_{\leq t}),
\qquad
Z^{\mathrm T}_{t+h}=f_{\bar\theta}(X_{t+h}),
\qquad
\widehat Z_{t+h}=P_\phi(Z_t,h,\xi_{t:t+h}),
\end{equation*}
where $\xi$ contains side information such as elapsed time, action, target position, or known covariates. Training matches $\widehat Z_{t+h}$ to the stop-gradient target $Z^{\mathrm T}_{t+h}$.

\subsection{MCJEPA: replace the predictor by a Markov chain}

We now specialize the JEPA latent state $Z_t$, previously allowed to be a general continuous representation, to a categorical predictive state,
\begin{equation*}
Z_t\in\{1,\ldots,K\}.
\end{equation*}
We retain $Z_t$ for the JEPA state and reserve $S_t$ for the corresponding hidden state in the HMM notation.

The online encoder returns a soft state distribution
\begin{equation*}
q_t
=
q_\theta(Z_t\mid X_{\leq t})
\in\DeltaK,
\end{equation*}
and the EMA target encoder returns
\begin{equation*}
\bar q_{t+h}
=
q_{\bar\theta}(Z_{t+h}\mid X_{t+h})
\in\DeltaK.
\end{equation*}

The predictor is a constant, time-homogeneous, row-stochastic transition matrix
\begin{equation}
A\in[0,1]^{K\times K},
\qquad
\sum_{j=1}^K A_{ij}=1.
\label{eq:time_homogeneous_transition_matrix}
\end{equation}
Its entries represent the categorical JEPA transition probabilities
\begin{equation}
A_{ij}
=
p_\phi(Z_{t+1}=j\mid Z_t=i).
\label{eq:time_homogeneous_transition_matrix_elements}
\end{equation}

The one-step and $h$-step predictive state distributions are\footnote{The linear-algebra interpretation is straightforward. The current state belief $q_t\in\mathbb R^{1\times K}$ is a row vector whose $i$th entry,
\begin{equation*}
(q_t)_i
=
q_\theta(Z_t=i\mid X_{\leq t}),
\end{equation*}
is the probability assigned to current state $i$. The transition matrix $A\in\mathbb R^{K\times K}$ is row stochastic, with
\begin{equation*}
A_{ij}
=
p_\phi(Z_{t+1}=j\mid Z_t=i),
\end{equation*}
so its $i$th row is the next-state distribution conditional on currently occupying state $i$. Their product $\widehat q_{t+1}=q_tA\in\mathbb R^{1\times K}$ is again a probability row vector, and its $j$th entry is
\begin{equation*}
(\widehat q_{t+1})_j
=
(q_tA)_j
=
\sum_{i=1}^{K}(q_t)_iA_{ij}.
\end{equation*}
Thus, the predicted probability of being in state $j$ at time $t+1$ is obtained by summing, over all possible current states $i$, the probability of currently being in state $i$ multiplied by the probability of transitioning from $i$ to $j$. Equivalently, $q_tA$ is a convex combination of the rows of $A$ \citep{strang2016introduction,boyd2018applied}, weighted by the current state belief $q_t$. Because $q_t$ is normalized and $A$ is row stochastic, the resulting vector remains normalized. The same interpretation applies to $q_tA^h$, where $(A^h)_{ij}$ is the total probability of reaching state $j$ after $h$ steps when starting from state $i$, obtained by summing the probabilities of all length-$h$ paths through the possible intermediate states. Consequently, the resulting predictive distribution depends only on the total horizon $h$, not on how that horizon is decomposed into successive transition steps.}
\begin{equation}
\widehat q_{t+1}=q_tA,
\qquad
\widehat q_{t+h}=q_tA^h.
\label{eq:time_homogeneous_multistep_transition}
\end{equation}

We train the transition matrix and online encoder by matching each predicted distribution to the corresponding target-encoder distribution:
\begin{equation}
\mathcal L_{\mathrm{MC}}
=
\E\left[
\sum_{h\in\mathcal H}
w_h
\KL\left(
\sg(\bar q_{t+h})
\,\middle\|\,
q_tA^h
\right)
\right],
\label{eq:mc-loss}
\end{equation}
where $\mathcal H$ is the set of prediction horizons and $w_h\geq0$ controls the relative importance of horizon $h$. For example, one may use uniform weights,
\begin{equation*}
w_h=\frac{1}{|\mathcal H|},
\end{equation*}
or exponentially discounted weights,
\begin{equation*}
w_h
=
\frac{\gamma^{h-1}}
{\sum_{r\in\mathcal H}\gamma^{r-1}},
\qquad
\gamma\in(0,1],
\end{equation*}
which place greater emphasis on near-term predictions; $\gamma=1$ recovers uniform weighting.

The target branch remains a slowly moving representation target. The core JEPA objective does not require observation reconstruction, although an explicit decoder or inverse target-encoder map may be used to map the predicted state back to $X_{t+h}$ for observation forecasting or HMM-style likelihood modeling.

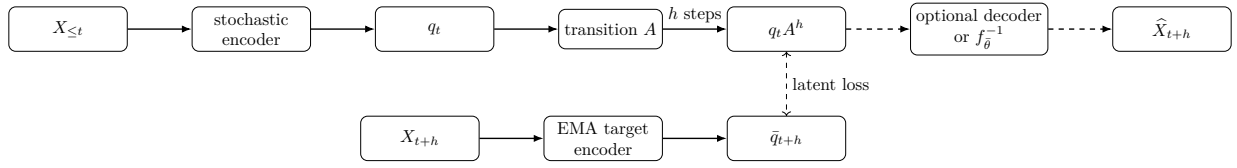
\begin{figure}[t]
\centering
\resizebox{0.98\linewidth}{!}{%
\begin{tikzpicture}[
node distance=8mm and 13mm,
box/.style={draw,rounded corners,minimum width=24mm,minimum height=9mm,align=center},
small/.style={draw,rounded corners,minimum width=19mm,minimum height=8mm,align=center},
arr/.style={-{Latex[length=2mm]},thick}
]
\node[box] (x) {$X_{\leq t}$};
\node[box,right=of x] (enc) {stochastic\\encoder};
\node[box,right=of enc] (qt) {$q_t$};
\node[small,right=of qt] (A) {transition $A$};
\node[box,right=of A] (pred) {$q_tA^h$};
\node[box,right=of pred] (emit) {optional decoder\\or $f_{\bar\theta}^{-1}$};
\node[box,right=of emit] (xhat) {$\widehat X_{t+h}$};
\node[box,below=13mm of pred] (target) {$\bar q_{t+h}$};
\node[box,left=of target] (tenc) {EMA target\\encoder};
\node[box,left=of tenc] (xf) {$X_{t+h}$};
\draw[arr] (x)--(enc);
\draw[arr] (enc)--(qt);
\draw[arr] (qt)--(A);
\draw[arr] (A)--node[above] {$h$ steps}(pred);
\draw[arr,dashed] (pred)--(emit);
\draw[arr,dashed] (emit)--(xhat);
\draw[arr] (xf)--(tenc);
\draw[arr] (tenc)--(target);
\draw[<->,dashed] (pred) -- node[right] {latent loss} (target);
\end{tikzpicture}}
\caption{Markov-Chain JEPA. The context encoder infers a predictive-state distribution, the transition matrix propagates it to future latent states, and training matches the prediction to an EMA target distribution. The dashed observation branch is optional: a decoder or inverse target encoder can map the predicted latent state back to observation space when observation forecasting or an explicit emission model is required.}
\label{fig:mcjepa}
\end{figure}

\subsection{Exact multi-horizon path consistency}

For temporal JEPA, a \textit{direct} prediction from $t$ to $t+h_1+h_2$ agrees exactly with a prediction composed through the \textit{intermediate} horizon:
\begin{equation}
q_tA^{h_1+h_2}
=
(q_tA^{h_1})A^{h_2}.
\label{eq:exact-path}
\end{equation}
This is the Chapman--Kolmogorov law\footnote{The Chapman--Kolmogorov equation states that a transition across two consecutive intervals is obtained by marginalizing over every possible intermediate state:
\(
P_{h_1+h_2}(i,k)
=
\sum_{j=1}^{K}
P_{h_1}(i,j)P_{h_2}(j,k).
\)
For a time-homogeneous chain, $P_h=A^h$, so $A^{h_1+h_2}=A^{h_1}A^{h_2}$; left-multiplying by the current state distribution $q_t$ gives \cref{eq:exact-path}.} for a time-homogeneous finite-state chain.

\begin{proposition}[Exact path consistency]
\label{prop:exact_path_consistency}
Assume that the latent dynamics form a time-homogeneous finite-state Markov chain with a fixed row-stochastic transition matrix $A$. Then, for any state distribution $q_t$ and nonnegative integers $h_1,h_2$,
\begin{equation*}
q_tA^{h_1+h_2}
=
\left(q_tA^{h_1}\right)A^{h_2}.
\end{equation*}
Consequently, all prediction paths whose transition lengths sum to the same total horizon produce the same predictive distribution.
\end{proposition}

A proof is provided in \cref{app:proof-exact-path-consistency}.
Proposition~\ref{prop:exact_path_consistency} is useful for combining short- and long-horizon planning because a long-horizon prediction can be computed either directly or by composing shorter transitions without introducing path-dependent discrepancies. This supports hierarchical planning and temporal abstraction while ensuring that every decomposition of the same total horizon yields a consistent predictive distribution.

A softer, less constrained alternative learns a separate matrix $A_h$ for each horizon and penalizes
\begin{equation*}
\mathcal L_{\mathrm{CK}}
=
\sum_{h_1,h_2}
\left\|
A_{h_1+h_2}-A_{h_1}A_{h_2}
\right\|_F^2.
\end{equation*}
This variant tests whether one homogeneous chain is adequate or whether the data require horizon-dependent dynamics.

\subsection{Avoiding discrete-state collapse}

The Markov-chain prediction objective in \cref{eq:mc-loss} does not by itself guarantee that the categorical latent states contain useful information. Because the online encoder, target encoder, and transition matrix are learned jointly, they may agree through a degenerate representation. This is the discrete-state analogue of representation collapse in deterministic self-supervised learning.

For each time index $t$, the online encoder produces
\begin{equation*}
q_t
=
\bigl(q_{t1},\ldots,q_{tK}\bigr),
\qquad
q_{tk}
=
q_\theta(Z_t=k\mid X_{\leq t}),
\end{equation*}
where $q_{tk}$ is the probability that the current observation history is represented by latent state $k$.

Two simple degenerate solutions are particularly important. First, all observations may be assigned to the same state:
\begin{equation*}
q_t\approx e_j
\qquad
\text{for every }t,
\end{equation*}
where $e_j\in\{0,1\}^K$ is the $j$th standard basis vector, with $(e_j)_j=1$ and $(e_j)_k=0$ for all $k\neq j$. The transition matrix can then place nearly all probability on the self-transition $A_{jj}$, allowing the online and target branches to agree without learning meaningful temporal structure. We refer to this failure mode as \emph{single-state collapse}.

Second, every observation may receive the same uniform assignment:
\begin{equation*}
q_t
\approx
\operatorname{Unif}(K)
=
\left(\frac{1}{K},\ldots,\frac{1}{K}\right).
\end{equation*}
A transition matrix that preserves the uniform distribution can again produce consistent predictions even though the latent state contains no information about the observation. We refer to this as \emph{uniform-assignment collapse}.

To measure how the available states are used across a minibatch $\mathcal B$, define the average state occupancy
\begin{equation*}
\bar q_{\mathcal B}
=
\frac{1}{|\mathcal B|}
\sum_{t\in\mathcal B}q_t.
\end{equation*}
We encourage aggregate use of the state space through
\begin{equation*}
\KL\left(
\bar q_{\mathcal B}
\,\middle\|\,
\operatorname{Unif}(K)
\right).
\end{equation*}
This term is zero when aggregate occupancy is uniform and increases when most probability mass is concentrated on only a few states, thereby discouraging single-state collapse and unused states.

Aggregate diversity alone is not sufficient. If every observation receives the uniform distribution, then
\begin{equation*}
\bar q_{\mathcal B}
=
\operatorname{Unif}(K),
\end{equation*}
so the occupancy penalty is again zero. We therefore also control the entropy of each assignment,
\begin{equation*}
H(q_t)
=
-\sum_{k=1}^{K}
q_{tk}\log q_{tk}.
\end{equation*}
Entropy is maximal at $\log K$ for a uniform assignment and minimal at zero for a one-hot assignment. Because this entropy appears with a positive weight in a minimized loss, it encourages comparatively confident, low-entropy assignments.

Combining the two effects gives
\begin{equation}
\mathcal L_{\mathrm{state}}
=
\underbrace{
\lambda_{\mathrm{occ}}
\KL\left(
\bar q_{\mathcal B}
\,\middle\|\,
\operatorname{Unif}(K)
\right)
}_{\text{occupancy}}
+
\underbrace{
\lambda_{\mathrm{ent}}
\frac{1}{|\mathcal B|}
\sum_{t\in\mathcal B}
H(q_t)
}_{\text{entropy}},
\label{eq:state-reg}
\end{equation}
where $\lambda_{\mathrm{occ}},\lambda_{\mathrm{ent}}\geq0$. The complete basic MCJEPA objective is therefore
\begin{equation*}
\mathcal L
=
\mathcal L_{\mathrm{MC}}
+
\mathcal L_{\mathrm{state}}.
\end{equation*}

The two regularizers in $\mathcal L_{\mathrm{state}}$ play complementary roles. The occupancy term encourages \emph{diversity across observations}, while the entropy term encourages \emph{confidence within each observation}. Their combination therefore favors balanced but informative assignments: different observations can occupy different states while each individual observation receives a comparatively concentrated state distribution.

The relative weights must nevertheless be selected carefully. If $\lambda_{\mathrm{occ}}$ is too large, the model may artificially force every minibatch to use all states even when the underlying state distribution is imbalanced. If $\lambda_{\mathrm{ent}}$ is too large, the encoder may make prematurely hard and unstable assignments. In practice, the entropy weight may be introduced gradually, and the uniform occupancy target may be replaced by a nonuniform prior $\pi$ when unequal state frequencies are expected:
\begin{equation*}
\KL\left(
\bar q_{\mathcal B}
\,\middle\|\,
\pi
\right).
\end{equation*}

\section{From a Transition Matrix to a Neural Markov Kernel}

The taxonomy in \cref{sec:markov-taxonomy} places the basic MCJEPA construction in the \emph{discrete-time, discrete-state} class. Its transition matrix is the simplest realization of latent Markov dynamics: the predictive state is categorical, time advances in discrete steps, and, in the time-homogeneous case, the same transition matrix is applied at every step. As shown in \cref{eq:time_homogeneous_multistep_transition}, this gives the particularly simple multi-step prediction
\begin{equation*}
\widehat q_{t+h}
=
q_tA^h.
\end{equation*}

The Markov principle itself is more general. It requires that the next-state distribution depend on the past only through the current predictive state and the transition-relevant side information\footnote{See sufficiency of predictive state \citep{huang2026vjepa}.}:
\begin{equation}
p(Z_{t+1}\mid Z_{\leq t},\xi_{\leq t})
=
p(Z_{t+1}\mid Z_t,\xi_t).
\label{eq:markov-property}
\end{equation}
The transition may therefore be fixed or conditioned, and the latent state may be discrete or continuous. We now generalize the fixed transition matrix progressively while preserving this conditional-independence structure.

\subsection{Neural discrete-state transitions}

The fixed matrix $A$ assumes that the transition law is time homogeneous and independent of external information. A more expressive model allows a neural network to produce a row-stochastic transition matrix conditioned on side information:
\begin{equation}
A_t
=
A_\phi(\xi_t),
\qquad
A_t\in[0,1]^{K\times K},
\qquad
\sum_{j=1}^{K}(A_t)_{ij}=1.
\label{eq:conditioned-transition}
\end{equation}
The side information $\xi_t$ may contain an action, elapsed time, goal, control input, regime indicator, or exogenous covariates. Its entries have the interpretation
\begin{equation*}
(A_t)_{ij}
=
p_\phi(Z_{t+1}=j\mid Z_t=i,\xi_t).
\end{equation*}

For a sequence of conditioned transitions, the $h$-step predictive distribution is
\begin{equation}
\begin{aligned}
\widehat q_{t+h}
&=
q_t
\prod_{j=0}^{h-1}
A_\phi(\xi_{t+j})
\\
&=
q_t
A_\phi(\xi_t)
A_\phi(\xi_{t+1})
\cdots
A_\phi(\xi_{t+h-1}),
\end{aligned}
\label{eq:conditioned-multistep}
\end{equation}
where the product is ordered chronologically from left to right. The transition matrices at different steps need not commute, so this ordering matters. In the time-homogeneous, unconditioned case,
\begin{equation*}
A_\phi(\xi_{t+j})=A
\qquad
\text{for all }j,
\end{equation*}
and \cref{eq:conditioned-multistep} reduces to the $q_tA^h$ construction in \cref{eq:time_homogeneous_multistep_transition}.

This construction supports action-conditioned dynamics, changing goals or regimes, exogenous covariates, and irregular temporal intervals when elapsed time is included in $\xi_t$, while retaining a discrete and interpretable latent state space. It also preserves exact path composition whenever the same chronologically ordered transition sequence is used along both prediction paths.

\subsection{Continuous-state stochastic transitions}

A categorical state may be too restrictive when the predictive representation varies continuously. In this case, the transition matrix is replaced by a Markov kernel
\begin{equation*}
p_\phi(Z_{t+1}\mid Z_t,\xi_t),
\end{equation*}
which assigns a probability distribution over the next continuous latent state for each current state and side-information value.

A simple example is a Gaussian transition,
\begin{equation}
p_\phi(Z_{t+1}\mid Z_t,\xi_t)
=
\mathcal N\!\left(
Z_{t+1};
\mu_\phi(Z_t,\xi_t),
\Sigma_\phi(Z_t,\xi_t)
\right),
\label{eq:continuous-kernel}
\end{equation}
where the neural network outputs a conditional mean $\mu_\phi$ and a valid covariance matrix $\Sigma_\phi$. The mean describes the expected latent evolution, while the covariance represents stochastic uncertainty and unresolved variation around that mean.

The two-step transition is obtained by marginalizing over the intermediate latent state:
\begin{equation*}
\begin{aligned}
&p_\phi^{(2)}
\left(
Z_{t+2}
\mid
Z_t,\xi_{t:t+1}
\right)
\\
&\qquad=
\int
p_\phi(Z_{t+2}\mid Z_{t+1},\xi_{t+1})
p_\phi(Z_{t+1}\mid Z_t,\xi_t)
\,dZ_{t+1}.
\end{aligned}
\end{equation*}
More generally,
\begin{equation}
\begin{aligned}
&p_\phi^{(h)}
\left(
Z_{t+h}
\mid
Z_t,\xi_{t:t+h-1}
\right)
\\
&\qquad=
\int
\prod_{j=0}^{h-1}
p_\phi(
Z_{t+j+1}
\mid
Z_{t+j},
\xi_{t+j})
\,dZ_{t+1}\cdots dZ_{t+h-1}.
\end{aligned}
\label{eq:kernel-composition}
\end{equation}
This is the continuous-state analogue of multiplying transition matrices in \cref{eq:time_homogeneous_multistep_transition,eq:conditioned-multistep}: all possible intermediate latent states are marginalized out. Closed-form composition is available only for restricted transition families, so neural models may instead use sampling, Monte Carlo integration, moment propagation, or learned approximations to multi-step prediction.

A single Gaussian kernel cannot represent a genuinely multimodal conditional distribution. When multiple distinct futures are important, the same construction can instead use mixtures \citep{huang2026GMM}, normalizing flows, diffusion-based transitions, or other expressive conditional distributions. The central requirement is not Gaussianity, but the Markov factorization in \cref{eq:markov-property}.

\subsection{Continuous-time extensions}

Suppose that observations are recorded at physical times
\begin{equation*}
\tau_1<\tau_2<\cdots,
\end{equation*}
with elapsed interval
\begin{equation*}
\Delta t_t
=
\tau_{t+1}-\tau_t.
\end{equation*}
Irregular sampling can already be handled within a discrete-time model by including the observed interval in the side information,
\begin{equation*}
\xi_t
=
(\widetilde\xi_t,\Delta t_t),
\end{equation*}
where $\widetilde\xi_t$ contains the remaining actions, controls, goals, or exogenous covariates. The resulting discrete-time transition model then learns how state evolution changes with the supplied propagation interval.

A more explicit alternative is to model the latent dynamics directly in continuous time. For a discrete latent state, a continuous-time Markov chain is specified by a generator matrix
\begin{equation*}
Q_\phi(\widetilde\xi_t),
\end{equation*}
whose off-diagonal entries are nonnegative transition rates and whose rows sum to zero. If the generator is held fixed over a propagation interval of duration $\Delta t\geq0$, the corresponding transition matrix is
\begin{equation}
A_\phi(\Delta t,\widetilde\xi_t)
=
\exp\!\left(
\Delta t\,Q_\phi(\widetilde\xi_t)
\right).
\label{eq:ctmc-transition}
\end{equation}
Here, $\Delta t$ denotes a generic continuous-time propagation duration, whereas $\Delta t_t=\tau_{t+1}-\tau_t$ denotes the particular interval between the $t$th and $(t+1)$th observations. Under a piecewise-constant conditioning assumption over this interval,
\begin{equation*}
A_t
=
A_\phi(\Delta t_t,\widetilde\xi_t)
=
\exp\!\left(
\Delta t_t\,Q_\phi(\widetilde\xi_t)
\right).
\end{equation*}
The same generator can therefore propagate the latent state across observation intervals of different lengths or across arbitrary forecasting horizons. If the conditioning variables vary continuously within an interval, the corresponding transition requires the appropriate time-varying generator composition rather than a single matrix exponential.

For a continuous latent state, a continuous-time stochastic transition may instead be represented by a stochastic differential equation. Using $\tau$ for continuous physical time to distinguish it from the discrete observation index,
\begin{equation}
dZ(\tau)
=
b_\phi(Z(\tau),\xi(\tau))\,d\tau
+
G_\phi(Z(\tau),\xi(\tau))\,dW(\tau),
\label{eq:latent-sde}
\end{equation}
where $b_\phi$ is the drift, $G_\phi$ controls the diffusion, and $W(\tau)$ is a Wiener process. The drift describes systematic latent evolution, while the diffusion represents stochastic transition uncertainty. Setting $G_\phi=0$ recovers deterministic continuous-time dynamics such as a neural ordinary differential equation.

These continuous-time constructions are not required for our main MCJEPA development, but they show that the transition-matrix formulation belongs to a broader family of latent Markov models.

\subsection{Deterministic temporal JEPA as a degenerate kernel}
\label{sec:deterministic-jepa-degenerate}

A deterministic temporal JEPA uses a predictor
\begin{equation*}
\widehat Z_{t+1}
=
g_\phi(Z_t,\xi_t).
\end{equation*}
Probabilistically, this is the Dirac transition kernel
\begin{equation}
p_\phi(Z_{t+1}\mid Z_t,\xi_t)
=
\delta\!\left(
Z_{t+1}-g_\phi(Z_t,\xi_t)
\right).
\label{eq:dirac-transition}
\end{equation}
All conditional probability mass is concentrated at the predictor output, so the transition contains no intrinsic stochastic uncertainty. For example, if the Gaussian mean in \cref{eq:continuous-kernel} satisfies
\begin{equation*}
\mu_\phi(Z_t,\xi_t)=g_\phi(Z_t,\xi_t),
\end{equation*}
then the Gaussian transition approaches this deterministic case as
\begin{equation*}
\Sigma_\phi(Z_t,\xi_t)\rightarrow 0.
\end{equation*}

Classical deterministic temporal JEPA is therefore not separate from the latent Markov-kernel perspective. A deterministic encoder may be represented probabilistically by a point-mass state distribution, while a deterministic predictor is represented by the Dirac transition in \cref{eq:dirac-transition}. MCJEPA, conditioned discrete-state JEPA, continuous probabilistic VJEPA, and deterministic JEPA can thus be viewed within the same state-space framework, differing primarily in their state representation and transition family.

\begin{table}[t]
\centering
\small
\renewcommand{\arraystretch}{1.08}
\caption{Representative hierarchy of latent Markov transitions for temporal JEPA. The main development focuses on discrete-time models; continuous-time variants situate the transition-matrix formulation within the broader state-space family.}
\label{tab:hierarchy}
\begin{tabular}{@{}p{0.25\linewidth}p{0.17\linewidth}p{0.18\linewidth}p{0.31\linewidth}@{}}
\toprule
Model & Time & State & Transition \\
\midrule

MCJEPA
&
Discrete
&
Discrete
&
Fixed matrix $A$
\\

Conditioned MCJEPA
&
Discrete
&
Discrete
&
$A_\phi(\xi_t)$
\\

Continuous-state Markov JEPA
&
Discrete
&
Continuous
&
Kernel $p_\phi(Z_{t+1}\mid Z_t,\xi_t)$
\\

Deterministic temporal JEPA
&
Discrete
&
Typically continuous
&
Dirac kernel at $g_\phi(Z_t,\xi_t)$
\\

Continuous-time MCJEPA
&
Continuous
&
Discrete
&
Generator $Q_\phi$ and $\exp(\Delta t Q_\phi)$
\\

Continuous-time stochastic JEPA
&
Continuous
&
Continuous
&
Latent SDE or continuous-time kernel
\\

\bottomrule
\end{tabular}
\end{table}

\subsection{When a recurrent predictor is Markov}

A recurrent predictor may appear to violate the first-order Markov assumption because its prediction can depend on a summary of the entire preceding latent history. Let $M_t$ denote a recurrent memory state and suppose that the next latent state is predicted according to
\begin{equation} \label{eq:MC_memory}
p_\phi(Z_{t+1}\mid Z_t,M_t,\xi_t).
\end{equation}
After sampling or predicting $Z_{t+1}$, the recurrent memory may be updated deterministically as
\begin{equation*}
M_{t+1}
=
r_\phi(M_t,Z_{t+1},\xi_t).
\end{equation*}

The process need not be first-order Markov in $Z_t$ alone, because two histories yielding the same $Z_t$ but different memories $M_t$ may induce different next-state distributions. However, defining the augmented state
\begin{equation*}
\widetilde Z_t
=
(Z_t,M_t)
\end{equation*}
restores a first-order representation:
\begin{equation*}
p_\phi(
\widetilde Z_{t+1}
\mid
\widetilde Z_{\leq t},
\xi_{\leq t})
=
p_\phi(
\widetilde Z_{t+1}
\mid
\widetilde Z_t,
\xi_t).
\end{equation*}

This distinction motivates a central representation-learning objective of PIB-VJEPA: ideally, the learned predictive state $Z_t$ itself should summarize the information from the observation history that is relevant to future prediction. When this succeeds, a simple first-order transition in $Z_t$ is sufficient. When substantial predictive information remains outside $Z_t$, the model must either enlarge the state, augment it with memory, use higher-order dynamics, or accept that the latent process is not first-order Markov in the chosen representation.

\section{A Probabilistic JEPA Is Secretly an HMM}

\subsection{The three distributions in PIB-VJEPA}

The full, time-indexed PIB-VJEPA considered here contains three conditional distributions \citep{huang2026ibvjepa}:
\begin{align*}
q_\theta(Z_t\mid X_{\leq t})
&\quad\text{current-state encoder},
\\
p_\phi(Z_{t+1}\mid Z_t,\xi_t)
&\quad\text{latent transition predictor},
\\
q_{\bar\theta}(Z_{t+1}\mid X_{t+1})
&\quad\text{future target encoder}.
\end{align*}
The online encoder $q_\theta$ maps the observation history $X_{\leq t}$ to a distribution over the current predictive state $Z_t$. The transition model $p_\phi$ propagates that state to a distribution over $Z_{t+1}$, possibly conditioned on side information $\xi_t$ such as an action, elapsed time, or exogenous covariates. The target encoder $q_{\bar\theta}$ provides the future latent distribution against which the prediction is trained. Here, $\bar\theta$ denotes the slowly updated target-encoder parameters, typically obtained as an exponential moving average of the online parameters $\theta$.

A compact form of the PIB-VJEPA objective is \citep{huang2026ibvjepa}:
\begin{equation}
\begin{aligned}
\mathcal L_{\mathrm{PIB\text{-}VJEPA}}
={}&
\E\left[
-\log p_\phi(Z_{t+1}\mid Z_t,\xi_t)
\right]
\\
&+
\gamma_{\mathrm S}
\E\KL\left(
q_\theta(Z_t\mid X_{\leq t})
\,\middle\|\,
p_{\mathrm{ref}}^{\mathrm S}(Z_t)
\right)
\\
&+
\beta_{\mathrm T}
\E\KL\left(
q_{\bar\theta}(Z_{t+1}\mid X_{t+1})
\,\middle\|\,
p_{\mathrm{ref}}^{\mathrm T}(Z_{t+1})
\right).
\end{aligned}
\label{eq:pib-vjepa}
\end{equation}
The expectation is taken over training sequences, side information, and latent samples from the online and target encoders. The superscripts $\mathrm S$ and $\mathrm T$ label the \emph{source-side} current state and \emph{target-side} future state, respectively. Accordingly, $p_{\mathrm{ref}}^{\mathrm S}$ and $p_{\mathrm{ref}}^{\mathrm T}$ are reference prior distributions for the current and future latent states. Depending on the latent family, these may be standard Gaussian, uniform categorical, or other suitably chosen simple distributions.

The nonnegative coefficients
\begin{equation*}
\gamma_{\mathrm S}\geq0,
\qquad
\beta_{\mathrm T}\geq0
\end{equation*}
control the strengths of the source- and target-side regularization. Increasing $\gamma_{\mathrm S}$ places greater pressure on the current representation to compress the observation history, while increasing $\beta_{\mathrm T}$ more strongly regularizes the future target representation. These coefficients therefore trade predictive accuracy against latent compression and regularity.

The first term in \cref{eq:pib-vjepa} encourages the current state to preserve information needed to predict the future target state. The second term promotes compression of the observation history by regularizing $q_\theta(Z_t\mid X_{\leq t})$ toward the reference prior $p_{\mathrm{ref}}^{\mathrm S}$, while the third regularizes the future target distribution toward $p_{\mathrm{ref}}^{\mathrm T}$. Together, the three terms encourage a predictive latent state while controlling the information retained in its stochastic representation.

\subsection{The encode--transition--emit correspondence}

A conventional HMM factorizes as \citep{rabiner1989hmm}
\begin{equation}
p(s_{1:T},x_{1:T})
=
p(s_1)
\prod_{t=1}^{T-1}
p(s_{t+1}\mid s_t)
\prod_{t=1}^{T}
p(x_t\mid s_t).
\label{eq:hmm}
\end{equation}
The direct correspondence\footnote{For notational simplicity, \cref{eq:hmm} shows the unconditioned case. Corresponding to JEPA, when observed side information is present, replace $p(s_{t+1}\mid s_t)$ by $p(s_{t+1}\mid s_t,\xi_t)$ and interpret the sequence factorization conditional on $\xi_{1:T-1}$. To keep the encoder notation compact, we usually suppress past side information in $q_\theta(Z_t\mid X_{\leq t})$; when relevant, it should be read as $q_\theta(Z_t\mid X_{\leq t},\xi_{<t})$.} was summarized in \cref{tab:hmm-jepa-dictionary}. Observation-level data play the role of HMM observations, $Z_t$ is the predictive latent state, and $p_\phi(Z_{t+1}\mid Z_t,\xi_t)$ specifies its transition dynamics. The history-dependent context encoder
\begin{equation*}
q_\theta(Z_t\mid X_{\leq t})
\end{equation*}
plays the \emph{state-inference} role. It coincides with the Bayesian filtering distribution of the corresponding HMM only under the filtering-consistency conditions developed below.

The remaining HMM component is the state-to-observation direction. There are three ways in which this direction can be supplied.

\paragraph{Explicit decoder.}
A probabilistic decoder
\begin{equation*}
p_\psi(X_t\mid Z_t)
\end{equation*}
is the literal analogue of an HMM emission distribution. Such a decoder may be trained jointly with the latent model or added after representation learning, depending on whether observation-space likelihood or forecasting is part of the training objective.

\paragraph{Inverse target encoder.}
If the target encoder is bijective on the modeled data domain, its inverse supplies a deterministic state-to-observation map:
\begin{equation*}
Z_t^{\mathrm T}
=
f_{\bar\theta}(X_t),
\qquad
X_t
=
f_{\bar\theta}^{-1}(Z_t^{\mathrm T}).
\end{equation*}
The associated emission can be represented as a Dirac kernel concentrated at $f_{\bar\theta}^{-1}(Z_t)$. A deterministic inverse alone, however, does not provide a non-degenerate observation density. When normalized likelihood evaluation is required, the inverse must form part of a tractable probabilistic density model, for example through an appropriate change-of-variables construction or explicit observation-noise model. Standard JEPA target encoders are typically compressive and therefore not exactly invertible, so inverse target encoding is an alternative architectural realization rather than an assumption of the general theory.

\paragraph{Implicit emission.}
When neither an explicit decoder nor an invertible target encoder is available, a local stochastic encoder can still induce a state-to-observation conditional. We develop this construction next.

In all three cases, the stochastic encoder itself should not be identified with the emission distribution: it maps observations to latent states, whereas an HMM emission maps latent states to observations.

\subsection{When no decoder is present: an implicit emission}

Suppose that a local stochastic encoder is available,
\begin{equation*}
q_\theta(z\mid x),
\end{equation*}
and let $p_{\mathrm{data}}(x)$ denote the observation marginal. Define the induced latent marginal
\begin{equation*}
q_\theta(z)
=
\int
p_{\mathrm{data}}(x)
q_\theta(z\mid x)
\,dx.
\end{equation*}
Whenever $q_\theta(z)>0$, define
\begin{equation}
p_\theta^{\mathrm{imp}}(x\mid z)
=
\frac{
p_{\mathrm{data}}(x)
q_\theta(z\mid x)
}{
q_\theta(z)
}.
\label{eq:implicit-emission}
\end{equation}

\begin{proposition}[Implicit emission completion]
\label{prop:implicit-emission-completion}
The conditional distribution in \cref{eq:implicit-emission} is normalized and satisfies
\begin{equation*}
q_\theta(z\mid x)
=
\frac{
p_\theta^{\mathrm{imp}}(x\mid z)
q_\theta(z)
}{
p_{\mathrm{data}}(x)
}.
\end{equation*}
Thus, every local stochastic encoder together with the data marginal defines a one-time latent-variable model for which the encoder is the exact posterior.
\end{proposition}

A proof of normalization and Bayes consistency is provided in \cref{app:proof-implicit-emission-completion}.
The construction is a direct application of Bayes' rule and establishes an exact \emph{static} observation--state correspondence. It does not by itself establish a sequence-level HMM. For that stronger claim, the induced state distributions must evolve consistently with the latent transition, and the history-dependent context encoder must agree with the Bayesian filtering distribution induced by the transition and emission models.

\subsection{Four levels of correspondence}

The statement that probabilistic temporal JEPA is ``secretly an HMM'' is not all-or-nothing. The correspondence becomes progressively stronger as additional state-space semantics are imposed. We distinguish four levels.

\paragraph{1. Computational correspondence.}
At the weakest level, the two architectures expose the same computational roles:
\begin{equation*}
\text{observation-to-state inference}
\;\longrightarrow\;
\text{state transition}
\;\longrightarrow\;
\text{state-to-observation prediction}.
\end{equation*}
For PIB-VJEPA these roles are implemented by the context encoder, latent predictor, and one of the observation-map constructions above. This correspondence does not by itself imply a common joint distribution or training objective.

\paragraph{2. Emission-complete latent-state representation.}
The correspondence becomes probabilistically more explicit once a valid state-to-observation conditional is available. This conditional may be parameterized directly by a decoder, supplied deterministically by an invertible target encoder, or induced implicitly through \cref{eq:implicit-emission}. Together with a valid latent transition, these components provide the ingredients of a latent Markov model. They do not yet guarantee, however, that the learned history encoder is the Bayesian filter of that model or that its state marginals are dynamically consistent.

\paragraph{3. Sequence-level HMM equivalence.}
A stronger statement holds when the transition, emission, latent marginals, and history-dependent encoder are mutually consistent. Under these conditions, the latent-state model admits an HMM factorization at the sequence level rather than merely sharing its components.

\begin{theorem}[Sufficient conditions for an exact HMM representation]
\label{thm:exact-hmm-representation}
Consider a probabilistic temporal JEPA with latent state $Z_t$ and observation process $X_t$. The full JEPA system is consistent with an exact HMM representation, meaning that its latent transition and emission define an HMM sequence model and its history-dependent encoder coincides with the corresponding filtering distribution, provided that:
\begin{enumerate}
    \item the latent dynamics satisfy the first-order Markov property;
    \item a valid state-to-observation conditional is specified by an explicit decoder, an invertible target encoder, or the implicit construction in \cref{eq:implicit-emission};
    \item the latent-state marginals are consistent with the transition kernel; and
    \item the history-dependent encoder coincides with the filtering posterior induced by the corresponding transition and emission models.
\end{enumerate}
For the implicit-emission construction, the local encoder must additionally satisfy the required observation-locality and Bayes-consistency conditions. Under these assumptions, the resulting joint sequence distribution admits the HMM factorization in \cref{eq:hmm}, and the history-dependent encoder is the corresponding filtering distribution. These conditions are sufficient rather than necessary and are not guaranteed by the standard JEPA training objective.
\end{theorem}

A complete statement of the marginal-consistency and filtering conditions, together with the proof, is provided in \cref{app:hmm-equivalence}.

\paragraph{4. Model-and-objective equivalence.}
Given sequence-level HMM equivalence, the strongest correspondence additionally concerns how that probabilistic model is trained. Standard probabilistic JEPA training predicts target representations and regularizes their information content; it does not generally maximize the observation-sequence likelihood
\begin{equation*}
\log p(X_{1:T})
\end{equation*}
or optimize a conventional HMM sequence-evidence objective. Full model-and-objective equivalence requires the resulting HMM-compatible sequence model to be trained directly by its observation-sequence likelihood, or by the corresponding sequence-evidence objective when exact marginalization is unavailable. Hybrid objectives occupy an intermediate regime because they retain the original JEPA latent-prediction objective alongside HMM-style sequence and filtering supervision rather than replacing it. We develop these alternatives in \cref{app:hmm-style-training} and examine their behavior experimentally in \cref{sec:exp-hmm-training}.

The hierarchy therefore separates four distinct claims: sharing HMM-like computational roles, possessing an emission-complete latent-state representation, defining the same sequence-level probabilistic model, and additionally training that model with an HMM-style sequence objective.

\section{Information Bottleneck Learning as Markovization}

At the information-theoretic level, predictive information bottleneck learning seeks a representation that compresses the past while retaining predictive information about the future \citep{tishby1999ib,bialek2001predictive,alemi2017vib}. A schematic latent-space objective is
\begin{equation}
\min_{q_\theta(Z_t\mid X_{\leq t})}
\MI(X_{\leq t};Z_t)
-
\lambda\MI(Z_t;Z_{t+1}),
\label{eq:pib}
\end{equation}
where the first term penalizes information retained from the observation history and the second rewards predictive dependence between the current and future latent states. The trade-off parameter $\lambda$ controls the relative emphasis on compression and prediction. The practical PIB-VJEPA objective in \cref{eq:pib-vjepa} implements this (variational) principle through latent prediction together with variational bottleneck regularization \citep{huang2026ibvjepa}.

Importantly, minimizing \cref{eq:pib} does not by itself guarantee that the learned representation is predictively sufficient. A stronger ideal target is that $Z_t$ retain all information in the observation history that is relevant to the future. In conditional-independence form,
\begin{equation}
X_{>t}
\perp
X_{\leq t}
\mid
Z_t.
\label{eq:predictive-sufficiency}
\end{equation}
This is the predictive-state condition: once $Z_t$ is known, the remaining observation history provides no additional information about the future. It is closely related to predictive-state representations \citep{littman2001psr}.

For one-step latent prediction, a corresponding Markov-sufficiency condition is
\begin{equation}
\MI(Z_{t+1};X_{<t}\mid Z_t)=0.
\label{eq:markovization}
\end{equation}
Thus, after conditioning on the current predictive state, older observation history contains no additional information about the next latent state. When transition-relevant side information $\xi_t$ is present, the analogous diagnostic additionally conditions on $\xi_t$.

\begin{proposition}[Predictive sufficiency implies one-step Markov sufficiency]
\label{prop:predictive-sufficiency-markov}
Assume that
\begin{equation*}
X_{>t}
\perp
X_{\leq t}
\mid
Z_t,
\end{equation*}
and that the future target state is generated from the next observation as
\begin{equation*}
Z_{t+1}
=
g_{\bar\theta}(X_{t+1},U_{t+1}),
\end{equation*}
where the target-encoder randomness satisfies
\begin{equation*}
U_{t+1}
\perp
X_{\leq t}
\mid
(X_{t+1},Z_t).
\end{equation*}
Then
\begin{equation*}
\MI(Z_{t+1};X_{<t}\mid Z_t)=0.
\end{equation*}
Hence, predictive sufficiency implies that the learned state screens off older observation history from the next latent state, giving a one-step Markov-sufficient representation at the chosen prediction scale.
\end{proposition}

The proof follows from the conditional data-processing inequality and is provided in \cref{app:proof-predictive-sufficiency}.

\section{Residual Predictability as a Diagnostic of Markov Sufficiency}
\label{sec:residual-diagnostic}

The Markov-sufficiency condition in \cref{eq:markovization} is difficult to verify directly in a learned, high-dimensional representation. A more operational approach is to ask whether prediction errors retain systematic dependence on information preceding the current state. If older observations or latent states improve prediction after the current representation and transition-relevant side information have been accounted for, then the current state--predictor pair has not captured all transition-relevant information.

Residual diagnostics test consequences of Markov sufficiency rather than the full conditional-independence condition itself. In particular, the diagnostics below focus primarily on conditional-mean predictability. Detecting residual predictability from older history therefore provides evidence against Markov sufficiency, whereas failing to detect it does not prove the full Markov property.

\subsection{Categorical probability innovations}

For categorical MCJEPA, let
\begin{equation*}
q_t
=
q_\theta(Z_t\mid X_{\leq t})
\end{equation*}
denote the current-state distribution, and let
\begin{equation*}
\bar q_{t+1}
=
q_{\bar\theta}(Z_{t+1}\mid X_{t+1})
\end{equation*}
denote the target-encoder distribution at the next time step. Write $A_t=A$ for the time-homogeneous model and $A_t=A_\phi(\xi_t)$ for an input-conditioned transition. The predicted next-state distribution is
\begin{equation*}
\widehat q_{t+1}
=
q_tA_t.
\end{equation*}
We define the categorical probability innovation as
\begin{equation}
R_{t+1}
=
\sg(\bar q_{t+1})
-
\widehat q_{t+1}
=
\sg(\bar q_{t+1})
-
q_tA_t.
\label{eq:cat-residual}
\end{equation}
This vector measures the discrepancy between the target-encoder distribution and the transition-based prediction for each latent category. The stop-gradient operator ensures that, when auxiliary diagnostic models are fitted to these residuals, their gradients are not propagated into the target encoder.

To distinguish transition fitting from state sufficiency, define the current predictor information
\begin{equation*}
\mathcal G_t
=
\sigma(q_t,\xi_t)
\end{equation*}
and the full observed-history filtration
\begin{equation*}
\mathcal F_t
=
\sigma(X_{\leq t},\xi_{\leq t}).
\end{equation*}
For fixed model parameters, $\mathcal G_t\subseteq\mathcal F_t$, since $q_t$ is computed from the observation history.
The transition predictor is conditionally mean-correct with respect to its own inputs when
\begin{equation*}
q_tA_t
=
\E\left[
\sg(\bar q_{t+1})
\mid
\mathcal G_t
\right].
\end{equation*}
This property is naturally associated with the forward-KL objective used by MCJEPA.\footnote{Let $Y=\sg(\bar q_{t+1})\in\DeltaK$ and condition on $\mathcal G_t$. For any predicted distribution $p\in\DeltaK$,
\[
\E[\KL(Y\|p)\mid\mathcal G_t]
=
\E\!\left[
\sum_j Y_j\log Y_j
\middle|
\mathcal G_t
\right]
-
\sum_j
\E[Y_j\mid\mathcal G_t]\log p_j.
\]
The first term is independent of $p$, so minimizing the conditional expected KL is equivalent to minimizing the cross-entropy with
$m_t=\E[Y\mid\mathcal G_t]$. The population optimum is therefore $p^\star=m_t$. Hence, when the transition family can represent the optimum of the MCJEPA objective, $q_tA_t=\E[\sg(\bar q_{t+1})\mid\mathcal G_t]$. A restricted or imperfectly optimized transition family need not satisfy this equality exactly.}
Under this condition,
\begin{equation*}
\E[R_{t+1}\mid\mathcal G_t]
=
0.
\end{equation*}
Thus, after conditioning on the inputs already available to the transition predictor, the residual has no remaining predictable conditional mean.

Markov sufficiency requires a stronger invariance: older history should not alter the conditional prediction once the current predictive state and side information are known. At the level of the target-encoder distribution, the corresponding conditional-mean implication is
\begin{equation*}
\E\left[
\sg(\bar q_{t+1})
\mid
\mathcal F_t
\right]
=
\E\left[
\sg(\bar q_{t+1})
\mid
\mathcal G_t
\right].
\end{equation*}
Combining this condition with a conditionally mean-correct transition gives
\begin{equation}
\E[R_{t+1}\mid\mathcal F_t]
=
0.
\label{eq:mds}
\end{equation}
Assuming integrability, \cref{eq:mds} gives the martingale-difference property of $\{R_{t+1}\}$ with respect to $\{\mathcal F_t\}$: once the current state and the available history are known, the residual has no systematic predictable component.

A consequence of \cref{eq:mds} is that $R_{t+1}$ is uncorrelated with any square-integrable function measurable with respect to $\mathcal F_t$. A simple linear diagnostic is therefore
\begin{equation*}
\mathcal D_{\mathrm{lin}}
=
\sum_{k=1}^{K_{\mathrm r}}
\left\|
\operatorname{Cov}
\left(
R_{t+1},
q_{t-k}
\right)
\right\|_F^2,
\end{equation*}
where $K_{\mathrm r}$ is the maximum lag examined and $\|\cdot\|_F$ denotes the Frobenius norm. A large value indicates that some components of the residual remain linearly associated with earlier latent states. A value near zero rules out only this particular form of linear dependence and does not establish Markov sufficiency.

\subsection{Testing incremental predictability from older history}

The covariance diagnostic detects only linear dependence. A stronger test asks whether an auxiliary model can predict the residual from older history beyond what can already be predicted from the current state and side information. Consider a restricted residual predictor
\begin{equation*}
\widehat R_{t+1}^{(0)}
=
g_{\omega_0}(q_t,\xi_t)
\end{equation*}
and a history-augmented predictor
\begin{equation*}
\widehat R_{t+1}^{(1)}
=
g_{\omega_1}
\left(
q_t,\xi_t,
q_{t-1},\ldots,q_{t-K_{\mathrm r}}
\right).
\end{equation*}
Their held-out prediction errors can be compared through
\begin{equation}
\begin{aligned}
\Delta_{\mathrm{hist}}
={}
\widehat{\E}
\left[
\left\|
R_{t+1}
-
\widehat R_{t+1}^{(0)}
\right\|_2^2
\right]
-
\widehat{\E}
\left[
\left\|
R_{t+1}
-
\widehat R_{t+1}^{(1)}
\right\|_2^2
\right].
\end{aligned}
\label{eq:residual-history-gain}
\end{equation}
A reliably positive value of $\Delta_{\mathrm{hist}}$ means that older latent history improves prediction beyond $(q_t,\xi_t)$. This provides evidence against the conditional-mean sufficiency of the current state--predictor pair\footnote{The comparison should be evaluated on held-out data or through cross-fitting. The restricted and augmented auxiliary predictors should also have comparable capacity and regularization; otherwise an apparent history gain may reflect unequal model flexibility rather than genuinely additional predictive information.}.

The same principle can be implemented by comparing restricted and history-augmented predictors of the target itself rather than predictors of the residual. If a common baseline prediction is used, the two formulations are equivalent because predicting
$R_{t+1}=Y_{t+1}-\widehat Y_{t+1}$
is equivalent to correcting the baseline prediction $\widehat Y_{t+1}$. Experiment~3 uses the direct-prediction version of this diagnostic, comparing prediction from $Z_t$ with prediction from $(Z_t,Z_{t-1})$.

\subsection{Distinguishing state insufficiency from predictor misspecification}

Residual predictability can arise because of either the representation or the transition model. First, the current representation may fail to summarize all past information relevant to predicting the future. Second, the transition model may be too restricted or insufficiently optimized even when the representation itself is sufficient.

The distinction between $\mathcal G_t$ and $\mathcal F_t$ helps separate these effects. Predictability of $R_{t+1}$ from $(q_t,\xi_t)$ alone indicates that the fitted transition has not captured the conditional mean available from its own inputs. The restricted auxiliary model $g_{\omega_0}$ can absorb part of this current-input misspecification. Additional held-out improvement after introducing $(q_{t-1},\ldots,q_{t-K_{\mathrm r}})$ then asks a more specific question: does older latent history contain predictive information not recoverable from the current state and side information?

Attributing a positive $\Delta_{\mathrm{hist}}$ specifically to representation insufficiency nevertheless requires care. The current-input predictor and residual correction must be sufficiently expressive and well fitted, the restricted and augmented diagnostic models should be compared under matched capacity and regularization, and evaluation should be performed out of sample. Under these conditions, incremental predictability from older history is evidence that the current representation has omitted transition-relevant information rather than merely that the original transition parameterization was imperfect.

\subsection{Continuous-state diagnostics}

For a continuous probabilistic predictor, suppose that
\begin{equation*}
p_\phi(Z_{t+1}\mid Z_t,\xi_t)
=
\mathcal N(\mu_t,\Sigma_t),
\end{equation*}
and let
\begin{equation*}
Z_{t+1}^{\mathrm T}
\sim
q_{\bar\theta}(Z_{t+1}\mid X_{t+1})
\end{equation*}
denote a target-encoder latent sample. When $\Sigma_t$ is positive definite, a standardized innovation may be defined as
\begin{equation*}
R_{t+1}^{\mathrm{std}}
=
\Sigma_t^{-1/2}
\left(
\sg(Z_{t+1}^{\mathrm T})
-
\mu_t
\right).
\end{equation*}
For singular or nearly singular covariance matrices, a regularized or pseudoinverse square root should be used instead.

Under a correctly specified conditional Gaussian model, the standardized innovation has conditional mean zero and conditional covariance equal to the identity. Markov sufficiency further implies that older history should not systematically predict this innovation once the current state and side information are given. One may therefore examine lagged residual dependence, residual covariance, squared-residual dependence, or auxiliary history-prediction gains. For non-Gaussian predictors, score-based diagnostics, probability-integral-transform diagnostics in suitable scalar settings, or appropriate multivariate calibration diagnostics can be used instead.

These diagnostics test different consequences of correct conditional prediction. Zero autocorrelation, for example, rules out linear temporal dependence but does not exclude nonlinear or higher-order dependence. Likewise, calibrated marginal probability-integral-transform values do not establish the conditional independence required for Markov sufficiency.

Residual analysis should therefore be interpreted asymmetrically. Predictability from older history provides evidence against Markov sufficiency of the current state--predictor pair. Failure to detect such predictability means only that the chosen diagnostics do not reject sufficiency; it does not prove that the learned representation is Markov sufficient.

\section{Experiments}
\label{sec:experiments}

Our experiments are deliberately small and diagnostic. Rather than targeting state-of-the-art forecasting performance, we test the structural claims developed in the preceding sections in settings where the latent states, transition laws, filtering distributions, and predictive sufficiency structure are known. This allows us to separate the effect of the proposed Markov structure from model capacity and uncontrolled properties of real-world data.

\Cref{tab:experiment-overview} summarizes the 4 experiments. They follow the conceptual progression of the paper: Experiment~1 asks whether MCJEPA recovers a coherent finite-state transition; Experiment~2 isolates the distinction between local observation evidence and filtering; Experiment~3 tests predictive compression, Markovization, and residual sufficiency; and Experiment~4 studies the strongest correspondence by training the same latent-state architecture with an HMM sequence objective. Full data-generation parameters, architectures, optimization settings, and hyperparameters are deferred to Appendix~\ref{app:experimental-details}.

\begin{table}[t]
\centering
\small
\renewcommand{\arraystretch}{1.08}
\caption{Overview of the experimental questions. All experiments use controlled synthetic processes for which the relevant latent structure is known.}
\label{tab:experiment-overview}
\begin{tabular}{@{}p{0.08\linewidth}p{0.34\linewidth}p{0.49\linewidth}@{}}
\toprule
Exp. & Structural question & Main evidence \\
\midrule
1
&
Does a shared Markov transition recover coherent finite-state dynamics?
&
State and transition recovery, multi-horizon prediction, path consistency, collapse ablations
\\
2
&
When is history-based filtering necessary?
&
Local versus filtered state accuracy, NLL, and posterior trajectories under emission ambiguity
\\
3
&
Can predictive compression construct a compact sufficient state?
&
Exact information trade-off, deterministic-partition frontier, learned compression path, residual predictability
\\
4
&
Can the same architecture be trained by an HMM sequence objective?
&
Sequence NLL, transition recovery, latent-state recovery, and filtering agreement
\\
\bottomrule
\end{tabular}
\end{table}

\paragraph{Common protocol.}
All reported learned-model results use five random seeds, with means and standard deviations reported across seeds. Categorical latent-state labels are identifiable only up to permutation. ARI and NMI are themselves permutation invariant, whereas transition matrices and predicted categorical distributions are aligned to ground-truth state order using a Hungarian assignment computed from the training-set hard state assignments. Multi-step evaluation uses horizons
\begin{equation*}
\mathcal H=\{1,2,4,8\}.
\end{equation*}

We use several common metrics to distinguish state recovery, transition recovery, predictive accuracy, and observation-level probabilistic fit. Full definitions are collected in Appendix~\ref{app:experimental-metrics}.

\emph{Adjusted Rand index} (ARI; Eq.~\ref{eq:ARI}) and \emph{normalized mutual information} (NMI; Eq.~\ref{eq:NMI}) measure agreement between inferred and ground-truth state partitions. Larger values indicate better recovery, with $1$ corresponding to exact partition agreement. For reference,
\begin{equation}
\operatorname{NMI}(S,\widehat S)
=
\frac{
2I(S;\widehat S)
}{
H(S)+H(\widehat S)
}.
\tag{Eq.~\ref{eq:NMI}}
\end{equation}
ARI additionally corrects pairwise partition agreement for agreement expected by chance; its full expression is given in Eq.~\ref{eq:ARI}.

Transition recovery is measured by the normalized permutation-aligned Frobenius error
\begin{equation}
\mathcal E_A
=
\frac{
\left\|
\widehat A_{\mathrm{aligned}}-A^\star
\right\|_F
}{
\|A^\star\|_F
}
=
\frac{
\left\|
P^\top\widehat A P-A^\star
\right\|_F
}{
\|A^\star\|_F
}.
\tag{Eq.~\ref{eq:transition_matrix_recovery}}
\end{equation}
Here, $\widehat A$ is the learned transition matrix, $A^\star$ is the ground-truth transition matrix, and $P$ is the learned-to-ground-truth permutation matrix obtained from training-set state alignment. Lower $\mathcal E_A$ indicates more faithful recovery of the latent dynamics.

Predictive quality is measured using negative log-likelihood. When the ground-truth future state is available, the horizon-$h$ true-state NLL is
\begin{equation}
\mathcal L_h^{\mathrm{state}}
=
-\E_t
\left[
\log
\widehat q_{t+h}(S_{t+h})
\right],
\tag{Eq.~\ref{eq:state_prediction_nll}}
\end{equation}
where $S_{t+h}$ is the ground-truth latent state and $\widehat q_{t+h}$ is the predicted categorical distribution after state alignment. Thus, $\widehat q_{t+h}(S_{t+h})$ is the probability assigned to the realized future state, and lower NLL indicates better probabilistic prediction.

When an explicit transition--emission model is available, observation-space fit is measured by sequence NLL per time step,
\begin{equation}
\mathcal L_{\mathrm{seq}}
=
-\frac{1}{T}
\E
\left[
\log p_{\phi,\psi}(X_{1:T})
\right].
\tag{Eq.~\ref{eq:sequence_nll}}
\end{equation}
Here, $p_{\phi,\psi}(X_{1:T})$ is the marginal observation-sequence density obtained after marginalizing the latent-state trajectory under the transition and emission models. This differs from true-state NLL: $\mathcal L_h^{\mathrm{state}}$ evaluates prediction of the known synthetic latent state, whereas $\mathcal L_{\mathrm{seq}}$ evaluates the probability density assigned to the observed sequence under the complete probabilistic model.

Experiment~1 additionally evaluates multi-horizon structural consistency through the path-disagreement metric
\begin{equation}
\mathcal D_{\mathrm{path}}(h_1,h_2)
=
\E_t
\left[
\left\|
q_tA_{h_1+h_2}
-
(q_tA_{h_1})A_{h_2}
\right\|_1
\right].
\tag{Eq.~\ref{eq:path_disagreement}}
\end{equation}
For MCJEPA with a single shared transition matrix $A_h=A^h$, this quantity is identically zero by construction.

Experiment~2 additionally reports state accuracy,
\begin{equation}
\operatorname{Acc}
=
\frac{1}{N}
\sum_t
\mathbf{1}
\left\{
\arg\max_k q_t(k)=S_t
\right\},
\tag{Eq.~\ref{eq:accuracy}}
\end{equation}
the multiclass Brier score,
\begin{equation}
\operatorname{Brier}
=
\frac{1}{N}
\sum_t
\sum_k
\left(
q_t(k)-\mathbf{1}\{S_t=k\}
\right)^2,
\tag{Eq.~\ref{eq:brier_score}}
\end{equation}
and mean posterior entropy,
\begin{equation}
\overline H
=
\frac{1}{N}
\sum_t H(q_t).
\tag{Eq.~\ref{eq:mean_posterior_entropy}}
\end{equation}
Accuracy evaluates hard state recovery, while NLL and Brier score retain information about probabilistic confidence. Posterior entropy is descriptive rather than a stand-alone performance criterion.

Finally, Experiment~4 measures agreement between an exact model-based filter and the amortized context encoder through
\begin{equation}
\mathcal D_{\mathrm{filter}}
=
\E_t
\left[
\KL
\left(
q_t^{\mathrm{exact}}
\,\middle\|\,
q_t^{\mathrm{enc}}
\right)
\right].
\tag{Eq.~\ref{eq:filtering_kl}}
\end{equation}
Filtering KL is training aligned for the HMM+filter and hybrid regimes because both explicitly optimize filtering distillation, so we interpret it primarily as a diagnostic of whether the context encoder has acquired the intended filtering role.

Other experiment-specific quantities, including effective state count, assignment entropy for collapse analysis, mutual-information quantities in the predictive bottleneck, and residual history gain $\Delta_{\mathrm{hist}}$, are defined where they are first introduced. The systems and networks are intentionally small because the aim is structural diagnosis rather than scaling. Detailed data-generation procedures, architectures, optimization settings, metric definitions, and supplementary results are provided in Appendix~\ref{app:experimental-details}.

\subsection{Experiment 1: finite-HMM recovery and Markov composition}
\label{sec:exp-finite-hmm}

We first test the basic MCJEPA construction on a finite HMM with four latent states and continuous observations. We consider a separated-emission regime, in which observations are highly informative about the state, and an ambiguous-emission regime, in which state inference becomes more difficult. MCJEPA uses a categorical history encoder and one shared row-stochastic transition matrix $A$, yielding
\begin{equation*}
\widehat q_{t+h}=q_tA^h.
\end{equation*}
We compare it with a horizon-specific categorical JEPA that learns an independent $A_h$ for each prediction horizon, and with a correctly specified Gaussian HMM.

\paragraph{Recovering states and transitions.}
\Cref{tab:exp1-recovery} reports ARI, aligned transition error $\mathcal E_A$, and true-state prediction NLL at multiple horizons.
In the separated regime, all three methods recover the latent process well. MCJEPA attains
\begin{equation*}
\mathrm{ARI}=0.9796\pm0.0055
\end{equation*}
and transition error
\begin{equation*}
\mathcal E_A=0.0191\pm0.0019,
\end{equation*}
close to the correctly specified HMM.

The ambiguous regime exposes the structural trade-off more clearly. The HMM remains strongest because it explicitly models the correct emission family and performs probabilistic filtering. The horizon-specific predictor obtains somewhat better state recovery than MCJEPA, but its aligned one-step transition error is
\begin{equation*}
0.2410\pm0.0029,
\end{equation*}
compared with MCJEPA's
\begin{equation*}
0.0901\pm0.0091.
\end{equation*}
Thus, independently fitting each horizon provides additional predictive flexibility but yields a substantially less faithful underlying transition law.

\begin{table}[t]
\centering
\small
\renewcommand{\arraystretch}{1.08}
\caption{Finite-HMM recovery. Values are mean $\pm$ standard deviation over 5 seeds. $\mathcal E_A$ is the permutation-aligned transition-matrix error; $\mathcal L_1$ and $\mathcal L_8$ are true-state prediction NLL at horizons $1$ and $8$.}
\label{tab:exp1-recovery}
\resizebox{\linewidth}{!}{
\begin{tabular}{llcccc}
\toprule
Regime & Model & ARI $\uparrow$ & $\mathcal E_A\downarrow$ & $\mathcal L_1\downarrow$ & $\mathcal L_8\downarrow$ \\
\midrule
Separated
&
Gaussian HMM
&
$\mathbf{\underline{0.9947\pm0.0006}}$
&
$\mathbf{\underline{0.0082\pm0.0020}}$
&
$\mathbf{\underline{0.5240\pm0.0026}}$
&
$\mathbf{\underline{1.3027\pm0.0050}}$
\\
&
Horizon-specific $A_h$
&
$0.9836\pm0.0034$
&
$0.0231\pm0.0018$
&
$0.5433\pm0.0019$
&
$1.3036\pm0.0058$
\\
&
MCJEPA shared $A$
&
$0.9796\pm0.0055$
&
$0.0191\pm0.0019$
&
$0.5438\pm0.0018$
&
$1.3048\pm0.0049$
\\
\midrule
Ambiguous
&
Gaussian HMM
&
$\mathbf{\underline{0.7543\pm0.0115}}$
&
$\mathbf{\underline{0.0091\pm0.0027}}$
&
$\mathbf{\underline{0.6676\pm0.0103}}$
&
$\mathbf{\underline{1.3127\pm0.0060}}$
\\
&
Horizon-specific $A_h$
&
$0.5581\pm0.0476$
&
$0.2410\pm0.0029$
&
$0.8488\pm0.0306$
&
$1.3288\pm0.0041$
\\
&
MCJEPA shared $A$
&
$0.5323\pm0.0513$
&
$0.0901\pm0.0091$
&
$0.8500\pm0.0349$
&
$1.3491\pm0.0093$
\\
\bottomrule
\end{tabular}
}
\end{table}

Importantly, these results do not imply that the shared-matrix constraint universally minimizes predictive NLL. Under ambiguity, the independently parameterized $A_h$ model is slightly better at several horizons. The benefit of MCJEPA is instead structural: all horizons are generated by one transition mechanism and must therefore compose consistently.

\begin{figure}[t]
\centering
\begin{minipage}[t]{0.48\linewidth}
\centering
\includegraphics[width=\linewidth]{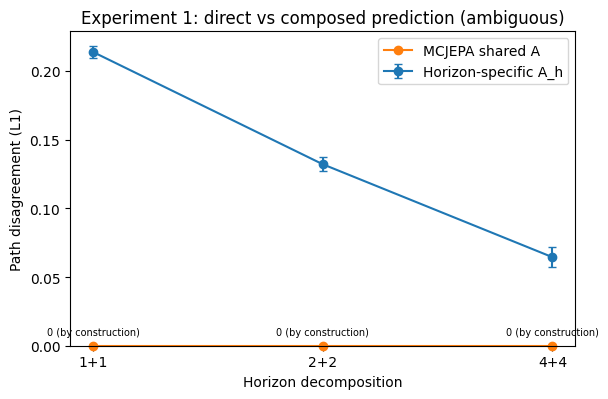}
\end{minipage}
\hfill
\begin{minipage}[t]{0.5\linewidth}
\centering
\includegraphics[width=\linewidth]{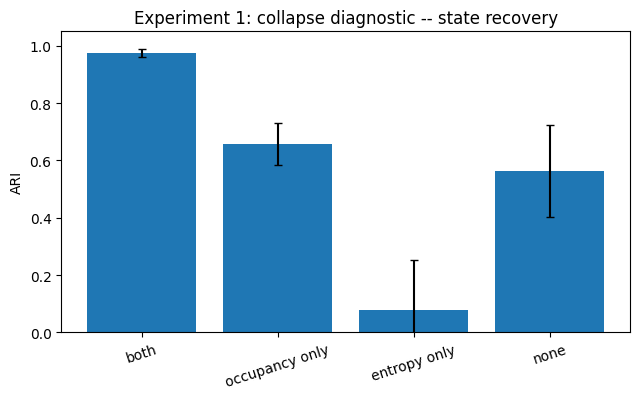}
\end{minipage}
\caption{Structural diagnostics for Experiment~1. \emph{Left:} direct-versus-composed prediction disagreement $\mathcal D_{\mathrm{path}}$ from \cref{eq:empirical-path-disagreement} in the ambiguous-emission regime. The labels $1+1$, $2+2$, and $4+4$ compare direct predictions at total horizons $2$, $4$, and $8$ with predictions obtained by composing two successive predictions of horizons $1$, $2$, and $4$, respectively. MCJEPA is exactly path-consistent because all horizons are generated by powers of one shared transition matrix; independently learned $A_h$ are not. \emph{Right:} visualization of the ARI column of \cref{tab:exp1-collapse} under the four state-regularization ablations. Occupancy and entropy regularization are complementary, and using both gives the strongest state recovery. Error bars denote mean $\pm$ one standard deviation over 5 seeds.}
\label{fig:exp1-structural}
\end{figure}

\paragraph{Exact path composition.}
The left panel of \cref{fig:exp1-structural} illustrates the consequence of using a shared Markov transition. To quantify disagreement between a direct prediction and a composed prediction with the same total horizon, we define
\begin{equation}
\mathcal D_{\mathrm{path}}(h_1,h_2)
=
\E_t\left[
\left\|
q_tA_{h_1+h_2}
-
(q_tA_{h_1})A_{h_2}
\right\|_1
\right],
\label{eq:empirical-path-disagreement}
\end{equation}
where the expectation is taken over valid evaluation time points. Thus, $\mathcal D_{\mathrm{path}}=0$ means that the direct and composed predicted state distributions agree exactly.

The labels $1+1$, $2+2$, and $4+4$ denote three decompositions of the same total prediction horizon. Specifically, $1+1$ compares a direct two-step prediction with two successive one-step predictions,
\begin{equation*}
q_tA_2
\quad\text{versus}\quad
(q_tA_1)A_1,
\end{equation*}
$2+2$ compares a direct four-step prediction with two successive two-step predictions,
\begin{equation*}
q_tA_4
\quad\text{versus}\quad
(q_tA_2)A_2,
\end{equation*}
and $4+4$ compares a direct eight-step prediction with two successive four-step predictions,
\begin{equation*}
q_tA_8
\quad\text{versus}\quad
(q_tA_4)A_4.
\end{equation*}

For MCJEPA, $A_h=A^h$, so
\begin{equation*}
q_tA^{h_1+h_2}
=
(q_tA^{h_1})A^{h_2}
\end{equation*}
exactly, and hence $\mathcal D_{\mathrm{path}}=0$ by construction. Independently learned horizon-specific matrices $A_h$, however, are not constrained to satisfy these composition identities. In the ambiguous regime, their path disagreement is
\begin{equation*}
0.2138\pm0.0041,\qquad
0.1321\pm0.0053,\qquad
0.0645\pm0.0072
\end{equation*}
for the $1+1$, $2+2$, and $4+4$ decompositions, respectively. The corresponding separated-regime disagreement is smaller but remains nonzero, confirming that independently trained horizon predictors need not define one coherent Markov chain.

\paragraph{Preventing discrete-state collapse.}
We next ablate the occupancy and entropy components of the state-use regularizer $\mathcal L_{\mathrm{state}}$ in \cref{eq:state-reg}. \Cref{tab:exp1-collapse} shows that the two terms address complementary failure modes. Using both gives
\begin{equation*}
\mathrm{ARI}=0.9748\pm0.0134
\end{equation*}
and approximately four effective hard states. Occupancy regularization alone maintains broad state use but leaves assignments highly uncertain, with mean assignment entropy
\begin{equation*}
1.1037\pm0.1216.
\end{equation*}
Entropy regularization alone instead makes assignments confident but usually collapses them onto a single state: four of the five runs use one effective state, and the average effective hard-state count is only
\begin{equation*}
1.1703\pm0.3807.
\end{equation*}
Using neither term produces less severe but unstable state use and substantially weaker recovery.

\begin{table}[t]
\centering
\small
\renewcommand{\arraystretch}{1.08}
\caption{Collapse ablation in Experiment~1. Values are mean $\pm$ standard deviation over 5 seeds. $K_{\mathrm{eff}}^{\mathrm{hard}}$ is the effective number of states induced by hard assignments. Assignment entropy measures per-example confidence and should be interpreted jointly with state usage: very low entropy can also arise from single-state collapse. The occupancy and entropy terms are complementary: the former encourages global state use, while the latter encourages confident per-example assignments.}
\label{tab:exp1-collapse}
\begin{tabular}{lccc}
\toprule
Regularization
&
ARI $\uparrow$
&
$K_{\mathrm{eff}}^{\mathrm{hard}}$
&
Assignment entropy
\\
\midrule
Both
&
$\mathbf{0.9748\pm0.0134}$
&
$3.9551\pm0.0084$
&
$0.0716\pm0.0987$
\\
Occupancy only
&
$0.6582\pm0.0736$
&
$3.8092\pm0.2249$
&
$1.1037\pm0.1216$
\\
Entropy only
&
$0.0776\pm0.1734$
&
$1.1703\pm0.3807$
&
$0.0015\pm0.0034$
\\
None
&
$0.5626\pm0.1612$
&
$3.4007\pm0.5026$
&
$1.1062\pm0.0837$
\\
\bottomrule
\end{tabular}
\end{table}

The right panel of \cref{fig:exp1-structural} visualizes the ARI column of \cref{tab:exp1-collapse}. Together, the state-recovery and collapse diagnostics support our intended interpretation: occupancy prevents global state under-use, whereas the entropy term prevents diffuse per-sample assignments; both are needed to obtain confident and diverse state assignments without collapse.

The complete multi-horizon curves, the separated-regime path-consistency result, and the state-usage and assignment-confidence ablations are provided in Appendix~\ref{app:experimental-details}.

\subsection{Experiment 2: filtering resolves emission ambiguity}
\label{sec:exp-filtering}

Experiment~2 isolates the distinction between local observation evidence and filtering. We use a persistent two-state HMM whose emission distributions are made progressively more overlapping\footnote{The two states have Gaussian emissions centered at $-\mu$ and $+\mu$ with common standard deviation $\sigma$. We control emission ambiguity through the separation ratio $\mu/\sigma\in\{2.0,1.25,0.75,0.45\}$; decreasing $\mu/\sigma$ increases the overlap between the two emission distributions and therefore makes the current observation less informative about the latent state.}. 
Because the generating model is known, we can compute both
\begin{equation*}
\text{local evidence: } p(S_t\mid X_t)
\end{equation*}
and
\begin{equation*}
\text{filtering: } p(S_t\mid X_{\leq t})
\end{equation*}
exactly. In MCJEPA terms, these two quantities are the oracle counterparts of a local encoder $q_\theta(Z_t\mid X_t)$ and a history-dependent encoder $q_\theta(Z_t\mid X_{\leq t})$, respectively. We use the exact posteriors here rather than learned encoders so that the experiment isolates the informational value of observation history without representation-learning or optimization confounds; it is therefore not a comparison between an HMM and MCJEPA.

\begin{figure}[t]
\centering
\begin{minipage}[t]{0.44\linewidth}
\centering
\includegraphics[width=\linewidth]{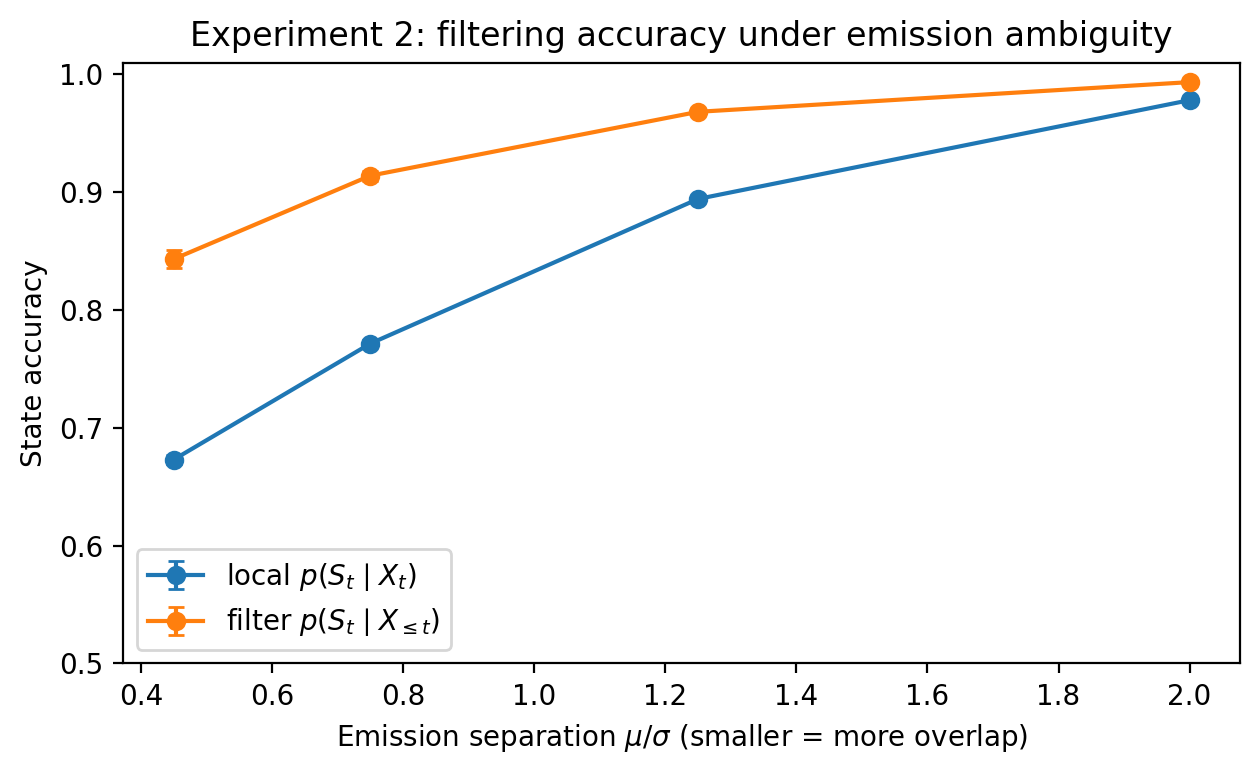}
\end{minipage}
\hfill
\begin{minipage}[t]{0.54\linewidth}
\centering
\includegraphics[width=\linewidth]{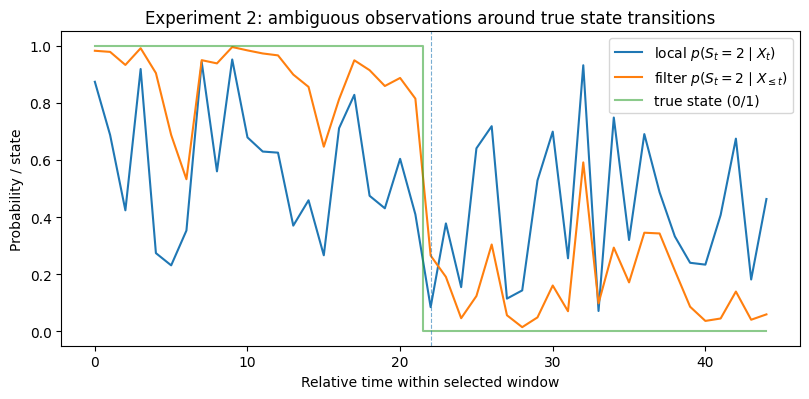}
\end{minipage}
\caption{Filtering under emission ambiguity. \emph{Left:} comparison of the exact local-evidence and filtering distributions associated with the local and history-dependent encoder roles in MCJEPA. Filtering increasingly outperforms local evidence as emission overlap grows. Error bars denote mean $\pm$ one standard deviation over 5 seeds. \emph{Right:} a representative ambiguous sequence around a true state transition. The local posterior reacts strongly to individual noisy observations, whereas the filtered belief integrates temporal evidence through the transition model. These are oracle posteriors computed from the known generating process, rather than separately trained HMM and MCJEPA models.}
\label{fig:exp2-filtering}
\end{figure}

\Cref{fig:exp2-filtering} shows that the oracle filtering distribution becomes increasingly more informative than the oracle local-evidence distribution as individual observations become ambiguous. At the most overlapping setting, local evidence reaches state accuracy
\begin{equation*}
0.6730\pm0.0037,
\end{equation*}
whereas filtering reaches
\begin{equation*}
0.8433\pm0.0080.
\end{equation*}
The corresponding state NLL decreases from
\begin{equation*}
0.6022\pm0.0030
\end{equation*}
to
\begin{equation*}
0.3683\pm0.0088.
\end{equation*}
The advantage diminishes as the emissions become locally separable, as expected.

The right panel of \cref{fig:exp2-filtering} illustrates the mechanism around a true state transition. Local evidence fluctuates strongly with individual observations. Filtering instead combines the current observation with the propagated state belief, remaining stable through many locally ambiguous measurements and changing when the accumulated evidence supports a transition. This directly supports the probabilistic distinction made earlier in the paper: $p(S_t\mid X_t)$ is local observation evidence, whereas $p(S_t\mid X_{\leq t})$ is the HMM filtering belief.

The corresponding NLL curve is provided in Appendix~\ref{app:experimental-details}.

\subsection{Experiment 3: predictive compression and Markovization}
\label{sec:exp-markovization}

Experiment~3 asks a simple question: can predictive compression discard unnecessary history while retaining exactly the information needed to predict the future?

We construct a binary second-order process satisfying
\begin{equation}
p(X_{t+1}\mid X_{\leq t})
=
p(X_{t+1}\mid X_{t-1},X_t).
\label{eq:exp3-second-order}
\end{equation}
Thus, although the entire observation history is available, only the two most recent observations are needed to predict $X_{t+1}$.

We deliberately give the encoder a longer three-step history,
\begin{equation}
H_t=(X_{t-2},X_{t-1},X_t),
\end{equation}
which has eight possible values. The known minimal predictive state is
\begin{equation}
Z_t^\star=(X_{t-1},X_t),
\end{equation}
which has only four possible values. The older bit $X_{t-2}$ is therefore redundant once $Z_t^\star$ is known.

This construction gives us a controlled ground truth for what predictive compression should do:
\begin{equation*}
\underbrace{(X_{t-2},X_{t-1},X_t)}_{\text{too much history}}
\;\longrightarrow\;
\underbrace{(X_{t-1},X_t)}_{\text{just enough}}
\;\longrightarrow\;
\underbrace{X_t}_{\text{too little}}.
\end{equation*}
In MCJEPA terms, these are three controlled choices of latent state supplied to the predictor: an overcomplete state $Z_t=H_t$, the minimal sufficient state $Z_t=Z_t^\star$, and an insufficient state $Z_t=X_t$.

The middle representation also explains the term \emph{Markovization}. Although the observation process is second-order in $X_t$, defining
\begin{equation*}
Z_t^\star=(X_{t-1},X_t)
\end{equation*}
turns it into a first-order state process: the information needed for the next transition is contained in the current state $Z_t^\star$, without requiring older history. Predictive compression should therefore remove $X_{t-2}$, but should not remove $X_{t-1}$.

\paragraph{Does compression recover the correct predictive state?}
We first compare the three controlled representations exactly. Because the process is finite, their information quantities and optimal one-step prediction losses can be computed without representation-learning or optimization error.

\Cref{tab:exp3-controls} gives the key result. The full three-bit history retains
\begin{equation*}
\MI(H_t;Z_t)=1.7356,
\end{equation*}
whereas the four-state predictive pair retains only
\begin{equation*}
\MI(H_t;Z_t)=1.3378.
\end{equation*}
Despite this compression, the two representations contain exactly the same information about the next observation,
\begin{equation*}
\MI(Z_t;X_{t+1})=0.2807,
\end{equation*}
and achieve the same prediction NLL,
\begin{equation*}
0.3978.
\end{equation*}
Hence, removing $X_{t-2}$ reduces the amount of past information stored in the state without sacrificing prediction.

Compressing further to $Z_t=X_t$, however, removes information that is genuinely needed. Predictive information falls from $0.2807$ to $0.0192$, and prediction NLL increases from $0.3978$ to $0.6593$. The controlled construction therefore has a known sufficiency--minimality boundary: eight states are predictively sufficient but redundant, four states are sufficient and minimal for this process, and two states are insufficient.

\begin{table}[t]
\centering
\small
\renewcommand{\arraystretch}{1.08}
\caption{Exact representation controls for Experiment~3. The minimal predictive pair removes redundant history while preserving all one-step predictive information. Compressing further to the current observation alone loses information required for prediction.}
\label{tab:exp3-controls}
\begin{tabular}{lcccc}
\toprule
Representation
&
$\MI(H_t;Z_t)$
&
$\MI(Z_t;X_{t+1})$
&
NLL
&
States
\\
\midrule
Full history $H_t$
&
1.7356
&
0.2807
&
0.3978
&
8
\\
Minimal predictive pair
&
1.3378
&
0.2807
&
0.3978
&
4
\\
Current observation $X_t$
&
0.6785
&
0.0192
&
0.6593
&
2
\\
\bottomrule
\end{tabular}
\end{table}

\paragraph{Is the four-state solution truly optimal, or just a favorable example?}
The comparison above considers only three hand-specified representations. We therefore use the small history space to perform an exhaustive check over \emph{every deterministic compression} of the eight possible histories represented by $H_t$.

A deterministic encoder
\begin{equation*}
Z_t=f(H_t)
\end{equation*}
groups histories that are assigned to the same latent state. We enumerate all such groupings and evaluate each one using
\begin{equation}
\mathcal L_{\mathrm{PIB}}^{\mathrm{exp}}
=
\mathcal L_{\mathrm{pred}}
+
\beta\,\MI(H_t;Z_t),
\label{eq:exp-pib-surrogate}
\end{equation}
where the first term rewards accurate prediction and the second penalizes retaining unnecessary information about the history.

Because there are only eight possible histories, all $4140$ deterministic partitions can be enumerated exactly. This provides a global deterministic reference rather than relying on a few hand-designed candidates. For every tested positive compression weight
\begin{equation*}
0<\beta\leq0.014,
\end{equation*}
the globally optimal deterministic representation is exactly the known four-state predictive state
\begin{equation*}
Z_t^\star=(X_{t-1},X_t).
\end{equation*}
Thus, when compression is strong enough to penalize redundant history but not so strong that predictive information is sacrificed, the predictive-bottleneck objective selects the known minimal sufficient Markov state.

At $\beta=0$, several predictively equivalent deterministic representations attain the same minimum prediction loss; the four-state state is therefore not identified by prediction alone. Once $\beta>0$, however, redundant stored history is penalized. At $\beta=0.015$, the deterministic optimum changes to a two-state representation. Its retained predictive information decreases from $0.2807$ to $0.2711$, indicating that compression has begun to remove information useful for prediction. With still stronger compression, the optimum eventually collapses to a single state. The resulting progression is therefore
\begin{equation*}
\text{redundant representation}
\;\longrightarrow\;
\text{minimal predictive state}
\;\longrightarrow\;
\text{over-compressed state}.
\end{equation*}

\begin{figure}[t]
\centering
\begin{minipage}[t]{0.5\linewidth}
\centering
\includegraphics[width=\linewidth]{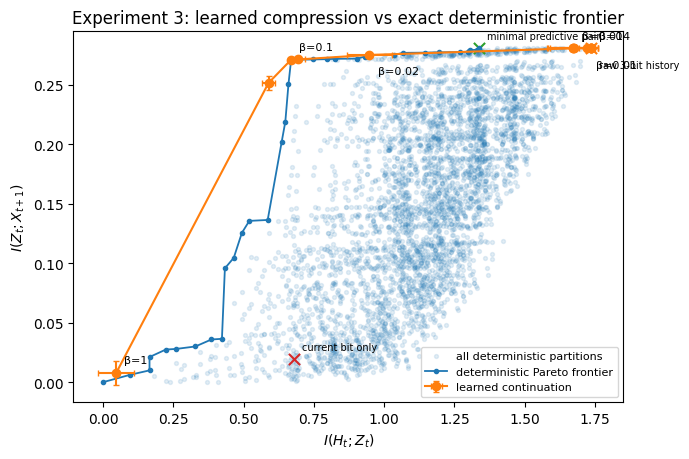}
\end{minipage}
\hfill
\begin{minipage}[t]{0.48\linewidth}
\centering
\includegraphics[width=\linewidth]{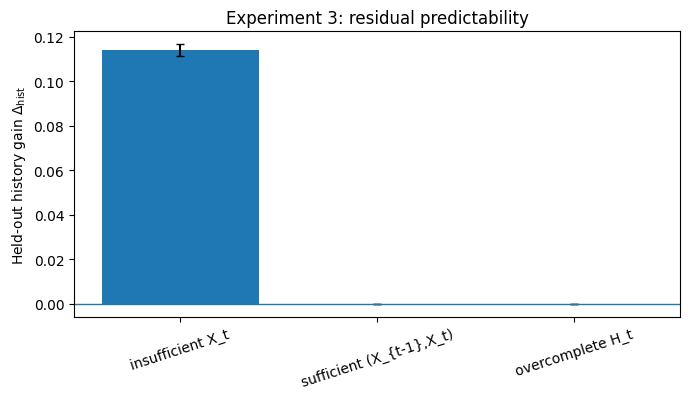}
\end{minipage}
\caption{Predictive compression and Markovization. \emph{Left:} exhaustive deterministic compression provides a global reference for the prediction--compression trade-off, while the learned stochastic encoder traces a warm-started continuation path as the compression weight $\beta$ is increased. Moderate compression can remove redundant history while preserving predictive information, whereas excessive compression eventually sacrifices information required for prediction. \emph{Right:} residual history gain $\Delta_{\mathrm{hist}}$ diagnoses predictive insufficiency by testing whether adding the previous representation state $Z_{t-1}$ improves held-out prediction beyond using $Z_t$ alone. This augmentation substantially helps the insufficient representation $Z_t=X_t$, for which $Z_{t-1}=X_{t-1}$ restores omitted transition-relevant information, but provides essentially no additional predictive benefit for either the minimal sufficient or the overcomplete representation. Error bars denote mean $\pm$ one standard deviation over 5 seeds where applicable.}
\label{fig:exp3-markovization}
\end{figure}

The left panel of \cref{fig:exp3-markovization} visualizes the compression--prediction trade-off in two complementary ways. Each light-blue point corresponds to one of the $4140$ deterministic partitions of the eight possible histories, positioned according to the amount of history information it retains, $\MI(H_t;Z_t)$ on the horizontal axis, and the amount of predictive information it preserves, $\MI(Z_t;X_{t+1})$ on the vertical axis. The blue curve connects the nondominated deterministic solutions and therefore gives the exact deterministic reference frontier.

The orange curve is obtained differently. We initialize a stochastic encoder at the overcomplete eight-history representation and follow a warm-started continuation path as the compression weight $\beta$ is increased. Each orange marker shows the representation obtained at one value of $\beta$. At $\beta=0$, the overcomplete initialization is retained with no compression pressure. For subsequent values, increasing $\beta$ makes representations with smaller $\MI(H_t;Z_t)$ increasingly preferable. The learned solution is therefore encouraged to move leftward in the information plane. Ideally, this removes redundant history while remaining near the top of the plot, where predictive information is preserved. If $\beta$ becomes too large, however, compression also removes information needed for prediction and the trajectory moves downward.

The labeled $\beta$ values do not represent different data-generating processes; they are different settings of the same predictive-compression objective and trace how the learned representation changes as compression pressure increases. The learned trajectory need not coincide with the exact blue frontier because the encoder is stochastic and is optimized by gradient descent, whereas the blue frontier is obtained by exhaustive enumeration over deterministic partitions. We therefore use the deterministic frontier as a global reference and the orange continuation path as a practical illustration of how predictive compression behaves during learning. Full enumeration details, the complete $\beta$ sweep, continuation optimization settings, and the corresponding state-count trajectories are provided in Appendix~\ref{app:experimental-details}.

\paragraph{Did compression remove too much information?}
The previous results identify which representations achieve a favorable trade-off between compression and prediction. We next ask a complementary question: can we detect when compression has gone too far and removed information that is still useful for predicting the future?

We instantiate the residual-history diagnostic from \cref{sec:residual-diagnostic} using one additional step of representation history. The intuition is simple. If $Z_t$ already contains all information needed to predict $X_{t+1}$, then additionally conditioning on the previous representation state $Z_{t-1}$ should not improve held-out prediction. Conversely, if $Z_{t-1}$ still contains transition-relevant information that is absent from $Z_t$, then the current representation is predictively insufficient.

We test the same three controlled representations:
\begin{equation*}
Z_t^{\mathrm{over}}=(X_{t-2},X_{t-1},X_t),
\qquad
Z_t^{\mathrm{suff}}=(X_{t-1},X_t),
\qquad
Z_t^{\mathrm{under}}=X_t.
\end{equation*}

For each representation, we compare two predictors of $X_{t+1}$. The \emph{restricted} predictor uses only the current representation,
\begin{equation*}
\widehat p_{\mathrm{restricted}}
=
\widehat p(X_{t+1}=1\mid Z_t),
\end{equation*}
whereas the \emph{history-augmented} predictor additionally receives the previous representation state,
\begin{equation*}
\widehat p_{\mathrm{history\text{-}augmented}}
=
\widehat p(X_{t+1}=1\mid Z_t,Z_{t-1}).
\end{equation*}
We define the residual history gain as
\begin{equation}
\Delta_{\mathrm{hist}}
=
\mathrm{MSE}_{\mathrm{restricted}}
-
\mathrm{MSE}_{\mathrm{history\text{-}augmented}}.
\label{eq:exp3-delta-hist}
\end{equation}
Thus,
\begin{equation*}
\Delta_{\mathrm{hist}}>0
\end{equation*}
means that the previous representation state contains predictive information not already captured by $Z_t$. By contrast,
\begin{equation*}
\Delta_{\mathrm{hist}}\approx0
\end{equation*}
means that adding one further step of representation history provides essentially no additional predictive benefit. The lookup predictors, chronological train--test split, and fitting procedure are detailed in Appendix~\ref{app:experimental-details}.

The three controlled representations make this diagnostic especially transparent. For the insufficient representation
\begin{equation*}
Z_t^{\mathrm{under}}=X_t,
\end{equation*}
the previous representation is simply
\begin{equation*}
Z_{t-1}^{\mathrm{under}}=X_{t-1}.
\end{equation*}
Thus, the history-augmented predictor restores exactly the variable omitted from $Z_t$ that is required by the second-order transition law. As expected, this produces a substantial held-out prediction gain,
\begin{equation*}
\Delta_{\mathrm{hist}}
=
0.1139\pm0.0026.
\end{equation*}
This is the intended positive control: $Z_t=X_t$ is insufficient because the omitted $X_{t-1}$ remains informative about $X_{t+1}$.

For the minimal sufficient representation,
\begin{equation*}
Z_t^{\mathrm{suff}}=(X_{t-1},X_t),
\end{equation*}
we have
\begin{equation*}
Z_{t-1}^{\mathrm{suff}}=(X_{t-2},X_{t-1}).
\end{equation*}
Since $X_{t-1}$ is already contained in $Z_t$, the only genuinely additional observation supplied by $Z_{t-1}$ is $X_{t-2}$, which is redundant for predicting $X_{t+1}$ by construction. Correspondingly,
\begin{equation*}
\Delta_{\mathrm{hist}}
\approx
-6.8\times10^{-6}.
\end{equation*}

For the overcomplete representation,
\begin{equation*}
Z_t^{\mathrm{over}}=(X_{t-2},X_{t-1},X_t),
\end{equation*}
the previous representation
\begin{equation*}
Z_{t-1}^{\mathrm{over}}
=
(X_{t-3},X_{t-2},X_{t-1})
\end{equation*}
adds only still older information beyond what is already available in $Z_t$. We obtain
\begin{equation*}
\Delta_{\mathrm{hist}}
\approx
-3.7\times10^{-5}.
\end{equation*}
Both near-zero values are negligible at the scale of the experiment. The tiny negative values are attributable to finite-sample fitting variation rather than a meaningful advantage of the restricted predictor. Once the current representation already contains all transition-relevant information, adding $Z_{t-1}$ does not improve held-out prediction.

Viewed together, the three controlled cases reveal a clear sufficiency--minimality boundary. Compressing from the overcomplete eight-state representation to the four-state minimal representation reduces $\MI(H_t;Z_t)$ from $1.7356$ to $1.3378$ while leaving $\Delta_{\mathrm{hist}}$ effectively zero, indicating that redundant history has been removed without sacrificing predictive sufficiency. Compressing further to the two-state representation reduces $\MI(H_t;Z_t)$ to $0.6785$, but $\Delta_{\mathrm{hist}}$ rises sharply to $0.1139\pm0.0026$: the previous representation $Z_{t-1}=X_{t-1}$ now contains substantial transition-relevant information missing from $Z_t=X_t$. Thus, in this controlled example, the four-state representation lies at the natural elbow between retaining redundant history and compressing away information required for prediction.

This result highlights the distinction between \emph{sufficiency} and \emph{minimality}. The residual-history diagnostic tests sufficiency: it correctly identifies $Z_t=X_t$ as missing predictive information, while both the four-state and eight-state representations pass because their current state already contains all information required for one-step prediction. The diagnostic cannot, however, determine that the eight-state representation stores redundant history. The predictive bottleneck supplies this complementary notion of minimality by preferring the smaller four-state representation among predictively sufficient alternatives.

Together, the two diagnostics play complementary roles. Predictive compression asks how much of the past can be discarded while preserving future prediction, thereby favoring a compact Markov state. Residual predictability asks whether compression has discarded too much: a positive $\Delta_{\mathrm{hist}}$ indicates that the previous representation state contains transition-relevant information not already captured by $Z_t$. In this controlled process, the four-state representation is the known minimal sufficient target: it preserves all one-step predictive information while storing less history than the overcomplete eight-state representation. Detailed enumeration, optimization settings, complete compression sweeps, residual-predictor specifications, and supplementary plots are provided in Appendix~\ref{app:experimental-details}.

\subsection{Experiment 4: HMM-style training of PIB-VJEPA}
\label{sec:exp-hmm-training}

Experiment~4 examines the final and strongest level of correspondence considered in this paper: \emph{model-and-objective equivalence}. Even when JEPA and HMM formulations share an emission-complete latent-state representation, or satisfy the conditions for sequence-level HMM equivalence, they need not be trained by the same objective. Standard JEPA training optimizes prediction in latent space, whereas HMM training additionally optimizes the probability of the observed sequence through an explicit transition--emission model.

To isolate this distinction, we use the same four-state latent family and separated-emission data-generating process as Experiment~1, but train all three Experiment~4 regimes independently. Because the latent state is categorical and the observation model is Gaussian, observation-sequence likelihood and filtering posteriors can be evaluated exactly using the HMM forward recursion. We compare three regimes---JEPA-only, HMM-style, and their hybrid---while keeping the latent-state family and amortized context-encoder architecture fixed.

\paragraph{Training regimes.}
The first regime is the \emph{JEPA latent objective}. It uses the same MCJEPA construction as Experiment~1:
\begin{equation}
\mathcal L_{\mathrm{JEPA}}
=
\mathcal L_{\mathrm{MC}}
+
\mathcal L_{\mathrm{state}},
\label{eq:exp4-jepa-objective}
\end{equation}
where $\mathcal L_{\mathrm{MC}}$ is defined in \cref{eq:mc-loss} and $\mathcal L_{\mathrm{state}}$ in \cref{eq:state-reg}. The online context encoder produces
\begin{equation*}
q_\theta(Z_t\mid X_{\leq t}),
\end{equation*}
while the EMA target encoder produces the local future target
\begin{equation*}
q_{\bar\theta}(Z_{t+h}\mid X_{t+h}).
\end{equation*}
A single row-stochastic transition matrix generates all horizons through $A^h$. This regime therefore represents the latent-prediction viewpoint: no observation-sequence likelihood or filtering target influences representation learning. For evaluation of sequence NLL, a Gaussian observation model is fitted only after JEPA training and consequently does not influence the learned representation.

The second regime is \emph{HMM sequence + filter distillation}. Here, the latent transition model
\begin{equation*}
p_\phi(Z_{t+1}\mid Z_t)
\end{equation*}
and Gaussian emission model
\begin{equation*}
p_\psi(X_t\mid Z_t)
\end{equation*}
are trained through exact observation-sequence negative log-likelihood,
\begin{equation}
\mathcal L_{\mathrm{HMM}}
=
-\E\left[
\frac{1}{T}
\log p_{\phi,\psi}(X_{1:T})
\right],
\label{eq:exp4-hmm-loss}
\end{equation}
computed by the HMM forward algorithm.

Sequence likelihood trains the generative transition--emission model, but it does not by itself require the amortized context encoder to represent the corresponding HMM filtering belief. We therefore additionally distill the exact filtering posterior into the context encoder through
\begin{equation}
\mathcal L_{\mathrm{filter}}
=
\E_t\left[
\KL\left(
\sg(\widetilde q_t)
\,\middle\|\,
q_\theta(Z_t\mid X_{\leq t})
\right)
\right].
\label{eq:exp4-filter-loss}
\end{equation}
where
\begin{equation*}
\widetilde q_t
=
p_{\phi,\psi}(Z_t\mid X_{\leq t})
\end{equation*}
denotes the exact filtering posterior under the current HMM transition and emission models.\footnote{Both $\widetilde q_t$ and $q_\theta(Z_t\mid X_{\leq t})$ represent a belief over the current latent state after observations through time $t$ have been assimilated. The exact HMM filter first propagates the previous filtering belief through the transition kernel,
\[
\pi_t^-(z_t)
=
\sum_{z_{t-1}}
p_\phi(z_t\mid z_{t-1})
\widetilde q_{t-1}(z_{t-1}),
\]
and then incorporates the current observation through the emission likelihood,
\[
\widetilde q_t(z_t)
=
\frac{
p_\psi(X_t\mid z_t)\,
\pi_t^-(z_t)
}{
\sum_{z_t'}
p_\psi(X_t\mid z_t')\,
\pi_t^-(z_t')
}.
\]
Thus, $\phi$ determines how probability mass is propagated between latent states, whereas $\psi$ determines how the current observation updates that predictive prior. For a discrete HMM with transition matrix $A$,
\[
\widetilde q_t(j)
=
\frac{
p_\psi(X_t\mid Z_t=j)
\sum_i\widetilde q_{t-1}(i)A_{ij}
}{
\sum_{j'}
p_\psi(X_t\mid Z_t=j')
\sum_i\widetilde q_{t-1}(i)A_{ij'}
}.
\]
Hence $\mathcal L_{\mathrm{filter}}$ compares like with like: it distills the exact current-state filtering belief into the amortized context encoder rather than comparing the context encoder with the pre-observation predictive prior.}

The complete HMM-style regime uses
\begin{equation}
\mathcal L_{\mathrm{HMM+filter}}
=
\lambda_{\mathrm{seq}}\mathcal L_{\mathrm{HMM}}
+
\lambda_{\mathrm{filter}}\mathcal L_{\mathrm{filter}}
+
\mathcal L_{\mathrm{state}}.
\label{eq:exp4-hmm-filter-objective}
\end{equation}
Importantly, this regime contains \emph{no JEPA latent-prediction loss $\mathcal L_{\mathrm{MC}}$}.

The third regime is the \emph{hybrid HMM + latent} objective:
\begin{equation}
\begin{aligned}
\mathcal L_{\mathrm{hybrid}}
={}&
\underbrace{
\lambda_{\mathrm{seq}}\mathcal L_{\mathrm{HMM}}
}_{\text{sequence evidence}}
+
\underbrace{
\lambda_{\mathrm{latent}}\mathcal L_{\mathrm{MC}}
}_{\text{JEPA latent prediction}}
\\
&+
\underbrace{
\lambda_{\mathrm{filter}}\mathcal L_{\mathrm{filter}}
}_{\text{filtering-posterior alignment}}
+
\underbrace{
\mathcal L_{\mathrm{state}}
}_{\text{state-use regularization}}.
\end{aligned}
\label{eq:exp4-hybrid-objective}
\end{equation}
Importantly, $\mathcal L_{\mathrm{MC}}$ is implemented in exactly the same way as in the JEPA-only regime: the future latent target is produced by the EMA target encoder,
\begin{equation*}
q_{\bar\theta}(Z_{t+h}\mid X_{t+h}),
\end{equation*}
and the context prediction is generated by the shared transition,
\begin{equation*}
q_\theta(Z_t\mid X_{\leq t})A^h.
\end{equation*}
The exact HMM filtering posterior $\widetilde q_t$ is used only in $\mathcal L_{\mathrm{filter}}$ and does not replace the JEPA target in $\mathcal L_{\mathrm{MC}}$. Moreover, the transition matrix used by the HMM sequence model is the same transition matrix used by the latent-prediction objective, so both training signals act on the same latent dynamics.

The state-use regularizer and its coefficients are shared across all three encoder-training regimes. The remaining objective weights, initialization scheme, and optimization settings are given in Appendix~\ref{app:experimental-details}.

\paragraph{Motivation and comparison design.}
The three regimes form a controlled objective-level comparison. The \emph{JEPA latent objective} uses latent predictive alignment but no observation-sequence likelihood or filtering target. The \emph{HMM sequence + filter distillation} regime does the converse: it uses observation-sequence likelihood and filtering-posterior supervision but no JEPA latent-prediction loss. The \emph{hybrid} regime adds both HMM-style signals to the same MCJEPA latent-prediction objective used by the JEPA-only baseline.

This comparison addresses three related questions. First, does adding HMM-style sequence and filtering supervision improve MCJEPA relative to latent-only training? Second, can the hybrid retain the genuine JEPA latent-prediction objective while approaching the probabilistic-model recovery achieved by HMM-style training? Third, is $\mathcal L_{\mathrm{MC}}$ necessary for training this latent-state architecture at all, or can the same architecture instead be trained through HMM-style sequence and filtering supervision?

\begin{table}[t]
\centering
\small
\renewcommand{\arraystretch}{1.10}
\caption{Objective-level comparison in Experiment~4. The three objective columns indicate which training signals are active: HMM sequence likelihood $\mathcal L_{\mathrm{HMM}}$, JEPA/Markov latent prediction $\mathcal L_{\mathrm{MC}}$, and filtering-posterior distillation $\mathcal L_{\mathrm{filter}}$. The state-use regularizer $\mathcal L_{\mathrm{state}}$ is shared across all three regimes. Values are mean $\pm$ standard deviation over five seeds. Boldface marks the numerically best mean and does not imply statistical significance. Filtering KL is training aligned for the HMM+filter and hybrid regimes because both explicitly optimize $\mathcal L_{\mathrm{filter}}$; for JEPA-only it is evaluated post hoc.}
\label{tab:exp4-objectives}
\resizebox{\linewidth}{!}{
\begin{tabular}{lccc|cccc}
\toprule
Training regime
&
$\mathcal L_{\mathrm{HMM}}$
&
$\mathcal L_{\mathrm{MC}}$
&
$\mathcal L_{\mathrm{filter}}$
&
ARI $\uparrow$
&
Seq.\ NLL $\downarrow$
&
$\mathcal E_A\downarrow$
&
Filter KL $\downarrow$
\\
\midrule
HMM sequence + filter distill
&
$\checkmark$
&
--
&
$\checkmark$
&
$\mathbf{0.9945\pm0.0008}$
&
$\mathbf{1.2659\pm0.0072}$
&
$0.0105\pm0.0014$
&
$\mathbf{0.00006\pm0.00001}$
\\
Hybrid HMM + latent
&
$\checkmark$
&
$\checkmark$
&
$\checkmark$
&
$0.9905\pm0.0008$
&
$1.2660\pm0.0073$
&
$\mathbf{0.0103\pm0.0017}$
&
$0.00436\pm0.00102$
\\
JEPA latent objective
&
--
&
$\checkmark$
&
--
&
$0.9802\pm0.0053$
&
$1.2753\pm0.0069$
&
$0.0191\pm0.0020$
&
$0.02546\pm0.00642$
\\
\bottomrule
\end{tabular}
}
\end{table}

\paragraph{Results.}
\Cref{tab:exp4-objectives} reveals a clear objective-level distinction. The JEPA-only model successfully learns a meaningful predictive latent state, reaching
\begin{equation*}
\mathrm{ARI}
=
0.9802\pm0.0053,
\end{equation*}
but its observation-sequence NLL is
\begin{equation*}
1.2753\pm0.0069,
\end{equation*}
and its transition-recovery error is
\begin{equation*}
\mathcal E_A
=
0.0191\pm0.0020.
\end{equation*}
This is consistent with what the objective directly supervises: $\mathcal L_{\mathrm{MC}}$ trains predictive agreement in latent space, but does not directly maximize observation-sequence likelihood or jointly train an emission model with the representation.

Adding HMM-style supervision produces a substantial improvement. The hybrid retains the same EMA-target MCJEPA loss but additionally optimizes sequence likelihood and filtering alignment. Its ARI rises to
\begin{equation*}
0.9905\pm0.0008,
\end{equation*}
its sequence NLL decreases to
\begin{equation*}
1.2660\pm0.0073,
\end{equation*}
and its transition error falls to
\begin{equation*}
\mathcal E_A
=
0.0103\pm0.0017.
\end{equation*}
Relative to JEPA-only training, this corresponds to an approximately $46\%$ reduction in transition-matrix error. Moreover, the improvement is seed-consistent: for each of the five random seeds, the hybrid improves over JEPA-only on ARI, NMI, sequence NLL, transition recovery, filtering KL, and true-state prediction NLL at every evaluated horizon $h\in\{1,2,4,8\}$. Thus, the gain from HMM-style supervision is not driven by a single favorable run.

The hybrid also nearly closes the observation-sequence likelihood gap to HMM-style training. The HMM+filter regime achieves sequence NLL
\begin{equation*}
1.265885\pm0.007220,
\end{equation*}
whereas the hybrid obtains
\begin{equation*}
1.266019\pm0.007271.
\end{equation*}
Their difference is only
\begin{equation*}
1.34\times10^{-4}
\end{equation*}
NLL per time step, compared with a JEPA-to-HMM gap of approximately
\begin{equation*}
9.41\times10^{-3}.
\end{equation*}
Equivalently, the hybrid closes approximately $98.6\%$ of the JEPA-only sequence-NLL gap to HMM-style training while retaining the genuine JEPA latent-prediction objective.

Transition recovery shows the same qualitative result. The hybrid has the numerically smallest mean error,
\begin{equation*}
0.0103\pm0.0017,
\end{equation*}
compared with
\begin{equation*}
0.0105\pm0.0014
\end{equation*}
for HMM+filter and
\begin{equation*}
0.0191\pm0.0020
\end{equation*}
for JEPA-only. The difference between hybrid and HMM-style training is small relative to the across-seed variability, so we interpret the two as achieving comparable transition recovery rather than claiming that the hybrid is superior to the correctly specified HMM objective. The important contrast is that both recover the transition substantially more faithfully than latent-only JEPA training.

The HMM+filter regime provides the complementary result. Despite containing no $\mathcal L_{\mathrm{MC}}$, it achieves the strongest state recovery,
\begin{equation*}
\mathrm{ARI}
=
0.9945\pm0.0008,
\end{equation*}
and
\begin{equation*}
\mathrm{NMI}
=
0.9884\pm0.0014,
\end{equation*}
together with the best sequence NLL and transition recovery comparable to the hybrid. Thus, the JEPA latent-prediction objective is not required to train this categorical latent-state architecture successfully: the same architecture can instead be trained using HMM-style sequence and filtering supervision, together with the common state-use regularizer.

The multi-horizon prediction results reinforce this conclusion. At every evaluated horizon $h\in\{1,2,4,8\}$, both HMM-style and hybrid training achieve lower true-state prediction NLL than JEPA-only training. The differences are largest at shorter horizons and diminish at longer horizons as the transition dynamics mix. Complete multi-horizon results are provided in Appendix~\ref{app:experimental-details}.

\paragraph{Filtering as a role diagnostic.}
Filtering KL requires a different interpretation from sequence NLL and transition recovery. The HMM+filter regime reaches
\begin{equation*}
\mathrm{KL}_{\mathrm{filter}}
=
0.000060\pm0.000007,
\end{equation*}
which is expected because its amortized encoder is explicitly trained to reproduce the exact HMM filtering posterior. The hybrid obtains
\begin{equation*}
0.004355\pm0.001015,
\end{equation*}
while JEPA-only gives
\begin{equation*}
0.025460\pm0.006423.
\end{equation*}
For the HMM+filter and hybrid regimes this quantity is training aligned and should therefore be interpreted as a diagnostic that the intended filtering role has been learned, rather than as an independent generalization metric. For JEPA-only, by contrast, the filtering distribution is constructed only after fitting the post-hoc observation model, so its filtering KL measures how closely latent-only representation learning happens to agree with the filter induced by that fitted probabilistic model.

The ordering is nevertheless informative about the roles induced by the different objectives. HMM+filter explicitly learns an amortized filter; the hybrid remains substantially aligned with that filtering interpretation while simultaneously satisfying the EMA-target JEPA objective; and JEPA-only has no requirement that its history encoder coincide with a Bayesian filtering belief. The corresponding diagnostic is reported separately in Appendix~\ref{app:experimental-details}.

\Cref{fig:exp4-objectives} focuses on the two metrics that most directly expose the objective-level distinction. The left panel reports observation-sequence NLL relative to HMM-style training. The hybrid lies almost on the HMM reference, whereas JEPA-only retains a clear positive gap. The right panel reports permutation-aligned transition recovery error: HMM-style and hybrid training form a closely matched pair, while JEPA-only exhibits substantially larger error.

\begin{figure}[t]
\centering
\begin{minipage}[t]{0.49\linewidth}
\centering
\includegraphics[width=\linewidth]{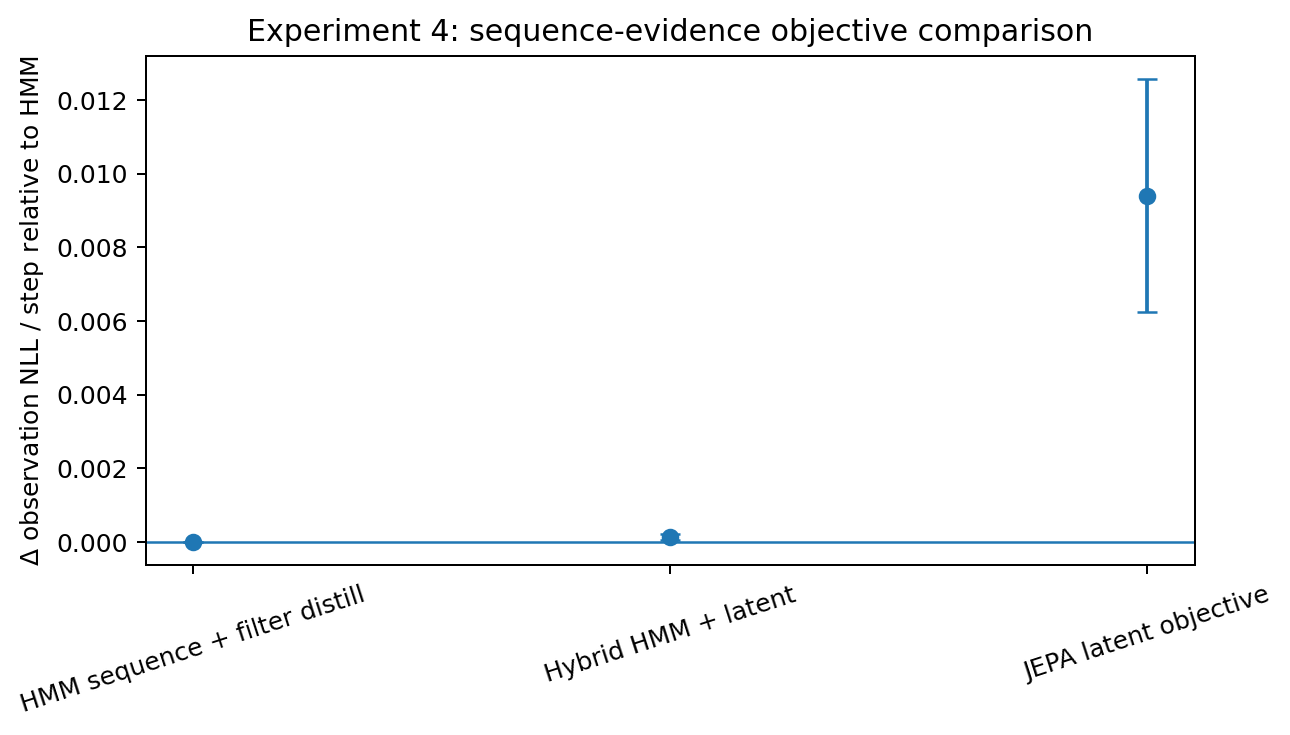}
\end{minipage}
\hfill
\begin{minipage}[t]{0.49\linewidth}
\centering
\includegraphics[width=\linewidth]{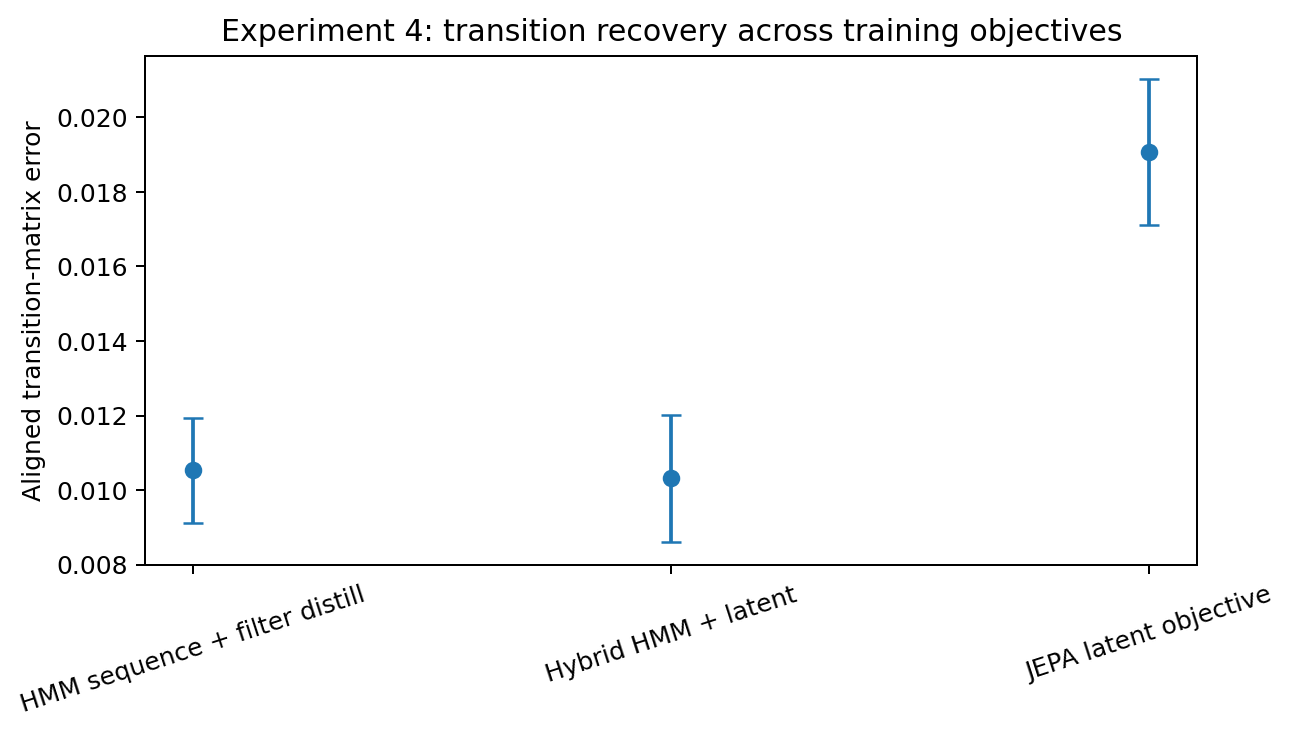}
\end{minipage}
\caption{Objective-level comparison in Experiment~4. \emph{Left:} observation-sequence NLL per time step relative to the HMM sequence + filter-distillation regime. Hybrid training nearly closes the entire JEPA-to-HMM sequence-likelihood gap while retaining the genuine MCJEPA latent-prediction objective. \emph{Right:} permutation-aligned transition-matrix recovery error. HMM-style and hybrid training achieve closely matched transition recovery, and both substantially outperform latent-only JEPA training. Error bars denote mean $\pm$ one standard deviation over five seeds.}
\label{fig:exp4-objectives}
\end{figure}

Taken together, these results support the distinction developed earlier between an HMM-compatible latent-state representation and full model-and-objective equivalence. The same HMM-compatible latent-state architecture can support JEPA-style latent prediction, HMM-style probabilistic sequence learning, or a combination of the two, but sharing the underlying probabilistic model class does not imply that different training objectives recover the same fitted model. Explicit sequence likelihood supplies transition--emission supervision that latent prediction alone does not provide, while filtering distillation connects the resulting HMM posterior back to the amortized PIB-VJEPA context encoder.

Experiment~4 supports two particularly important practical conclusions. First, incorporating HMM-style sequence and filtering supervision into MCJEPA improves recovery of the underlying probabilistic latent-state model while retaining the original JEPA latent-prediction objective. The hybrid improves over JEPA-only training on every reported metric for every seed, nearly matches HMM-style sequence likelihood, and recovers the transition dynamics at essentially the same level as HMM-style training. Second, the same latent-state architecture can be trained successfully without $\mathcal L_{\mathrm{MC}}$: the HMM sequence + filter-distillation regime attains the strongest state recovery and sequence likelihood despite omitting the JEPA latent-prediction objective altogether. Thus, the distinction between MCJEPA and an HMM is not determined by architecture alone; it also depends fundamentally on the objective used to train that architecture.

More generally, the three regimes expose a continuum of training objectives on the same latent-state family:
\begin{equation*}
\underbrace{
\mathcal L_{\mathrm{MC}}
}_{\text{JEPA-style training}}
\quad\longleftrightarrow\quad
\underbrace{
\mathcal L_{\mathrm{MC}}
+
\mathcal L_{\mathrm{HMM}}
+
\mathcal L_{\mathrm{filter}}
}_{\text{hybrid training}}
\quad\longleftrightarrow\quad
\underbrace{
\mathcal L_{\mathrm{HMM}}
+
\mathcal L_{\mathrm{filter}}
}_{\text{HMM-style training}},
\end{equation*}
with $\mathcal L_{\mathrm{state}}$ shared as a state-use regularizer. The hybrid demonstrates that HMM-style probabilistic sequence learning can be incorporated without abandoning JEPA-style latent prediction, while the HMM-style end of the spectrum shows that the same architecture can also be trained without latent-prediction supervision altogether. This objective continuum makes precise the paper's broader claim: probabilistic temporal JEPA and HMMs can share an underlying latent Markov architecture while differing in how strongly HMM-equivalent probabilistic semantics are enforced by training.

Further optimization details, complete multi-horizon results, and the training-aligned filtering diagnostic are reported in Appendix~\ref{app:experimental-details}.

\subsection{Summary of Experiments}
The four experiments test complementary and progressively stronger aspects of the proposed HMM interpretation of probabilistic temporal JEPA. Experiment~1 establishes the finite-state Markov structure: MCJEPA learns an explicit shared transition matrix whose powers generate all prediction horizons, thereby guaranteeing exact direct-versus-composed consistency. The correctly specified Gaussian HMM remains strongest when emissions are ambiguous, while independently trained horizon-specific predictors can gain some predictive flexibility at the cost of a less faithful and non-compositional transition law. The collapse ablations further show that occupancy and entropy regularization play complementary roles in stable discrete-state learning.

Experiment~2 isolates the inference role of the context encoder. Using exact oracle posteriors, it shows that history-based filtering $p(S_t\mid X_{\leq t})$ becomes increasingly more informative than local evidence $p(S_t\mid X_t)$ as emissions overlap. This supports the interpretation of a history-dependent context encoder $q_\theta(Z_t\mid X_{\leq t})$ as an amortized filtering belief rather than an emission model.

Experiment~3 addresses how such a Markov state can be constructed. In a controlled second-order binary process, predictive compression removes the redundant history variable $X_{t-2}$ and selects the known minimal sufficient state $Z_t^\star=(X_{t-1},X_t)$ without loss of predictive information, whereas further compression becomes insufficient. Exhaustive enumeration of all deterministic history partitions provides a global reference for the prediction--compression trade-off, while the residual-history diagnostic supplies the complementary sufficiency test: adding the previous representation state $Z_{t-1}$ provides essentially no predictive benefit once $Z_t$ is sufficient, but yields a large held-out gain when $Z_t=X_t$ has discarded the transition-relevant variable $X_{t-1}$. Together, these results separate \emph{minimality} from \emph{sufficiency} and show how predictive compression can Markovize an observation process by constructing a compact predictive state.

Finally, Experiment~4 makes the objective-level distinction explicit through a controlled comparison on the same latent-state family. The architecture can be trained with the original JEPA-style latent-prediction objective, HMM-style sequence likelihood and filtering distillation, or a hybrid containing all three signals. Adding HMM-style supervision to the genuine MCJEPA objective improves state recovery, observation-sequence likelihood, transition recovery, filtering agreement, and multi-horizon prediction relative to latent-only training. The hybrid nearly closes the entire JEPA-to-HMM sequence-likelihood gap and recovers the transition dynamics at essentially the same level as HMM-style training, while retaining the original EMA-target JEPA latent-prediction objective. Conversely, the HMM-style regime achieves the strongest state recovery and sequence likelihood despite using no $\mathcal L_{\mathrm{MC}}$, showing that the same latent-state architecture can also be trained successfully without JEPA latent-prediction supervision.

Taken together, the experiments support a progressively stronger view of probabilistic temporal JEPA: it can instantiate a coherent latent Markov architecture; its context encoder can acquire the role of a filtering distribution; predictive compression can construct a compact sufficient Markov state; and the same architecture can be trained along a continuum from JEPA-style latent prediction, through hybrid JEPA--HMM learning, to HMM-style probabilistic sequence learning without $\mathcal L_{\mathrm{MC}}$. The degree of HMM equivalence therefore depends not only on architectural structure, but also on the probabilistic components and, critically, the objective used to train them.

\section{Discussion}

\paragraph{What is ``secretly an HMM''?}
The central claim is structural and probabilistic, but not unconditional. Full, time-indexed PIB-VJEPA exposes the same three computational roles as an HMM: inference of a latent-state belief from observations, propagation of that state through a Markov transition, and a state-to-observation map. The correspondence becomes progressively stronger across the four levels developed in this paper: computational correspondence, emission-complete latent-state representation, sequence-level HMM equivalence, and model-and-objective equivalence. In particular, the stochastic encoder is \emph{not} an emission model; it plays the recognition or filtering role. The emission direction may instead be supplied by a decoder, by the inverse of an invertible target encoder, or implicitly through the Bayes-consistent conditional induced by a local stochastic encoder. Reaching sequence-level HMM equivalence further requires Markov, marginal-consistency, and filtering-consistency conditions, as formalized in \cref{thm:exact-hmm-representation}; reaching model-and-objective equivalence additionally requires HMM-compatible sequence-level probabilistic training.

\paragraph{Beyond MCJEPA: when is a general JEPA predictor Markov?}
The HMM correspondence is not specific to MCJEPA, nor does a neural-network predictor cease to be Markov merely because it is nonlinear or highly expressive. Markovianity is a conditional-independence property rather than a restriction on the functional form of the predictor. A general temporal JEPA may use an arbitrary neural transition
\begin{equation*}
p_\phi(Z_{t+1}\mid Z_t,\xi_t),
\end{equation*}
implemented, for example, by an MLP, Transformer, mixture model, or another conditional density estimator. It remains \textit{first-order} Markov with respect to $Z_t$ whenever
\begin{equation*}
p_\phi(Z_{t+1}\mid Z_{\leq t},\xi_{\leq t})
=
p_\phi(Z_{t+1}\mid Z_t,\xi_t).
\end{equation*}
The transition can therefore be arbitrarily nonlinear; for example,
\begin{equation*}
p_\phi(Z_{t+1}\mid Z_t,\xi_t)
=
\mathcal N\!\left(
Z_{t+1};
\mu_\phi(Z_t,\xi_t),
\Sigma_\phi(Z_t,\xi_t)
\right),
\end{equation*}
while a deterministic temporal JEPA is recovered through the Dirac kernel
\begin{equation*}
p_\phi(Z_{t+1}\mid Z_t,\xi_t)
=
\delta\!\left(
Z_{t+1}-P_\phi(Z_t,\xi_t)
\right).
\end{equation*}
MCJEPA is therefore an explicit finite-state instantiation of a broader latent-Markov interpretation of temporal JEPA: it replaces the general transition kernel by
\begin{equation*}
p_\phi(Z_{t+1}=j\mid Z_t=i)=A_{ij},
\end{equation*}
making the Markov property and multi-step Chapman--Kolmogorov composition especially transparent. Conceptually,
\begin{equation*}
\text{MCJEPA}
\subset
\text{Markov neural-predictor JEPA}
\subset
\text{general temporal JEPA}.
\end{equation*}
Thus, MCJEPA demonstrates the correspondence in its simplest explicit form; it does not create or solely represent the correspondence.

If the predictor genuinely depends on information beyond $Z_t$, however, the latent process need not be first-order Markov in $Z_t$ alone. For example, a recurrent predictor may use (Eq.\ref{eq:MC_memory})
\begin{equation*}
p_\phi(Z_{t+1}\mid Z_t,M_t,\xi_t),
\end{equation*}
where $M_t$ summarizes additional history (i.e. memory), or a higher-order predictor may depend directly on $(Z_t,\ldots,Z_{t-k+1})$. In such cases, exact HMM correspondence with $Z_t$ as the hidden state does not follow. However, a first-order representation can often be recovered by \textit{augmenting the state}. For recurrent dynamics, one may define
\begin{equation*}
\widetilde Z_t=(Z_t,M_t),
\end{equation*}
while for a $k$th-order predictor one may use
\begin{equation*}
\widetilde Z_t=
(Z_t,Z_{t-1},\ldots,Z_{t-k+1}).
\end{equation*}
If the augmented state contains all transition-relevant information from the past, then
\begin{equation*}
p(
\widetilde Z_{t+1}
\mid
\widetilde Z_{\leq t},
\xi_{\leq t})
=
p(
\widetilde Z_{t+1}
\mid
\widetilde Z_t,
\xi_t),
\end{equation*}
and the first-order latent-state interpretation is restored. The correspondence developed in this paper therefore applies beyond MCJEPA to general probabilistic temporal JEPA whenever the chosen latent state, possibly after augmentation, admits such a first-order Markov transition together with the additional emission and consistency conditions required for sequence-level equivalence.

\paragraph{Predictive representation learning as Markov-state construction.}
The HMM perspective changes the interpretation of the JEPA representation itself. Rather than viewing $Z_t$ only as a feature vector useful for predicting another feature vector, we can ask whether it constitutes a \emph{predictive state}: does it retain the information from the past that is needed for future prediction while discarding redundant history? This distinction also separates two notions that can otherwise be conflated. A predictor may be architecturally first-order,
\begin{equation*}
p_\phi(Z_{t+1}\mid Z_t,\xi_t),
\end{equation*}
without $Z_t$ being a sufficient Markov representation of the underlying process. In particular,
\begin{equation*}
\text{first-order predictor}
\;\not\Rightarrow\;
\text{sufficient Markov state}.
\end{equation*}
If older history remains predictive after conditioning on $Z_t$, for example if
\begin{equation*}
p(Z_{t+1}\mid Z_t,Z_{t-1})
\neq
p(Z_{t+1}\mid Z_t),
\end{equation*}
then the architecture is imposing a first-order transition on a representation that has not fully Markovized the process.

This gives predictive information-bottleneck learning a state-space interpretation. Ideally, the representation should satisfy predictive sufficiency,
\begin{equation*}
X_{>t}
\perp
X_{\leq t}
\mid
Z_t,
\end{equation*}
while retaining as little redundant information about the past as necessary. Compression therefore promotes minimality, whereas predictive sufficiency prevents over-compression. Together they can transform a non-Markov observation process into an approximately Markov latent process rather than merely forcing a Markov predictor onto an insufficient representation. Experiment~3 illustrates this distinction explicitly: compression removes redundant history to recover the known minimal sufficient state, while the residual-history diagnostic tests whether transition-relevant information remains outside the current representation. Thus, compression and residual predictability provide complementary tools for \emph{learning and testing} a Markov representation rather than assuming one a priori.

\paragraph{Architecture and objective are separate design choices.}
The HMM correspondence also clarifies a distinction that is easy to obscure: sharing an encode--transition--emit architecture does not imply sharing a training objective. Standard JEPA training may optimize only latent predictive alignment and never maximize observation-sequence likelihood. Experiment~4 makes this distinction operational. Adding HMM-style sequence and filtering supervision to the genuine MCJEPA objective substantially improves recovery of the probabilistic latent-state model, with the hybrid approaching HMM-level sequence likelihood and transition recovery while retaining EMA-target latent prediction. Conversely, the same latent-state architecture can be trained successfully using HMM sequence likelihood and filtering distillation without $\mathcal L_{\mathrm{MC}}$ at all. The resulting continuum
\begin{equation*}
\text{JEPA latent prediction}
\;\longleftrightarrow\;
\text{hybrid JEPA--HMM training}
\;\longleftrightarrow\;
\text{HMM-style sequence learning}
\end{equation*}
shows that the boundary between probabilistic temporal JEPA and classical state-space modeling is determined not only by model components, but also by which probabilistic semantics the objective enforces.

\paragraph{Why observation reconstruction can remain optional.}
An HMM explicitly models $p(X_t\mid S_t)$ because its likelihood is defined in observation space. JEPA may instead deliberately concentrate learning on the information required for future prediction, avoiding the cost of reconstructing high-entropy observation details that are irrelevant to the predictive task. A decoder can be introduced when observation forecasting or sequence likelihood is required, but it need not participate in the core representation-learning objective. An invertible target encoder provides another realization of the observation map, although invertibility limits the encoder's ability to discard nuisance information or reduce dimensionality and can therefore conflict with predictive compression. The implicit-emission construction establishes a probabilistic completion even without either explicit map, but that induced conditional need not be tractable enough for practical generation or likelihood evaluation.

\paragraph{Scope and limitations.}
The first-order Markov property should therefore be understood as a \emph{representation-design target}, not as a generic property of neural embeddings. A neural predictor that consumes only $Z_t$ is architecturally first-order, but this alone does not establish that $Z_t$ contains all transition-relevant information from the past. If residual history remains predictive, the representation is insufficient at the chosen temporal scale; the state can instead be enlarged, augmented with recurrent memory, modeled with higher-order dynamics, or predicted using an unrestricted history-dependent model. Likewise, finite categorical states and transition matrices improve structural interpretability but do not automatically produce semantically meaningful state labels. Such semantics must be established through observation statistics, transition behavior, interventions, or downstream tasks. Finally, our experiments are intentionally controlled and synthetic: they isolate composition, filtering, Markov-state construction, and objective-level behavior under known dynamics. Extending these diagnostics to high-dimensional video, control, and real-world partially observed systems is therefore an important empirical next step.

\section{Conclusion}

We developed a state-space interpretation of probabilistic temporal JEPA and made it concrete through \emph{Markov-Chain JEPA} (\ours). MCJEPA replaces the latent predictor by a learned row-stochastic transition matrix, so that multi-step prediction is generated by $A^h$ and direct and composed predictions satisfy exact Chapman--Kolmogorov consistency. Neural conditioned matrices, continuous-state Markov kernels, and continuous-time transitions extend this construction beyond finite homogeneous chains, while deterministic temporal JEPA appears as a degenerate transition kernel\footnote{As shown in \cref{sec:deterministic-jepa-degenerate}, a deterministic predictor is a Dirac Markov kernel, and a deterministic latent representation can likewise be viewed as a point-mass state distribution. Thus, the latent Markov perspective is not restricted to probabilistic JEPA: probabilistic formulations expose the state-space structure explicitly, while classical JEPA occupies its deterministic boundary.}.

The broader contribution is to make precise when this latent Markov view becomes an HMM interpretation. Observation-level data correspond to HMM observations; the stochastic context encoder plays the filtering role; the probabilistic predictor defines latent transition dynamics; and a decoder, inverse target encoder, or induced implicit conditional supplies the emission direction. We distinguish computational correspondence, emission-complete latent-state representation, sequence-level HMM equivalence, and model-and-objective equivalence, and give sufficient conditions in \cref{thm:exact-hmm-representation} under which the resulting model admits an exact sequence-level HMM representation. Because classical deterministic JEPA is recovered through point-mass latent distributions and Dirac transitions, the same computational and latent-Markov correspondence extends to the classical setting in the corresponding degenerate sense, although stronger HMM equivalence still requires the emission and consistency conditions identified above.

This perspective also yields a representation-learning principle: predictive information bottleneck learning can be understood as seeking a compact predictive state that approximately \emph{Markovizes} the observed process at the chosen prediction scale. Compression promotes \emph{minimality} by removing redundant history, while residual predictability tests \emph{sufficiency} by detecting transition-relevant information that remains outside the current state. Finally, the objective-level experiments show that the same latent-state architecture supports a continuum from JEPA latent prediction, through hybrid JEPA--HMM learning, to HMM-style sequence and filtering training. Probabilistic temporal JEPA is therefore not simply an HMM under a different name; rather, it exposes an HMM-compatible latent state-space structure whose probabilistic semantics become progressively stronger as emission completeness, transition and marginal consistency, filtering consistency, and HMM-style sequence training are imposed.

\bibliographystyle{tmlr}
\bibliography{tmlr}
\appendix

\section{Training Objectives and Minimal Algorithm}
\label{app:training}

For the basic time-homogeneous MCJEPA model with one shared transition matrix $A$, the training objective is
\begin{equation*}
\mathcal L
=
\mathcal L_{\mathrm{MC}}
+
\mathcal L_{\mathrm{state}},
\end{equation*}
where $\mathcal L_{\mathrm{MC}}$ is the multi-horizon latent-prediction objective in \cref{eq:mc-loss} and $\mathcal L_{\mathrm{state}}$ is the discrete-state regularizer in \cref{eq:state-reg}. Because all $h$-step predictions are generated by powers of the same matrix $A$, exact path consistency follows automatically from \cref{prop:exact_path_consistency}; no additional Chapman--Kolmogorov penalty is required.

A softer alternative may instead parameterize separate horizon-dependent transition matrices $A_h$. In that case, path consistency is no longer guaranteed and may be encouraged through
\begin{equation*}
\mathcal L_{\mathrm{CK}}
=
\sum_{h_1,h_2}
\left\|
A_{h_1+h_2}
-
A_{h_1}A_{h_2}
\right\|_F^2,
\end{equation*}
with an additional weight $\lambda_{\mathrm{CK}}\geq0$. This penalty belongs only to the horizon-dependent variant and is unnecessary for the shared-$A$ MCJEPA used in the main experiments.

For the basic shared-$A$ model, a minimal training step is:
\begin{enumerate}
    \item sample a time index $t$, prediction horizon $h\in\mathcal H$, observation history $X_{\leq t}$, future target $X_{t+h}$, and any required side information;
    \item compute the current-state distribution
    \[
    q_t
    =
    q_\theta(Z_t\mid X_{\leq t});
    \]
    \item compute the EMA target distribution
    \[
    \bar q_{t+h}
    =
    q_{\bar\theta}(Z_{t+h}\mid X_{t+h});
    \]
    \item propagate the current state through the shared transition matrix,
    \[
    \widehat q_{t+h}
    =
    q_tA^h;
    \]
    \item evaluate the latent-prediction loss in \cref{eq:mc-loss} using $\sg(\bar q_{t+h})$ as the target, and add the state-use regularizer $\mathcal L_{\mathrm{state}}$;
    \item update the online encoder parameters $\theta$ and the learnable transition parameters, then update the target encoder by exponential moving average,
    \[
    \bar\theta
    \leftarrow
    \tau\bar\theta
    +
    (1-\tau)\theta,
    \qquad
    \tau\in[0,1).
    \]
\end{enumerate}

For the conditioned discrete-state model in \cref{eq:conditioned-transition}, Step~4 is replaced by the ordered transition composition
\begin{equation*}
\widehat q_{t+h}
=
q_t
\prod_{j=0}^{h-1}
A_\phi(\xi_{t+j}),
\end{equation*}
as defined in \cref{eq:conditioned-multistep}. The remainder of the training procedure is unchanged.

The residual quantities introduced in \cref{sec:residual-diagnostic} are used as held-out diagnostics of state and transition sufficiency rather than as part of the default MCJEPA training objective. In particular, residual-history gain is evaluated after fitting the representation and transition model so that residual predictability can diagnose information omitted from the current state without directly training the representation to satisfy the diagnostic.

\section{Proofs}

\subsection{Proof of Proposition~\ref{prop:exact_path_consistency}}
\label{app:proof-exact-path-consistency}

For nonnegative integers $h_1$ and $h_2$, the definition of matrix powers together with associativity of matrix multiplication gives
\begin{equation*}
A^{h_1+h_2}
=
A^{h_1}A^{h_2}.
\end{equation*}
Left-multiplying by the current state distribution $q_t$ yields
\begin{equation*}
q_tA^{h_1+h_2}
=
\left(q_tA^{h_1}\right)A^{h_2},
\end{equation*}
which is \cref{eq:exact-path}.

More generally, let a total horizon $h$ be partitioned into nonnegative integers
\begin{equation*}
h
=
h_1+\cdots+h_m.
\end{equation*}
Repeated application of the same identity gives
\begin{equation*}
A^h
=
A^{h_1}\cdots A^{h_m},
\end{equation*}
and therefore
\begin{equation*}
q_tA^h
=
\bigl(\cdots((q_tA^{h_1})A^{h_2})\cdots\bigr)A^{h_m}.
\end{equation*}
Hence every decomposition of the same total horizon produces the same predictive distribution, proving the proposition.

\subsection{Proof of Proposition~\ref{prop:implicit-emission-completion}}
\label{app:proof-implicit-emission-completion}

Recall from \cref{eq:implicit-emission} that, whenever $q_\theta(z)>0$,
\begin{equation*}
p_\theta^{\mathrm{imp}}(x\mid z)
=
\frac{
p_{\mathrm{data}}(x)q_\theta(z\mid x)
}{
q_\theta(z)
},
\end{equation*}
where
\begin{equation*}
q_\theta(z)
=
\int
p_{\mathrm{data}}(x)
q_\theta(z\mid x)
\,dx.
\end{equation*}

For any $z$ with $q_\theta(z)>0$,
\begin{align*}
\int
p_\theta^{\mathrm{imp}}(x\mid z)
\,dx
&=
\frac{1}{q_\theta(z)}
\int
p_{\mathrm{data}}(x)
q_\theta(z\mid x)
\,dx
\\
&=
\frac{q_\theta(z)}{q_\theta(z)}
=
1.
\end{align*}
Thus, $p_\theta^{\mathrm{imp}}(x\mid z)$ is a normalized conditional distribution. For discrete observations, the corresponding integrals are replaced by sums.

Moreover, for any $x$ in the support of $p_{\mathrm{data}}$ and any $z$ with $q_\theta(z)>0$, direct substitution gives
\begin{align*}
\frac{
p_\theta^{\mathrm{imp}}(x\mid z)
q_\theta(z)
}{
p_{\mathrm{data}}(x)
}
&=
\frac{
p_{\mathrm{data}}(x)
q_\theta(z\mid x)
}{
q_\theta(z)
}
\frac{
q_\theta(z)
}{
p_{\mathrm{data}}(x)
}
\\
&=
q_\theta(z\mid x).
\end{align*}
Hence $q_\theta(z\mid x)$ is exactly the posterior associated with the prior $q_\theta(z)$ and the implicit emission $p_\theta^{\mathrm{imp}}(x\mid z)$.

Equivalently, the resulting one-time joint distribution satisfies
\begin{equation*}
p_\theta^{\mathrm{imp}}(x\mid z)q_\theta(z)
=
p_{\mathrm{data}}(x)q_\theta(z\mid x).
\end{equation*}
Its observation marginal is $p_{\mathrm{data}}(x)$, so the construction defines a valid static latent-variable model. As emphasized in the main text, this one-time Bayes completion does not by itself establish a sequence-level HMM; that stronger result additionally requires transition, marginal, and filtering consistency.

\subsection{Proof of Proposition~\ref{prop:predictive-sufficiency-markov}}
\label{app:proof-predictive-sufficiency}

By assumption, the future target state is generated as
\begin{equation*}
Z_{t+1}
=
g_{\bar\theta}(X_{t+1},U_{t+1}),
\end{equation*}
so $Z_{t+1}$ is a stochastic post-processing of $(X_{t+1},U_{t+1})$. The conditional data-processing inequality therefore gives
\begin{equation*}
\MI(Z_{t+1};X_{<t}\mid Z_t)
\leq
\MI(X_{t+1},U_{t+1};X_{<t}\mid Z_t).
\end{equation*}

By the chain rule for conditional mutual information,
\begin{align*}
\MI(X_{t+1},U_{t+1};X_{<t}\mid Z_t)
={}&
\MI(X_{t+1};X_{<t}\mid Z_t)
\\
&+
\MI(
U_{t+1};
X_{<t}
\mid
X_{t+1},Z_t
).
\end{align*}

Predictive sufficiency,
\begin{equation*}
X_{>t}
\perp
X_{\leq t}
\mid
Z_t,
\end{equation*}
implies
\begin{equation*}
\MI(X_{t+1};X_{<t}\mid Z_t)
=
0,
\end{equation*}
because $X_{t+1}$ is contained in $X_{>t}$ and $X_{<t}$ is contained in $X_{\leq t}$.

Likewise, the assumed conditional independence of the target-encoder randomness,
\begin{equation*}
U_{t+1}
\perp
X_{\leq t}
\mid
(X_{t+1},Z_t),
\end{equation*}
implies
\begin{equation*}
\MI(
U_{t+1};
X_{<t}
\mid
X_{t+1},Z_t
)
=
0.
\end{equation*}
Consequently,
\begin{equation*}
\MI(X_{t+1},U_{t+1};X_{<t}\mid Z_t)
=
0.
\end{equation*}

Combining this equality with the conditional data-processing inequality and the nonnegativity of conditional mutual information yields
\begin{equation*}
\MI(Z_{t+1};X_{<t}\mid Z_t)
=
0,
\end{equation*}
which proves the proposition. Thus, under predictive sufficiency and the stated target-encoder independence condition, the current representation $Z_t$ screens off older observation history from the next latent state $Z_{t+1}$. For a deterministic target encoder, the auxiliary randomness $U_{t+1}$ can be omitted.

\section{Three Realizations of the Emission Direction}
\label{app:emission-realizations}

The main text describes three alternative ways to complete the state-to-observation direction of the latent-state model. These constructions should not be interpreted as three progressively stronger notions of equivalence. Rather, each can supply the emission component required for an emission-complete latent-state representation. Exact sequence-level HMM equivalence additionally requires the transition, marginal-consistency, and filtering-consistency conditions developed in \cref{app:hmm-equivalence}.

\subsection{Explicit decoder}

The most direct realization introduces a probabilistic decoder
\begin{equation*}
p_\psi(X_t\mid Z_t),
\end{equation*}
which has the same state-to-observation direction as an HMM emission model.

When both the latent state and observation space are finite, this conditional may be represented by an emission matrix $B$, for example
\begin{equation*}
B_{jk}
=
p_\psi(X_t=k\mid Z_t=j).
\end{equation*}
For a categorical latent state with continuous observations, each latent state instead indexes an observation density. More generally, for images, signals, or other high-dimensional observations, $p_\psi(X_t\mid Z_t)$ may be parameterized by a Gaussian, discretized logistic, autoregressive, diffusion-based, or other suitable conditional observation model.

If the decoder is fitted only after JEPA representation learning while the latent model is held fixed, it acts as a post-hoc observation model or probe. If it participates jointly in training, it becomes part of the generative latent-state model and can contribute directly to observation-sequence likelihood.

\subsection{Invertible target encoder}

A second realization is available when the target encoder
\begin{equation*}
f_{\bar\theta}:\cX\rightarrow\cS
\end{equation*}
is bijective on the modeled data domain. Its target representation satisfies
\begin{equation*}
Z_t^{\mathrm T}
=
f_{\bar\theta}(X_t),
\qquad
X_t
=
f_{\bar\theta}^{-1}(Z_t^{\mathrm T}).
\end{equation*}
Identifying the latent state with this target representation gives the deterministic state-to-observation kernel
\begin{equation*}
p(X_t\mid Z_t)
=
\delta\!\left(
X_t-f_{\bar\theta}^{-1}(Z_t)
\right).
\end{equation*}
Thus, invertibility supplies the required state-to-observation direction without introducing a separate decoder.

A deterministic inverse should nevertheless be distinguished from a non-degenerate probabilistic emission. If $f_{\bar\theta}$ is a tractable invertible density model, the change-of-variables formula can be used to evaluate the observation density induced by a latent density. The conditional map $X_t=f_{\bar\theta}^{-1}(Z_t)$ itself remains deterministic, however. A non-degenerate conditional emission can instead be obtained by augmenting the inverse map with an observation-noise model, for example
\begin{equation*}
X_t
=
f_{\bar\theta}^{-1}(Z_t)
+
\varepsilon_t,
\end{equation*}
with a specified noise distribution for $\varepsilon_t$.

Exact invertibility imposes strong architectural constraints. In particular, it prevents unrestricted dimensionality reduction and may require the representation to preserve observation details that a predictive information bottleneck would otherwise discard. Standard compressed JEPA encoders therefore need not admit this construction.

\subsection{Implicit emission}

When no explicit decoder is parameterized and the target encoder is not invertible, a local stochastic encoder can still induce a state-to-observation conditional. As defined in \cref{eq:implicit-emission},
\begin{equation*}
p_\theta^{\mathrm{imp}}(x\mid z)
=
\frac{
p_{\mathrm{data}}(x)
q_\theta(z\mid x)
}{
q_\theta(z)
},
\end{equation*}
for $q_\theta(z)>0$. As shown in the preceding proof, this conditional is normalized and, together with $q_\theta(z)$, reproduces the one-time joint distribution
\begin{equation*}
p_\theta^{\mathrm{imp}}(x\mid z)q_\theta(z)
=
p_{\mathrm{data}}(x)q_\theta(z\mid x).
\end{equation*}

This construction provides an exact static probabilistic completion of the observation--state relationship, but it does not automatically provide a practical generative model. In particular, $p_\theta^{\mathrm{imp}}(x\mid z)$ depends explicitly on the generally unknown data marginal $p_{\mathrm{data}}(x)$, so direct sampling and likelihood evaluation may be intractable.

Moreover, the construction uses a \emph{local} encoder $q_\theta(z\mid x)$. An arbitrary history-dependent encoder $q_\theta(Z_t\mid X_{\leq t})$ cannot simply be reinterpreted as an emission model. To obtain an exact sequence-level HMM from the implicit construction, the resulting one-time conditionals must additionally be consistent with the latent transition and with the Bayesian filtering recursion, as formalized in \cref{app:hmm-equivalence}.

\section{HMM-Style Training Objectives for PIB-VJEPA}
\label{app:hmm-style-training}

The HMM correspondence suggests an alternative to purely latent-space JEPA training. Once a valid observation model is available, the latent transition and emission can be trained from observation-sequence likelihood, while the history-dependent context encoder can be aligned with the corresponding Bayesian filtering distribution. This appendix summarizes the probabilistic objectives underlying the model-and-objective correspondence developed in the main text.

\subsection{Sequence likelihood}

Suppose that the latent-state model is equipped with an initial-state distribution $p_0(Z_1)$, transition model
\begin{equation*}
p_\phi(Z_{t+1}\mid Z_t,\xi_t),
\end{equation*}
and explicit emission model
\begin{equation*}
p_\psi(X_t\mid Z_t).
\end{equation*}
The resulting conditional sequence model is
\begin{equation*}
\begin{aligned}
&p_{\phi,\psi}
\left(
X_{1:T},Z_{1:T}
\mid
\xi_{1:T-1}
\right)
\\
&\qquad=
p_0(Z_1)
\prod_{t=1}^{T}
p_\psi(X_t\mid Z_t)
\prod_{t=1}^{T-1}
p_\phi(Z_{t+1}\mid Z_t,\xi_t).
\end{aligned}
\end{equation*}
Marginalizing the latent trajectory gives the observation-sequence evidence
\begin{equation*}
p_{\phi,\psi}
\left(
X_{1:T}\mid\xi_{1:T-1}
\right)
=
\int
p_{\phi,\psi}
\left(
X_{1:T},Z_{1:T}
\mid
\xi_{1:T-1}
\right)
\,dZ_{1:T}.
\end{equation*}
Training from this evidence directly constrains the transition--emission model in observation space, in contrast to the standard JEPA objective, which is imposed primarily in latent space.

\subsection{Exact likelihood and filtering for categorical MCJEPA}

For categorical states $Z_t\in\{1,\ldots,K\}$, the sequence likelihood and filtering posterior can be evaluated exactly by the HMM forward recursion. Let
\begin{equation*}
\pi_j
=
p_0(Z_1=j),
\qquad
b_\psi(X_t\mid j)
=
p_\psi(X_t\mid Z_t=j),
\end{equation*}
and
\begin{equation*}
(A_t)_{ij}
=
p_\phi(Z_{t+1}=j\mid Z_t=i,\xi_t).
\end{equation*}
For the time-homogeneous MCJEPA model, $A_t=A$.

Define the forward message
\begin{equation*}
\alpha_t(j)
=
p_{\phi,\psi}
\left(
X_{1:t},Z_t=j
\mid
\xi_{1:t-1}
\right).
\end{equation*}
It satisfies
\begin{equation*}
\alpha_1(j)
=
\pi_j b_\psi(X_1\mid j)
\end{equation*}
and
\begin{equation*}
\alpha_{t+1}(j)
=
b_\psi(X_{t+1}\mid j)
\sum_{i=1}^{K}
\alpha_t(i)(A_t)_{ij}.
\end{equation*}
The sequence evidence is therefore
\begin{equation*}
p_{\phi,\psi}
\left(
X_{1:T}\mid\xi_{1:T-1}
\right)
=
\sum_{j=1}^{K}\alpha_T(j).
\end{equation*}
In practice, the recursion is evaluated in log space or with normalized forward messages for numerical stability.

Normalizing the forward messages also gives the exact filtering posterior
\begin{equation*}
\widetilde q_t
=
p_{\phi,\psi}
\left(
Z_t\mid X_{\leq t},\xi_{<t}
\right).
\end{equation*}
The history-dependent PIB-VJEPA context encoder can then be trained as an amortized filter using
\begin{equation*}
\mathcal L_{\mathrm{filter}}
=
\E_t
\left[
\KL\left(
\sg(\widetilde q_t)
\,\middle\|\,
q_\theta(Z_t\mid X_{\leq t})
\right)
\right].
\end{equation*}
Thus, the encoder learns to approximate in one forward pass the current-state posterior that the HMM computes recursively.

This distinction is important: sequence likelihood trains the transition and emission model, whereas filtering distillation trains the context encoder to reproduce the corresponding filtering belief. Sequence likelihood alone does not require an independently parameterized history encoder to equal that filter.

\subsection{Continuous-state extension}

For continuous or nonlinear latent states, exact marginalization of $Z_{1:T}$ is generally unavailable. Introducing an approximate sequence posterior
\begin{equation*}
q_\eta
\left(
Z_{1:T}\mid X_{1:T},\xi_{1:T-1}
\right)
\end{equation*}
gives the standard variational lower bound
\begin{equation*}
\begin{aligned}
\log p_{\phi,\psi}
\left(
X_{1:T}\mid\xi_{1:T-1}
\right)
\geq
\E_{q_\eta}\Bigg[
&
\sum_{t=1}^{T}
\log p_\psi(X_t\mid Z_t)
+
\log p_0(Z_1)
\\
&+
\sum_{t=1}^{T-1}
\log p_\phi(Z_{t+1}\mid Z_t,\xi_t)
-
\log q_\eta
\left(
Z_{1:T}\mid X_{1:T},\xi_{1:T-1}
\right)
\Bigg].
\end{aligned}
\end{equation*}
The approximate posterior may be causal when online filtering is required or smoothing when full-sequence information is available during training. We include this extension to show how the same model-and-objective interpretation extends beyond the finite categorical setting; the experiments in this paper use exact finite-state inference.

\subsection{Relation to JEPA and hybrid training}

HMM-style sequence learning and JEPA latent prediction are distinct objectives even when they operate on the same latent-state architecture. In the categorical setting studied in Experiment~4, the main text compares
\begin{equation*}
\underbrace{
\mathcal L_{\mathrm{MC}}
}_{\text{JEPA latent prediction}}
\qquad\text{and}\qquad
\underbrace{
\mathcal L_{\mathrm{HMM}}
+
\mathcal L_{\mathrm{filter}}
}_{\text{HMM-style training}},
\end{equation*}
with $\mathcal L_{\mathrm{state}}$ used as a shared state-use regularizer. The hybrid regime combines all three signals,
\begin{equation*}
\mathcal L_{\mathrm{HMM}}
+
\mathcal L_{\mathrm{MC}}
+
\mathcal L_{\mathrm{filter}},
\end{equation*}
with the corresponding weights given in \cref{sec:exp-hmm-training,app:experimental-details}.

Importantly, the exact HMM filtering posterior is used only as the target of $\mathcal L_{\mathrm{filter}}$. It does not replace the EMA future target in $\mathcal L_{\mathrm{MC}}$: the JEPA component retains the same target-encoder construction as the JEPA-only regime. The transition matrix is shared between the HMM sequence objective and the MCJEPA latent-prediction objective, so the two training signals constrain the same latent dynamics from observation-space and representation-space perspectives, respectively.

Consequently, adding an emission model alone does not make PIB-VJEPA training identical to HMM training. Model-and-objective equivalence additionally requires observation-sequence evidence and the corresponding sequence-inference semantics to participate in the learning objective.

\section{Exact HMM Representation Conditions}
\label{app:hmm-equivalence}

This appendix makes precise the sufficient conditions in \cref{thm:exact-hmm-representation}. We state the construction for a discrete latent state for clarity; the same argument extends to general Markov kernels by replacing sums with integrals. Side information $\xi_t$ is treated as observed, so all sequence distributions below are conditional on $\xi_{1:T-1}$.

Let the latent transition be
\begin{equation*}
A_t(i,j)
=
p_\phi(Z_{t+1}=j\mid Z_t=i,\xi_t),
\end{equation*}
and let
\begin{equation*}
b_t(x\mid j)
\end{equation*}
denote a valid state-to-observation conditional. This emission may be supplied by an explicit decoder, an invertible target encoder interpreted as a deterministic kernel, or the implicit construction described below. Given an initial distribution $\rho_1$, these components define
\begin{equation}
\begin{aligned}
p(
z_{1:T},x_{1:T}
\mid
\xi_{1:T-1}
)
\qquad=
\rho_1(z_1)
\prod_{t=1}^{T}
b_t(x_t\mid z_t)
\prod_{t=1}^{T-1}
A_t(z_t,z_{t+1}),
\end{aligned}
\label{eq:exact-hmm-factorization-app}
\end{equation}
which is the conditional HMM factorization.

\subsection{Marginal and filtering consistency}

Let
\begin{equation*}
\rho_t(j)
=
p(Z_t=j\mid\xi_{1:t-1})
\end{equation*}
denote the latent marginal before observing $X_t$, conditional on the side-information history. Dynamic consistency requires
\begin{equation}
\rho_{t+1}(j)
=
\sum_i
\rho_t(i)A_t(i,j).
\label{eq:latent-marginal-consistency}
\end{equation}
Thus, the one-time latent marginals must be generated by the same transition kernel used by the temporal model.

For a realized observation history, let
\begin{equation*}
q_t(j)
=
p(Z_t=j\mid X_{\leq t},\xi_{<t})
\end{equation*}
denote the filtering distribution. Its transition-based predictive prior is
\begin{equation*}
\pi_t(j)
=
\sum_i
q_{t-1}(i)A_{t-1}(i,j),
\end{equation*}
with $\pi_1=\rho_1$. Bayes' rule then gives the filtering recursion
\begin{equation}
q_t(j)
=
\frac{
b_t(X_t\mid j)\pi_t(j)
}{
\sum_k
b_t(X_t\mid k)\pi_t(k)
}.
\label{eq:filtering-update}
\end{equation}

The history-dependent PIB-VJEPA encoder is filtering-consistent when
\begin{equation}
q_\theta(Z_t\mid X_{\leq t},\xi_{<t})
=
q_t
\label{eq:encoder-filter-consistency}
\end{equation}
for the transition and emission model under consideration. An arbitrary history encoder need not satisfy this equality.

\subsection{Explicit and invertible emissions}

If a decoder directly specifies
\begin{equation*}
b_t(x\mid z)
=
p_\psi(x\mid z),
\end{equation*}
then \cref{eq:exact-hmm-factorization-app} follows immediately from the initial distribution, first-order transition, and emission model. If the target encoder is invertible, the deterministic kernel induced by
\begin{equation*}
x=f_{\bar\theta}^{-1}(z)
\end{equation*}
plays the same role. In either case, if the latent marginals satisfy \cref{eq:latent-marginal-consistency} and the history encoder satisfies \cref{eq:encoder-filter-consistency}, the resulting temporal JEPA admits the sequence-level HMM interpretation stated in \cref{thm:exact-hmm-representation}.

\subsection{Implicit-emission case}

The less direct case begins with a local evidence encoder
\begin{equation*}
e_{\theta,t}(z\mid x)
\end{equation*}
that depends only on the current observation. Let $p_t(x)$ be the one-time observation marginal and define its induced latent marginal
\begin{equation}
\rho_t(z)
=
\int
p_t(x)e_{\theta,t}(z\mid x)
\,dx.
\label{eq:implicit-local-marginal}
\end{equation}
For $\rho_t(z)>0$, define
\begin{equation*}
b_t^{\mathrm{imp}}(x\mid z)
=
\frac{
p_t(x)e_{\theta,t}(z\mid x)
}{
\rho_t(z)
}.
\end{equation*}
By the argument in the proof of implicit emission completion, this is a normalized state-to-observation conditional and satisfies
\begin{equation*}
e_{\theta,t}(z\mid x)
=
\frac{
b_t^{\mathrm{imp}}(x\mid z)\rho_t(z)
}{
p_t(x)
}.
\end{equation*}

The locality assumption is important: $e_{\theta,t}(z\mid x_t)$ supplies the observation-dependent evidence factor, whereas the history-dependent context encoder represents the filtering belief. The two should not be identified.

Substituting the implicit emission into the filtering recursion gives
\begin{align*}
q_t(z)
&\propto
\pi_t(z)b_t^{\mathrm{imp}}(x_t\mid z)
\\
&=
\pi_t(z)
\frac{
p_t(x_t)e_{\theta,t}(z\mid x_t)
}{
\rho_t(z)
}
\\
&\propto
\pi_t(z)
\frac{
e_{\theta,t}(z\mid x_t)
}{
\rho_t(z)
}.
\end{align*}
Hence the history-dependent filtering distribution may equivalently be written as
\begin{equation}
q_t(z)
\propto
\pi_t(z)
\frac{
e_{\theta,t}(z\mid x_t)
}{
\rho_t(z)
}.
\label{eq:implicit-filtering-update}
\end{equation}

For an exact sequence-level interpretation, the induced marginals in \cref{eq:implicit-local-marginal} must additionally be dynamically consistent with the transition:
\begin{equation}
\rho_{t+1}(z')
=
\sum_z
\rho_t(z)A_t(z,z').
\label{eq:implicit-marginal-consistency}
\end{equation}
Finally, the PIB-VJEPA history encoder must coincide with the filtering distribution generated by \cref{eq:implicit-filtering-update}.

\subsection{Completion of the proof for Theorem.\ref{thm:exact-hmm-representation}}

We can now verify the four sufficient conditions in \cref{thm:exact-hmm-representation}. First, $A_t$ defines first-order latent Markov dynamics. Second, one of the three constructions above supplies a valid state-to-observation conditional. Third, \cref{eq:latent-marginal-consistency}, or \cref{eq:implicit-marginal-consistency} in the implicit case, ensures that the latent marginals evolve under the same transition kernel. Fourth, \cref{eq:encoder-filter-consistency} identifies the history-dependent context encoder with the Bayesian filtering posterior of that transition--emission model.

Therefore the joint sequence distribution is precisely \cref{eq:exact-hmm-factorization-app}, and the context encoder represents its filtering distribution. This proves the sufficient-condition statement in \cref{thm:exact-hmm-representation}.

These conditions are stronger than architectural correspondence alone. In particular, a valid transition and emission specify an HMM-compatible generative model, but an arbitrary JEPA history encoder need not equal its Bayesian filter, and independently induced one-time latent marginals need not evolve according to the learned transition. The additional consistency conditions are what promote an emission-complete latent-state representation to exact sequence-level HMM equivalence.

\section{Experimental Details}
\label{app:experimental-details}

This appendix provides the data-generation procedures, model architectures, optimization settings, evaluation metrics, and supplementary results for the experiments in \cref{sec:experiments}. The experiments are deliberately small and synthetic because their purpose is to isolate the structural claims of the paper under known latent dynamics rather than to benchmark large-scale forecasting performance.

Unless otherwise stated, experiments involving sampled data or learned models use five random seeds,
\begin{equation*}
\{0,1,2,3,4\},
\end{equation*}
and report mean $\pm$ one standard deviation across seeds. Exact finite calculations in Experiment~3, such as deterministic-partition enumeration, are deterministic and are therefore reported without seed variability. We do not perform formal hypothesis tests; the experiments are intended as controlled structural diagnostics, and across-seed variability is reported to expose sampling and optimization variability.

For the finite-HMM experiments, the paper configuration uses $400$ training sequences and $160$ test sequences of length $80$, with prediction horizons
\begin{equation*}
\mathcal H=\{1,2,4,8\}.
\end{equation*}
No separate validation split is used because hyperparameter selection is not the purpose of these controlled diagnostics. Within a seed, competing methods are evaluated on the same generated data whenever a paired comparison is intended. Experiment~4 independently regenerates the separated-emission data used in Experiment~1 with the same data-generating process and seeds, but none of the fitted Experiment~1 models is reused.

\subsection{Evaluation metrics}
\label{app:experimental-metrics}

We collect here the evaluation metrics used across the experiments. This also separates \emph{representation recovery}, \emph{transition recovery}, \emph{predictive performance}, and \emph{probabilistic-model fit}, which measure different aspects of the proposed correspondence.

\paragraph{State recovery: ARI and NMI.}
When ground-truth latent states are available, we convert each learned categorical distribution $q_t$ to a hard state assignment
\begin{equation*}
\widehat S_t
=
\arg\max_k q_t(k).
\end{equation*}
We report the adjusted Rand index (ARI) and normalized mutual information (NMI) between the learned assignments $\widehat S_t$ and ground-truth states $S_t$.

For a contingency table with entries $n_{ij}$, row sums $a_i$, column sums $b_j$, and total sample size $N$, ARI is
\begin{equation} \label{eq:ARI}
\operatorname{ARI}
=
\frac{
\displaystyle
\sum_{ij}\binom{n_{ij}}{2}
-
\frac{
\left(\sum_i\binom{a_i}{2}\right)
\left(\sum_j\binom{b_j}{2}\right)
}{
\binom{N}{2}
}
}{
\displaystyle
\frac{1}{2}
\left[
\sum_i\binom{a_i}{2}
+
\sum_j\binom{b_j}{2}
\right]
-
\frac{
\left(\sum_i\binom{a_i}{2}\right)
\left(\sum_j\binom{b_j}{2}\right)
}{
\binom{N}{2}
}
}.
\end{equation}
ARI corrects the ordinary Rand index for agreement expected by chance. A value of $1$ denotes identical partitions, while values near $0$ correspond to chance-level agreement under the adjustment.

NMI is computed using the arithmetic normalization,
\begin{equation} \label{eq:NMI}
\operatorname{NMI}(S,\widehat S)
=
\frac{
2 I(S;\widehat S)
}{
H(S)+H(\widehat S)
}.
\end{equation}
NMI lies in $[0,1]$, with larger values indicating greater shared information between the learned and ground-truth state partitions. Both ARI and NMI are invariant to permutation of categorical state labels.

\paragraph{Permutation alignment.}
Although ARI and NMI do not require label alignment, transition matrices and predicted categorical probabilities do. We therefore compute a Hungarian assignment on the \emph{training-set} hard state assignments. Let $M$ denote the resulting permutation matrix mapping learned-state order to ground-truth-state order. A learned transition matrix $\widehat A$ is aligned as
\begin{equation*}
\widehat A_{\mathrm{aligned}}
=
M^\top \widehat A M.
\end{equation*}
The mapping is fitted only on training assignments and then held fixed for test evaluation.

\paragraph{Transition recovery.}
When the true transition matrix $A^\star$ is known, we measure normalized Frobenius error,
\begin{equation} \label{eq:transition_matrix_recovery}
\mathcal E_A
=
\frac{
\left\|
\widehat A_{\mathrm{aligned}}-A^\star
\right\|_F
}{
\left\|A^\star\right\|_F
}.
\end{equation}
Lower values indicate more faithful recovery of the underlying Markov transition law. This metric evaluates the learned dynamics themselves rather than only their downstream predictive consequences.

\paragraph{True-state prediction NLL.}
For prediction horizon $h$, let $\widehat q_{t+h}$ denote the predicted categorical distribution after alignment to ground-truth state order. We report
\begin{equation}
\mathcal L_h^{\mathrm{state}}
=
-\E_t
\left[
\log
\widehat q_{t+h}(S_{t+h})
\right].
\label{eq:state_prediction_nll}
\end{equation}
Thus the metric measures the probability assigned to the actual future latent state. Lower values are better. For the shared-transition models,
\begin{equation*}
\widehat q_{t+h}
=
q_tA^h.
\end{equation*}

\paragraph{Observation-sequence NLL.}
For models equipped with a transition--emission likelihood, observation-space fit is measured by negative log-likelihood per time step,
\begin{equation}
\mathcal L_{\mathrm{seq}}
=
-\frac{1}{T}
\E
\left[
\log p_{\phi,\psi}(X_{1:T})
\right],
\label{eq:sequence_nll}
\end{equation}
where $p_{\phi,\psi}(X_{1:T})$ denotes the marginal observation-sequence density induced by the latent transition and emission models after marginalizing the latent-state sequence. In the finite-state setting used in our experiments,
\begin{equation*}
p_{\phi,\psi}(X_{1:T})
=
\sum_{Z_{1:T}}
p_\phi(Z_1)
\left[
\prod_{t=1}^{T}
p_\psi(X_t\mid Z_t)
\right]
\left[
\prod_{t=1}^{T-1}
p_\phi(Z_{t+1}\mid Z_t)
\right].
\end{equation*}
Thus, the likelihood integrates out the unobserved latent trajectory rather than conditioning on the ground-truth latent states. In practice, this marginalization is evaluated exactly and efficiently by the HMM forward algorithm rather than by explicitly enumerating all possible latent-state sequences.

Unlike state-prediction NLL, this metric evaluates the probability density assigned to the observed sequence rather than the probability assigned to the known synthetic latent state. In Experiment~4, the same mathematical quantity plays different roles across training regimes. For the HMM+filter and hybrid regimes, it is optimized during training as $\mathcal L_{\mathrm{HMM}}$ and subsequently evaluated on held-out sequences. For the JEPA-only regime, no observation-sequence likelihood is optimized during representation learning; $\mathcal L_{\mathrm{seq}}$ is computed only after fitting the post-hoc emission model. It therefore serves as a common evaluation metric across the three regimes rather than a common training objective.

\paragraph{Path disagreement.}
To measure whether direct and composed multi-step predictions agree, we use
\begin{equation}
\mathcal D_{\mathrm{path}}(h_1,h_2)
=
\E_t
\left[
\left\|
q_tA_{h_1+h_2}
-
(q_tA_{h_1})A_{h_2}
\right\|_1
\right].
\label{eq:path_disagreement}
\end{equation}
For MCJEPA with one shared transition matrix,
\begin{equation*}
A_h=A^h,
\end{equation*}
and therefore
\begin{equation*}
\mathcal D_{\mathrm{path}}(h_1,h_2)=0
\end{equation*}
algebraically, up to numerical precision. For independently learned horizon-specific matrices no such guarantee exists.

\paragraph{Filtering KL.}
When both an exact model-based filtering distribution and an amortized context-encoder distribution are available, we report
\begin{equation}
\mathcal D_{\mathrm{filter}}
=
\E_t
\left[
\KL
\left(
q_t^{\mathrm{exact}}
\,\middle\|\,
q_t^{\mathrm{enc}}
\right)
\right].
\label{eq:filtering_kl}
\end{equation}
Lower values mean that the amortized encoder more closely reproduces the corresponding filtering belief. In Experiment~4 this quantity is training aligned for the HMM+filter and hybrid regimes because both explicitly optimize filtering distillation; it is therefore interpreted as a role diagnostic rather than an independent generalization metric.

\paragraph{Accuracy, Brier score, and entropy.}
Experiment~2 additionally reports state accuracy,
\begin{equation} \label{eq:accuracy}
\operatorname{Acc}
=
\frac{1}{N}
\sum_t
\mathbf{1}
\left\{
\arg\max_k q_t(k)=S_t
\right\},
\end{equation}
the multiclass Brier score,
\begin{equation} \label{eq:brier_score}
\operatorname{Brier}
=
\frac{1}{N}
\sum_t
\sum_k
\left(
q_t(k)-\mathbf{1}\{S_t=k\}
\right)^2,
\end{equation}
and mean posterior entropy,
\begin{equation} \label{eq:mean_posterior_entropy}
\overline H
=
\frac{1}{N}
\sum_t H(q_t).
\end{equation}
Accuracy measures hard classification correctness, whereas NLL and Brier score retain information about probabilistic confidence. Posterior entropy is descriptive and should not be interpreted as a performance metric by itself.

\subsection{Experiment 1: finite-HMM recovery and Markov composition}
\label{app:exp1-details}

\paragraph{Data generation.}
We generate observations from a four-state stationary Gaussian HMM. The ground-truth transition matrix is
\begin{equation*}
A^\star
=
\begin{bmatrix}
0.85 & 0.10 & 0.05 & 0.00\\
0.05 & 0.85 & 0.10 & 0.00\\
0.00 & 0.05 & 0.85 & 0.10\\
0.10 & 0.00 & 0.05 & 0.85
\end{bmatrix}.
\end{equation*}
The initial state is sampled from the stationary distribution of $A^\star$. Conditional on state $S_t=k$, the two-dimensional observation is generated as
\begin{equation*}
X_t\mid S_t=k
\sim
\mathcal N(\mu_k,\sigma^2 I_2),
\end{equation*}
with state means
\begin{equation*}
\mu_1=(-1,-1),\qquad
\mu_2=(-1,1),\qquad
\mu_3=(1,1),\qquad
\mu_4=(1,-1).
\end{equation*}
We consider two emission regimes:
\begin{equation*}
\sigma=0.35
\quad\text{(separated)},
\qquad
\sigma=0.80
\quad\text{(ambiguous)}.
\end{equation*}
For each seed and regime we independently generate $400$ training sequences and $160$ test sequences, each of length $80$.

\paragraph{MCJEPA architecture.}
The online context encoder is a one-layer GRU with hidden dimension $48$, followed by a linear projection to $K=4$ logits and a softmax:
\begin{equation*}
q_\theta(Z_t\mid X_{\leq t})
=
\operatorname{softmax}
\left(
Wh_t+b
\right).
\end{equation*}
The target encoder has the same architecture but processes each $X_t$ as an independent length-one sequence, producing a local target distribution. Its parameters are initialized from the online encoder and subsequently updated by exponential moving average.

For the shared-transition MCJEPA model, a trainable $4\times4$ logit matrix is row-normalized by softmax,
\begin{equation*}
A=\operatorname{softmax}_{\mathrm{row}}(L_A),
\end{equation*}
and horizon-$h$ prediction uses
\begin{equation*}
\widehat q_{t+h}=q_tA^h.
\end{equation*}
The transition logits are initialized with a mild diagonal bias,
\begin{equation*}
L_A=I_4.
\end{equation*}

The horizon-specific baseline uses the same online and target encoders but replaces the shared matrix with independent row-stochastic matrices
\begin{equation*}
A_1,\quad A_2,\quad A_4,\quad A_8.
\end{equation*}
Each matrix is initialized with the same diagonal logit bias, but no constraint requires
\begin{equation*}
A_h=A_1^h.
\end{equation*}

\paragraph{Warm start and optimization.}
To make the small synthetic recovery experiment stable and reproducible, both categorical JEPA variants receive an unsupervised K-means warm start. K-means with $K=4$ and $10$ initializations is fitted to individual training observations; ground-truth states are never used. The online encoder is then trained for $100$ warm-start updates with Adam at learning rate
\begin{equation*}
10^{-2}
\end{equation*}
to predict the K-means assignments, after which the target encoder is copied from the online encoder.

The main MCJEPA optimization runs for $180$ epochs with mini-batches of $64$ sequences and Adam learning rate
\begin{equation*}
3\times10^{-3}.
\end{equation*}
The prediction loss averages the target-to-prediction KL divergence over $\mathcal H=\{1,2,4,8\}$:
\begin{equation*}
\mathcal L_{\mathrm{pred}}
=
\frac{1}{|\mathcal H|}
\sum_{h\in\mathcal H}
\E
\left[
\KL
\left(
\bar q_{t+h}
\,\middle\|\,
q_tA_h
\right)
\right],
\end{equation*}
where $A_h=A^h$ for MCJEPA and $A_h$ is independently learned for the horizon-specific baseline.

The state-use terms are
\begin{align*}
\mathcal L_{\mathrm{occ}}
&=
\KL
\left(
\bar q
\middle\|
\operatorname{Unif}(K)
\right),
\\
\mathcal L_{\mathrm{ent}}
&=
\E_t
\left[
H(q_t)
\right],
\end{align*}
with
\begin{equation*}
\lambda_{\mathrm{occ}}=0.30,
\qquad
\lambda_{\mathrm{ent}}=0.02.
\end{equation*}
Because $\mathcal L_{\mathrm{ent}}$ is minimized, it encourages confident per-example assignments. The target encoder uses EMA coefficient
\begin{equation*}
\tau=0.995.
\end{equation*}
Gradients are clipped to norm $5$.

\paragraph{Gaussian-HMM baseline.}
The HMM baseline uses the correctly specified four-state family with a learned initial-state distribution, row-stochastic transition matrix, and state-conditional diagonal Gaussian emissions. The emission means are initialized from K-means cluster centers, and the diagonal variances are initialized from within-cluster variances with a small additive floor.

The HMM is trained directly through the observation-sequence NLL defined above, computed exactly by the forward algorithm. We use Adam with learning rate
\begin{equation*}
3\times10^{-2}
\end{equation*}
for $350$ optimization steps and clip gradients to norm $10$.

\paragraph{Permutation alignment and evaluation.}
The Hungarian alignment and common metrics follow \cref{app:experimental-metrics}. ARI and NMI evaluate recovery of the hidden-state partition, transition error evaluates recovery of $A^\star$, and true-state NLL evaluates future-state prediction at
\begin{equation*}
h\in\{1,2,4,8\}.
\end{equation*}

For path consistency we evaluate
\begin{equation*}
(h_1,h_2)\in\{(1,1),(2,2),(4,4)\}.
\end{equation*}
For the shared-transition model,
\begin{equation*}
q_tA^{h_1+h_2}
=
(q_tA^{h_1})A^{h_2}
\end{equation*}
exactly, so path disagreement is zero by construction. The horizon-specific baseline has no corresponding constraint.

\paragraph{State-usage metrics.}
In addition to the common evaluation metrics, we monitor both soft and hard effective state counts. If
\begin{equation*}
\bar q
=
\frac{1}{N}
\sum_n q_n
\end{equation*}
is the average soft assignment distribution, then
\begin{equation*}
K_{\mathrm{eff}}^{\mathrm{soft}}
=
\exp\left(H(\bar q)\right).
\end{equation*}
For hard assignments, let $\widehat p_k$ be the empirical frequency of state $k$. We define
\begin{equation*}
K_{\mathrm{eff}}^{\mathrm{hard}}
=
\exp\left(
-\sum_{k:\widehat p_k>0}
\widehat p_k\log\widehat p_k
\right).
\end{equation*}
The reported assignment entropy is
\begin{equation*}
\overline H_{\mathrm{assign}}
=
\E_t[H(q_t)].
\end{equation*}

Effective state count measures diversity of state usage, whereas assignment entropy measures confidence of individual assignments. Low assignment entropy is not desirable by itself: an encoder that confidently maps every observation to one state also has low entropy. State-use diversity and assignment confidence must therefore be interpreted jointly.

\paragraph{Collapse ablations.}
The collapse diagnostic is run separately from the warm-started recovery experiment. We generate a new separated-emission dataset with $\sigma=0.35$ and train the shared-$A$ model from random initialization, deliberately omitting the K-means warm start. The four settings are
\begin{equation*}
(\lambda_{\mathrm{occ}},\lambda_{\mathrm{ent}})
\in
\left\{
(0,0),\,
(0.30,0),\,
(0,0.02),\,
(0.30,0.02)
\right\}.
\end{equation*}
Each model is trained for $180$ epochs with batch size $64$; for this diagnostic the EMA coefficient is $\tau=0.99$. This deliberately creates a more collapse-prone optimization problem and isolates the complementary roles of the two penalties: occupancy regularization discourages global state under-use, whereas entropy regularization encourages confident per-example assignments.

\begin{figure}[t]
\centering
\begin{minipage}[t]{0.49\linewidth}
\centering
\includegraphics[width=\linewidth]{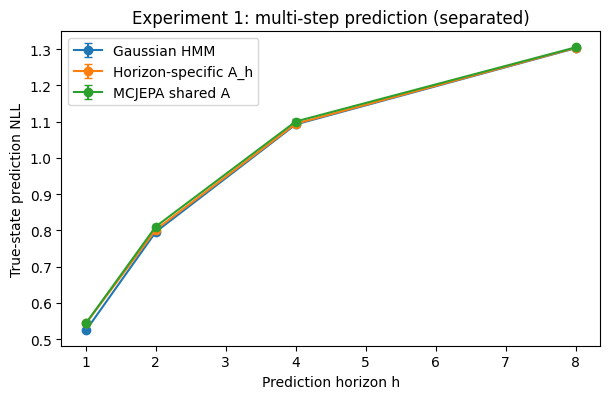}
\end{minipage}
\hfill
\begin{minipage}[t]{0.49\linewidth}
\centering
\includegraphics[width=\linewidth]{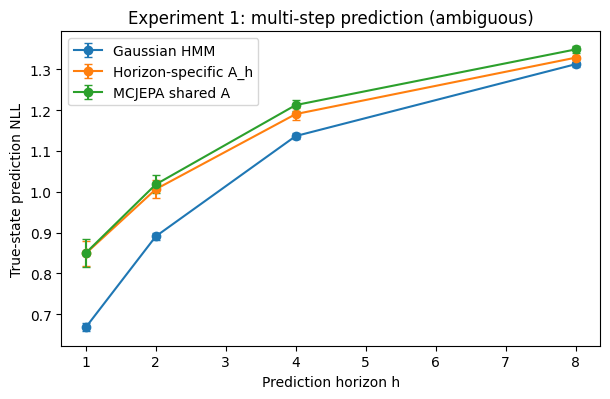}
\end{minipage}
\caption{Complete multi-horizon state-prediction results for Experiment~1. \emph{Left:} separated emissions. \emph{Right:} ambiguous emissions. Lower true-state NLL is better. Error bars denote mean $\pm$ one standard deviation over five seeds.}
\label{fig:app-exp1-multistep}
\end{figure}

\paragraph{Supplementary results.}
\Cref{fig:app-exp1-multistep} reports the complete multi-horizon prediction curves omitted from the main text. In the separated regime, all three models remain close across horizons. Under ambiguous emissions, the correctly specified HMM remains strongest, while the horizon-specific and shared-transition JEPA models exhibit similar predictive NLL despite their substantially different structural consistency.

\begin{figure}[t]
\centering
\includegraphics[width=0.52\linewidth]{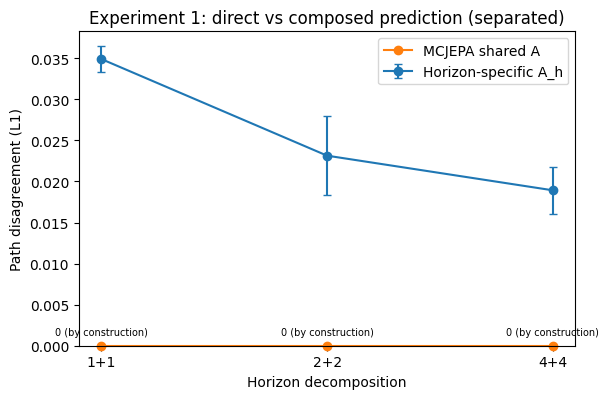}
\caption{Direct-versus-composed prediction disagreement in the separated-emission regime. MCJEPA has zero path disagreement by construction because all horizons are powers of one shared transition matrix.}
\label{fig:app-exp1-path}
\end{figure}

The separated-emission path-consistency result is shown in \cref{fig:app-exp1-path}. As in the ambiguous regime, the shared-$A$ model is exactly compositionally consistent, whereas independently trained horizon-specific matrices exhibit nonzero direct-versus-composed disagreement.

\begin{figure}[t]
\centering
\begin{minipage}[t]{0.49\linewidth}
\centering
\includegraphics[width=\linewidth]{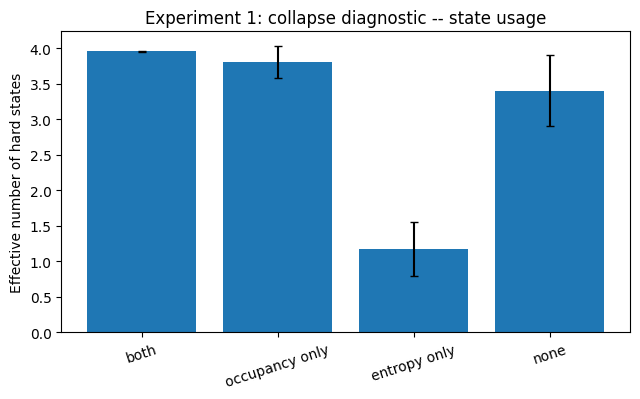}
\end{minipage}
\hfill
\begin{minipage}[t]{0.49\linewidth}
\centering
\includegraphics[width=\linewidth]{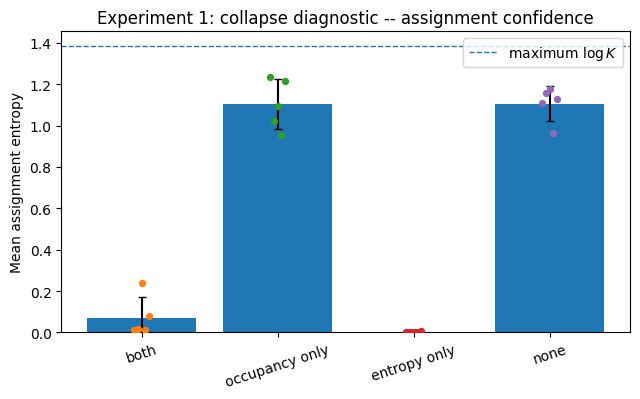}
\end{minipage}
\caption{Additional collapse diagnostics for Experiment~1. \emph{Left:} effective number of hard states. \emph{Right:} mean per-example assignment entropy. Occupancy regularization primarily promotes broad global state usage, whereas entropy regularization promotes confident assignments; neither diagnostic should be interpreted in isolation.}
\label{fig:app-exp1-collapse-extra}
\end{figure}

\Cref{fig:app-exp1-collapse-extra} supplements the main-text ARI ablation with state-usage and assignment-confidence diagnostics.

\subsection{Experiment 2: filtering under emission ambiguity}
\label{app:exp2-details}

\paragraph{Data generation.}
Experiment~2 uses a persistent two-state HMM with transition matrix
\begin{equation*}
A
=
\begin{bmatrix}
0.97 & 0.03\\
0.03 & 0.97
\end{bmatrix}.
\end{equation*}
Its stationary distribution is uniform,
\begin{equation*}
\pi=(0.5,0.5).
\end{equation*}
The scalar observation model is
\begin{equation*}
X_t\mid S_t
=
\begin{cases}
\mathcal N(-\mu,\sigma^2), & S_t=0,\\
\mathcal N(+\mu,\sigma^2), & S_t=1,
\end{cases}
\end{equation*}
with
\begin{equation*}
\sigma=1.
\end{equation*}
Emission ambiguity is controlled by
\begin{equation*}
\frac{\mu}{\sigma}
\in
\{2.0,1.25,0.75,0.45\}.
\end{equation*}
For every seed and separation value we generate $160$ sequences of length $80$, corresponding to $12{,}800$ state--observation pairs per setting. Because the comparison uses exact oracle posteriors, there is no learned train/test model split in this experiment; independent random seeds provide repeated sampled datasets.

\paragraph{Exact local evidence.}
The local posterior uses only the current observation and the stationary state prior:
\begin{equation*}
q_t^{\mathrm{local}}(k)
=
p(S_t=k\mid X_t)
=
\frac{
\pi_k p(X_t\mid S_t=k)
}{
\sum_j
\pi_j p(X_t\mid S_t=j)
}.
\end{equation*}
This is the oracle counterpart of a local encoder $q_\theta(Z_t\mid X_t)$.

\paragraph{Exact filtering.}
The filtering posterior incorporates both the propagated previous belief and the current emission evidence. At the first step,
\begin{equation*}
q_0^{\mathrm{filter}}(k)
\propto
\pi_k p(X_0\mid S_0=k).
\end{equation*}
Thereafter,
\begin{equation*}
q_t^{\mathrm{filter}}
\propto
\left(
q_{t-1}^{\mathrm{filter}}A
\right)
\odot
p(X_t\mid S_t),
\end{equation*}
followed by normalization across the two states. This quantity is exactly
\begin{equation*}
p(S_t\mid X_{\leq t}).
\end{equation*}

No learned HMM and MCJEPA models are being compared in Experiment~2. Both curves are oracle calculations under the same known generating process. This design isolates the informational value of temporal history from representation-learning and optimization effects.

\paragraph{Evaluation.}
We use the accuracy, state NLL, Brier score, and posterior entropy defined in \cref{app:experimental-metrics}. Accuracy gives the most immediately interpretable state-recovery comparison, while NLL measures whether the posterior assigns high probability to the realized state. Brier score provides a complementary proper probabilistic score, and entropy records posterior confidence.

\paragraph{Representative sequence selection.}
The representative trajectory in \cref{fig:exp2-filtering} is selected only for visualization; all quantitative results use all generated sequences. We use the most ambiguous setting,
\begin{equation*}
\mu/\sigma=0.45,
\end{equation*}
from the first seed and search $45$-step windows centered on genuine latent-state transitions. Windows containing one to three true state switches receive a small preference, and among candidate windows we favor those in which filtering gives a larger realized-state NLL improvement over local evidence. This produces a transition-rich example that visibly illustrates the mechanism quantified by the aggregate experiment rather than selecting the first sequence arbitrarily.

\begin{figure}[t]
\centering
\includegraphics[width=0.5\linewidth]{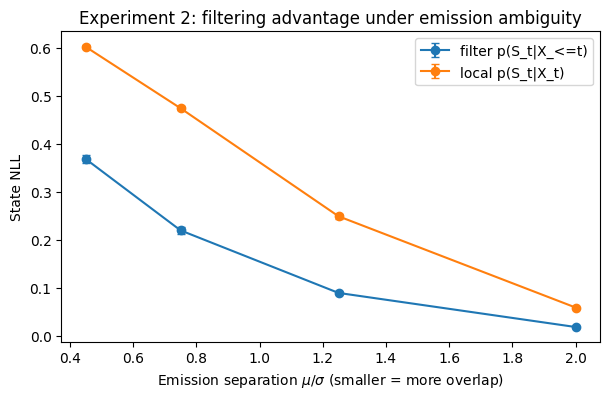}
\caption{State NLL for exact local evidence and exact filtering under increasing emission ambiguity. Smaller $\mu/\sigma$ corresponds to stronger overlap between the two Gaussian emissions. Error bars denote mean $\pm$ one standard deviation over five independently generated datasets.}
\label{fig:app-exp2-nll}
\end{figure}

\paragraph{Supplementary result.}
\Cref{fig:app-exp2-nll} gives the complete state-NLL comparison across emission separations. The filtering advantage grows as $\mu/\sigma$ decreases, matching the accuracy trend reported in the main text.

\subsection{Experiment 3: predictive compression and Markovization}
\label{app:exp3-details}

\paragraph{Second-order binary process.}
Let
\begin{equation*}
Y_t=X_{t+1}.
\end{equation*}
The data-generating process is
\begin{equation*}
p(X_{t+1}=1\mid X_{t-1},X_t)
=
\begin{cases}
0.10, & (X_{t-1},X_t)=(0,0),\\
0.90, & (X_{t-1},X_t)=(0,1),\\
0.80, & (X_{t-1},X_t)=(1,0),\\
0.20, & (X_{t-1},X_t)=(1,1).
\end{cases}
\end{equation*}
The corresponding first-order transition matrix on pair states
\begin{equation*}
(00,01,10,11)
\end{equation*}
is
\begin{equation*}
A_{\mathrm{pair}}
=
\begin{bmatrix}
0.9 & 0.1 & 0   & 0\\
0   & 0   & 0.1 & 0.9\\
0.2 & 0.8 & 0   & 0\\
0   & 0   & 0.8 & 0.2
\end{bmatrix}.
\end{equation*}
Its stationary distribution is
\begin{equation*}
\pi_{\mathrm{pair}}
=
\left(
\frac{16}{41},
\frac{8}{41},
\frac{8}{41},
\frac{9}{41}
\right)
\approx
(0.3902,0.1951,0.1951,0.2195).
\end{equation*}
The exact joint distribution of
\begin{equation*}
H_t=(X_{t-2},X_{t-1},X_t)
\end{equation*}
and $Y_t=X_{t+1}$ is constructed analytically from this stationary pair chain. Consequently, the exact representation controls and deterministic frontier do not require Monte Carlo estimation.

\paragraph{Exact representation controls.}
A deterministic representation is a mapping
\begin{equation*}
Z=f(H).
\end{equation*}
For such a mapping,
\begin{equation*}
I(H;Z)=H(Z),
\end{equation*}
because $H(Z\mid H)=0$. For each representation we construct the exact joint distribution $p(z,y)$ and compute
\begin{equation*}
I(Z;Y)
=
\sum_{z,y}
p(z,y)
\log
\frac{p(z,y)}{p(z)p(y)}.
\end{equation*}
All logarithms are natural, so information quantities and NLLs are measured in nats.

The optimal one-step probabilistic predictor for a fixed representation is the exact conditional distribution $p(Y\mid Z)$. Its prediction NLL is therefore
\begin{equation*}
\mathcal L_{\mathrm{pred}}
=
-\sum_{z,y}
p(z,y)\log p(y\mid z)
=
H(Y\mid Z).
\end{equation*}

The three control mappings are
\begin{align*}
Z^{\mathrm{over}}
&=
(X_{t-2},X_{t-1},X_t),
\\
Z^{\mathrm{suff}}
&=
(X_{t-1},X_t),
\\
Z^{\mathrm{under}}
&=
X_t.
\end{align*}
The first has eight possible values, the second four, and the third two. Their exact information and prediction quantities are reproduced in \cref{tab:app-exp3-controls} for completeness.

\begin{table}[t]
\centering
\small
\renewcommand{\arraystretch}{1.08}
\caption{Exact structural controls for Experiment~3. The four-state predictive pair removes redundant history while preserving all one-step predictive information.}
\label{tab:app-exp3-controls}
\begin{tabular}{lcccc}
\toprule
Representation
&
$I(H;Z)$
&
$I(Z;Y)$
&
$H(Y\mid Z)$
&
States
\\
\midrule
Three-bit history
&
1.7356
&
0.2807
&
0.3978
&
8
\\
Predictive pair
&
1.3378
&
0.2807
&
0.3978
&
4
\\
Current bit only
&
0.6785
&
0.0192
&
0.6593
&
2
\\
\bottomrule
\end{tabular}
\end{table}

\paragraph{Enumeration of deterministic partitions.}
Because $H_t$ has only eight possible values, every deterministic compression can be enumerated. A deterministic encoder identifies histories that share the same output label and therefore corresponds to a set partition of the eight histories. The number of such partitions is the eighth Bell number,
\begin{equation*}
B_8=4140.
\end{equation*}
The Bell number $B_n$ counts the number of partitions of an $n$-element set into nonempty unlabeled subsets. Here, each of the eight possible three-bit histories is an element and each subset collects histories assigned to the same latent state. The implementation uses restricted-growth strings to enumerate each partition exactly once, thereby eliminating duplicates caused solely by relabeling latent states.

For every partition we compute
\begin{equation*}
I(H;Z),
\qquad
I(Z;Y),
\qquad
H(Y\mid Z)
\end{equation*}
exactly. A representation belongs to the deterministic Pareto frontier if there is no other deterministic representation with no larger $I(H;Z)$ and strictly larger $I(Z;Y)$. A numerical tolerance of $10^{-12}$ is used when constructing this frontier. Of the $4140$ deterministic partitions, $36$ are nondominated.

\paragraph{Exact compression sweep.}
For each compression coefficient $\beta$, every deterministic partition is scored using
\begin{equation*}
\mathcal L_{\mathrm{PIB}}^{\mathrm{exp}}
=
H(Y\mid Z)
+
\beta I(H;Z).
\end{equation*}
The tested values are
\begin{equation*}
\beta
\in
\{
0,\,
0.0005,\,
0.001,\,
0.003,\,
0.005,\,
0.01,\,
0.012,\,
0.014,\,
0.015,\,
0.02,\,
0.03,\,
0.10,\,
0.30,\,
1.0
\}.
\end{equation*}

At score ties, using tolerance $10^{-10}$, we first select the candidate with lower $I(H;Z)$, then the candidate with fewer occupied states, and finally larger $I(Z;Y)$. This convention matters at $\beta=0$: several deterministic representations achieve the same minimum prediction loss, so the four-state minimal sufficient representation is the \emph{reported tie-broken representative}, not a uniquely preferred solution of prediction alone.

The exact optimum evolves as shown in \cref{tab:app-exp3-beta-optima}. For every tested positive value through $\beta=0.014$, the optimum is the known four-state predictive pair. At $\beta=0.015$, the optimum switches to a two-state compression and sacrifices a small amount of predictive information. At $\beta=1$, complete compression to a single state becomes optimal.

\begin{table}[t]
\centering
\small
\renewcommand{\arraystretch}{1.08}
\caption{Exact deterministic compression regimes in Experiment~3. At $\beta=0$, the four-state entry is the lower-information representative selected among predictively tied optima.}
\label{tab:app-exp3-beta-optima}
\begin{tabular}{lcccc}
\toprule
Compression range
&
$I(H;Z)$
&
$I(Z;Y)$
&
Pred.\ NLL
&
States
\\
\midrule
$\beta=0$ (tie-broken)
&
1.3378
&
0.2807
&
0.3978
&
4
\\
$0<\beta\leq0.014$
&
1.3378
&
0.2807
&
0.3978
&
4
\\
$0.015\leq\beta\leq0.30$
&
0.6689
&
0.2711
&
0.4074
&
2
\\
$\beta=1$
&
0
&
0
&
0.6785
&
1
\\
\bottomrule
\end{tabular}
\end{table}

\paragraph{Learned stochastic continuation.}
The orange continuation curve in \cref{fig:exp3-markovization} is generated separately from the exhaustive deterministic search. Its purpose is to show how gradient optimization behaves as compression pressure is gradually increased.

Because the underlying problem is finite, the learned encoder is represented directly as a categorical table
\begin{equation*}
q_\theta(z\mid h),
\qquad
h\in\{0,\ldots,7\},
\qquad
z\in\{0,\ldots,7\},
\end{equation*}
rather than by a neural sequence encoder. This removes architectural capacity as a confound. The predictor is a second categorical table,
\begin{equation*}
p_\omega(Y\mid Z).
\end{equation*}

The encoder is initialized near the overcomplete identity mapping $Z=H$. Specifically, the diagonal encoder logits are initialized to $+8$, the off-diagonal logits to $-8$, and independent noise of scale $10^{-3}$ is added to break exact symmetry. The predictor is initialized from the exact conditional distribution $p(Y\mid H)$ associated with the identity representation.

For each tested $\beta$, the model minimizes
\begin{equation*}
\mathcal L_{\mathrm{learned}}
=
\mathcal L_{\mathrm{pred}}
+
\beta I(H;Z)
\end{equation*}
using Adam with learning rate
\begin{equation*}
3\times10^{-2}.
\end{equation*}
Each nonzero compression stage receives $2200$ gradient steps. The solution at one value of $\beta$ initializes the next, so the orange curve is a \emph{continuation path} rather than a collection of independently initialized models. At $\beta=0$, the deliberately overcomplete eight-state initialization is retained without an optimization stage. Five independent perturbation seeds are used, and the plotted curve reports their mean with standard deviations.

This distinguishes the two $\beta=0$ constructions in the experiment. The exact deterministic sweep reports the lower-information four-state solution after tie-breaking among equally predictive partitions, whereas the learned continuation deliberately begins from the overcomplete eight-state solution so that the compression trajectory can be observed.

The learned encoder is stochastic, whereas the exact blue frontier contains only deterministic mappings. The learned continuation is therefore interpreted \emph{relative to} the deterministic global reference, not as an optimization method expected to lie exactly on that frontier.

\paragraph{Residual-predictability diagnostic.}
The residual diagnostic uses five independently generated sequences of length $120{,}000$, initialized from the stationary pair-state distribution. Each sequence is divided chronologically into $70\%$ training and $30\%$ held-out evaluation data.

The restricted predictor estimates
\begin{equation*}
\widehat p_0
=
\widehat p(X_{t+1}=1\mid Z_t),
\end{equation*}
whereas the history-augmented predictor receives one additional step of representation history,
\begin{equation*}
\widehat p_1
=
\widehat p(X_{t+1}=1\mid Z_t,Z_{t-1}).
\end{equation*}
Both are discrete lookup estimators fitted on the training portion with Laplace smoothing parameter
\begin{equation*}
\alpha=1.
\end{equation*}

For the three controlled representations, the augmented inputs specialize to
\begin{align*}
Z_t^{\mathrm{under}}=X_t
&:
&
(Z_t,Z_{t-1})
&=
(X_t,X_{t-1}),
\\
Z_t^{\mathrm{suff}}=(X_{t-1},X_t)
&:
&
(Z_t,Z_{t-1})
&=
\bigl((X_{t-1},X_t),(X_{t-2},X_{t-1})\bigr),
\\
Z_t^{\mathrm{over}}=(X_{t-2},X_{t-1},X_t)
&:
&
(Z_t,Z_{t-1})
&=
\bigl(
(X_{t-2},X_{t-1},X_t),
(X_{t-3},X_{t-2},X_{t-1})
\bigr).
\end{align*}

This makes the positive and negative controls transparent. For $Z_t=X_t$, the augmentation $Z_{t-1}=X_{t-1}$ restores exactly the transition-relevant variable omitted from the current state. For $Z_t=(X_{t-1},X_t)$, the only genuinely new observation supplied by $Z_{t-1}$ is $X_{t-2}$, which is redundant under the data-generating process. The overcomplete representation already contains still more history, so adding its previous state should likewise provide no one-step predictive benefit.

On the held-out portion we compute
\begin{align*}
\mathrm{MSE}_{\mathrm{restricted}}
&=
\E
\left[
(X_{t+1}-\widehat p_0)^2
\right],
\\
\mathrm{MSE}_{\mathrm{history\text{-}augmented}}
&=
\E
\left[
(X_{t+1}-\widehat p_1)^2
\right],
\end{align*}
and define
\begin{equation*}
\Delta_{\mathrm{hist}}
=
\mathrm{MSE}_{\mathrm{restricted}}
-
\mathrm{MSE}_{\mathrm{history\text{-}augmented}}.
\end{equation*}
Thus, positive $\Delta_{\mathrm{hist}}$ means that $Z_{t-1}$ contains predictive information absent from $Z_t$. Values near zero indicate that one further step of representation history does not improve held-out prediction.

The measured gains are
\begin{equation*}
0.1139\pm0.0026
\end{equation*}
for the insufficient representation,
\begin{equation*}
-6.8\times10^{-6}
\end{equation*}
for the minimal sufficient representation, and
\begin{equation*}
-3.7\times10^{-5}
\end{equation*}
for the overcomplete representation. The tiny negative values are finite-sample fitting variation and are effectively zero at the scale of the experiment.

Operationally, this diagnostic measures the held-out gain from adding $Z_{t-1}$ rather than fitting a separate neural regressor to residuals. It realizes the same sufficiency principle used in the main text: if omitted history still carries transition-relevant information, augmenting the predictor with an earlier representation state should reduce held-out prediction error.

\begin{figure}[t]
\centering
\includegraphics[width=0.6\linewidth]{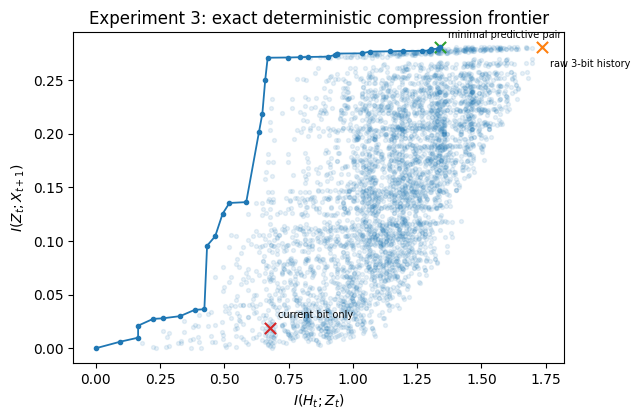}
\caption{Exact deterministic compression frontier for Experiment~3. All $4140$ deterministic partitions of the eight histories are evaluated exactly. The nondominated frontier gives the best achievable deterministic trade-offs between retaining history information and preserving information about $X_{t+1}$.}
\label{fig:app-exp3-frontier}
\end{figure}

\paragraph{Supplementary results.}
\Cref{fig:app-exp3-frontier} shows the exact deterministic frontier without the learned continuation overlay.

\begin{figure}[t]
\centering
\includegraphics[width=0.70\linewidth]{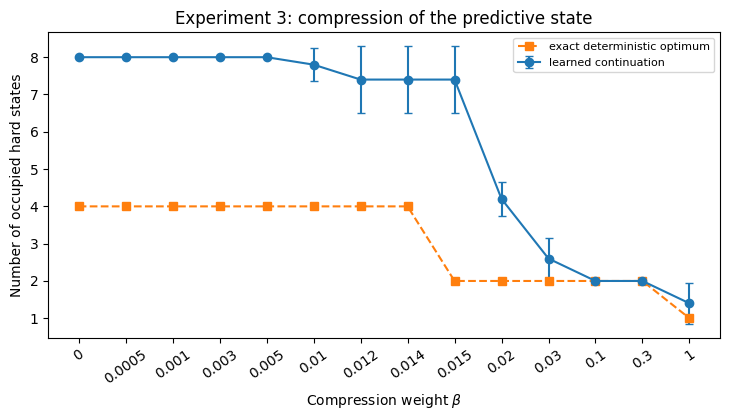}
\caption{Number of occupied hard states as the compression weight $\beta$ increases. The dashed reference gives the exact deterministic optimum, while the learned stochastic continuation follows the warm-started gradient trajectory.}
\label{fig:app-exp3-state-count}
\end{figure}

\Cref{fig:app-exp3-state-count} shows the number of occupied hard states along the compression sweep for both the exact deterministic optimum and the learned continuation. The learned warm-started path can differ from the global deterministic optimum because of stochastic parameterization and optimization path dependence.

\subsection{Experiment 4: HMM-style training of PIB-VJEPA}
\label{app:exp4-details}

\paragraph{Independent rerun with a shared latent-state family.}
Experiment~4 is implemented and run independently from Experiments~1--3. It regenerates the same separated-emission data used in Experiment~1 using the same ground-truth transition matrix, Gaussian state means, emission standard deviation
\begin{equation*}
\sigma=0.35,
\end{equation*}
and data seeds. For top-level seed $s\in\{0,\ldots,4\}$, the training generator uses
\begin{equation*}
1000+17s,
\end{equation*}
and the test generator uses
\begin{equation*}
2000+17s.
\end{equation*}
Consequently, Experiment~4 sees the same paired data realizations as the separated condition of Experiment~1, but none of the fitted Experiment~1 models is reused. All three Experiment~4 regimes are trained afresh.

Each seed contains $400$ training sequences and $160$ test sequences of length $80$. All regimes use $K=4$ categorical latent states and the same basic one-layer GRU context-encoder family with hidden dimension $48$.

\paragraph{Common initialization and state-use regularization.}
All categorical encoder regimes use an unsupervised K-means warm start with $K=4$ and $10$ K-means initializations; ground-truth states are never used. The context encoder is pretrained for $100$ updates at learning rate
\begin{equation*}
10^{-2}
\end{equation*}
to reproduce the K-means assignments.

The Gaussian HMM components in the HMM+filter and hybrid regimes are initialized from the same K-means partition. State-conditional means are initialized from cluster centers, diagonal variances from within-cluster variances plus $0.05$, the initial-state logits are initialized uniformly, and the transition logits receive a diagonal bias
\begin{equation*}
L_A=1.5I_4.
\end{equation*}

The same state-use regularization is applied to all three amortized context encoders:
\begin{equation*}
\mathcal L_{\mathrm{state}}
=
0.30\,\mathcal L_{\mathrm{occ}}
+
0.02\,\mathcal L_{\mathrm{ent}}.
\end{equation*}
Sharing these coefficients removes state-regularization strength as a confound in the objective-level comparison.

\paragraph{JEPA latent regime.}
The JEPA-only regime is trained from scratch using the same MCJEPA construction and optimization settings as the shared-$A$ model in Experiment~1. Its objective is
\begin{equation*}
\mathcal L_{\mathrm{JEPA}}
=
\mathcal L_{\mathrm{MC}}
+
\mathcal L_{\mathrm{state}},
\end{equation*}
where
\begin{equation*}
\mathcal L_{\mathrm{MC}}
=
\frac{1}{|\mathcal H|}
\sum_{h\in\mathcal H}
\E_t
\left[
\KL
\left(
\sg\bigl(
q_{\bar\theta}(Z_{t+h}\mid X_{t+h})
\bigr)
\,\middle\|\,
q_\theta(Z_t\mid X_{\leq t})A^h
\right)
\right].
\end{equation*}

The context encoder is trained for $180$ epochs with mini-batches of $64$ sequences, Adam learning rate
\begin{equation*}
3\times10^{-3},
\end{equation*}
EMA coefficient
\begin{equation*}
\tau=0.995,
\end{equation*}
and gradient clipping at norm $5$.

No observation model participates in this training. To evaluate observation-sequence NLL afterward, we fit a diagonal Gaussian emission distribution to each learned latent state using the soft training assignments:
\begin{equation*}
\widehat\mu_k
=
\frac{
\sum_{n,t}q_{nt}(k)X_{nt}
}{
\sum_{n,t}q_{nt}(k)
}.
\end{equation*}
The diagonal variance is the corresponding soft-assignment-weighted second moment around $\widehat\mu_k$, with a minimum variance of $0.03$. The initial-state distribution is estimated from the mean encoder distribution at the first time step. These post-hoc parameters do not backpropagate into either the JEPA encoder or transition matrix.

The fitted observation model and learned transition are then treated as a fixed HMM solely for evaluation of test sequence NLL and the post-hoc filtering diagnostic.

\paragraph{HMM sequence + filter-distillation regime.}
The second regime jointly maintains a Gaussian HMM and an amortized GRU context encoder but contains no JEPA latent-prediction loss. The HMM contributes
\begin{equation*}
\mathcal L_{\mathrm{HMM}}
=
-\frac{1}{T}
\E
\left[
\log p_{\phi,\psi}(X_{1:T})
\right].
\end{equation*}
At each update, its current exact filtering posterior is computed and detached,
\begin{equation*}
\widetilde q_t
=
\sg\left(
p_{\phi,\psi}(Z_t\mid X_{\leq t})
\right),
\end{equation*}
and the context encoder is trained through
\begin{equation*}
\mathcal L_{\mathrm{filter}}
=
\E_t
\left[
\KL
\left(
\widetilde q_t
\,\middle\|\,
q_\theta(Z_t\mid X_{\leq t})
\right)
\right].
\end{equation*}
The implemented objective is
\begin{equation*}
\mathcal L_{\mathrm{HMM+filter}}
=
1.0\,\mathcal L_{\mathrm{HMM}}
+
0.5\,\mathcal L_{\mathrm{filter}}
+
0.30\,\mathcal L_{\mathrm{occ}}
+
0.02\,\mathcal L_{\mathrm{ent}}.
\end{equation*}

The HMM and encoder are jointly optimized for $600$ full-data updates using Adam with learning rate
\begin{equation*}
10^{-2}.
\end{equation*}
Gradients are clipped to norm $10$. Because the filtering target is detached, $\mathcal L_{\mathrm{filter}}$ updates the amortized encoder but not the HMM parameters. Likewise, $\mathcal L_{\mathrm{state}}$ acts only on the encoder. Thus, the generative transition and emission parameters are learned through sequence likelihood, while the context encoder learns to amortize the corresponding Bayesian filtering operation.

\paragraph{Hybrid HMM + latent regime.}
The hybrid uses the same Gaussian HMM and amortized context-encoder families as the preceding regime but adds the genuine MCJEPA latent-prediction objective. A separate target encoder is initialized from the context encoder after K-means pretraining and subsequently updated only by EMA.

Its three principal losses are
\begin{align*}
\mathcal L_{\mathrm{HMM}}
&=
-\frac{1}{T}
\E
\left[
\log p_{\phi,\psi}(X_{1:T})
\right],
\\
\mathcal L_{\mathrm{filter}}
&=
\E_t
\left[
\KL
\left(
\sg(\widetilde q_t)
\,\middle\|\,
q_\theta(Z_t\mid X_{\leq t})
\right)
\right],
\\
\mathcal L_{\mathrm{MC}}
&=
\frac{1}{|\mathcal H|}
\sum_{h\in\mathcal H}
\E_t
\left[
\KL
\left(
\sg\bigl(
q_{\bar\theta}(Z_{t+h}\mid X_{t+h})
\bigr)
\,\middle\|\,
q_\theta(Z_t\mid X_{\leq t})A^h
\right)
\right].
\end{align*}

The critical implementation distinction is that the HMM filtering posterior
\begin{equation*}
\widetilde q_t
=
p_{\phi,\psi}(Z_t\mid X_{\leq t})
\end{equation*}
is used \emph{only} by $\mathcal L_{\mathrm{filter}}$. It is not substituted for the future JEPA target in $\mathcal L_{\mathrm{MC}}$. Instead, the latter uses the same EMA local-target construction as the JEPA-only baseline. The hybrid comparison is therefore objective-faithful: its $\mathcal L_{\mathrm{MC}}$ remains the original JEPA latent-prediction signal.

The transition matrix $A$ is shared between the HMM and JEPA objectives. Hence the same latent dynamics are trained simultaneously by observation-sequence evidence and latent predictive alignment. The complete implemented objective is
\begin{equation*}
\begin{aligned}
\mathcal L_{\mathrm{hybrid}}
={}&
1.0\,\mathcal L_{\mathrm{HMM}}
+
1.0\,\mathcal L_{\mathrm{MC}}
+
0.5\,\mathcal L_{\mathrm{filter}}
\\
&+
0.30\,\mathcal L_{\mathrm{occ}}
+
0.02\,\mathcal L_{\mathrm{ent}}.
\end{aligned}
\end{equation*}
The model is trained for $600$ full-data updates with Adam learning rate
\begin{equation*}
10^{-2},
\end{equation*}
EMA coefficient
\begin{equation*}
\tau=0.995,
\end{equation*}
and gradient clipping at norm $10$.

\paragraph{Which parameters are trained by each objective?}
For clarity, \cref{tab:app-exp4-gradient-flow} summarizes the effective parameter flow. Both the exact filtering target and EMA JEPA target are stop-gradient quantities.

\begin{table}[t]
\centering
\small
\renewcommand{\arraystretch}{1.10}
\caption{Effective parameter updates in the revised Experiment~4 implementation. The target encoder receives no gradient and is updated only by EMA.}
\label{tab:app-exp4-gradient-flow}
\begin{tabular}{lccc}
\toprule
Loss
&
Context encoder
&
Transition $A$
&
Emission / initial-state parameters
\\
\midrule
$\mathcal L_{\mathrm{HMM}}$
&
--
&
$\checkmark$
&
$\checkmark$
\\
$\mathcal L_{\mathrm{MC}}$
&
$\checkmark$
&
$\checkmark$
&
--
\\
$\mathcal L_{\mathrm{filter}}$
&
$\checkmark$
&
--
&
--
\\
$\mathcal L_{\mathrm{state}}$
&
$\checkmark$
&
--
&
--
\\
\bottomrule
\end{tabular}
\end{table}

This separation clarifies the interpretation of the comparison. In HMM+filter training, the generative parameters are learned from sequence evidence and the context encoder amortizes the resulting filter. In hybrid training, the transition additionally receives the JEPA latent-prediction signal, while the emission and initial-state parameters remain trained through sequence evidence.

\paragraph{Forward algorithm and numerical stabilization.}
All HMM sequence likelihoods are evaluated exactly in log space. Let
\begin{equation*}
\ell_t(k)
=
\log p_\psi(X_t\mid Z_t=k).
\end{equation*}
The forward recursion is initialized as
\begin{equation*}
\alpha_1(k)
=
\log\pi_k+\ell_1(k)
\end{equation*}
and updated by
\begin{equation*}
\alpha_t(j)
=
\ell_t(j)
+
\operatorname{logsumexp}_i
\left[
\alpha_{t-1}(i)+\log A_{ij}
\right].
\end{equation*}
The sequence log-likelihood is
\begin{equation*}
\log p(X_{1:T})
=
\operatorname{logsumexp}_k\alpha_T(k).
\end{equation*}

Learned HMM log variances are clamped to
\begin{equation*}
[-7,4]
\end{equation*}
before evaluating Gaussian emissions. A numerical floor
\begin{equation*}
\varepsilon=10^{-8}
\end{equation*}
is used when taking logarithms or normalizing probabilities.

\paragraph{Evaluation protocol.}
The common metrics and Hungarian alignment follow \cref{app:experimental-metrics}. For every regime, the alignment is obtained from training-set hard assignments and held fixed during test evaluation. We report test ARI and NMI, transition error $\mathcal E_A$, observation-sequence NLL per time step, filtering KL, and true-state prediction NLL at
\begin{equation*}
h\in\{1,2,4,8\}.
\end{equation*}

The three groups of metrics have distinct interpretations. ARI and NMI evaluate representation recovery; transition error and true-state NLL evaluate learned latent dynamics; sequence NLL evaluates the complete transition--emission model. Filtering KL is treated separately because it is an explicitly optimized quantity for two of the three regimes.

\paragraph{Paired sequence-evidence comparison.}
All three regimes within a seed are evaluated on exactly the same test realization. The sequence-evidence figure therefore uses a paired difference. For regime $r$ and seed $s$, we compute
\begin{equation*}
\Delta\mathrm{NLL}_{r,s}
=
\mathrm{NLL}_{r,s}
-
\mathrm{NLL}_{\mathrm{HMM+filter},s}
\end{equation*}
before averaging across seeds. This removes variability caused by different sampled test sequences and makes the objective-induced difference easier to see.

\begin{table}[t]
\centering
\small
\renewcommand{\arraystretch}{1.08}
\caption{Paired observation-sequence NLL difference relative to HMM+filter training. Differences are computed within each seed before aggregation. Lower is better.}
\label{tab:app-exp4-delta-nll}
\begin{tabular}{lc}
\toprule
Training regime
&
$\Delta$ sequence NLL / step
\\
\midrule
HMM sequence + filter distill
&
$0$
\\
Hybrid HMM + latent
&
$0.000134\pm0.000092$
\\
JEPA latent objective
&
$0.009409\pm0.003164$
\\
\bottomrule
\end{tabular}
\end{table}

Thus, the hybrid retains only a very small sequence-evidence gap relative to HMM-style training, whereas the latent-only JEPA model remains clearly separated. We use this paired comparison descriptively rather than as a formal hypothesis test.

\paragraph{Multi-horizon prediction.}
For each regime we propagate the current context distribution through powers of its learned transition matrix,
\begin{equation*}
\widehat q_{t+h}
=
q_tA^h,
\end{equation*}
align the result to ground-truth state order, and evaluate true-state NLL. The complete results are shown in \cref{tab:app-exp4-multihorizon}.

\begin{table}[t]
\centering
\small
\renewcommand{\arraystretch}{1.08}
\caption{True-state multi-horizon prediction NLL in Experiment~4. Values are mean $\pm$ standard deviation over five seeds. Lower is better.}
\label{tab:app-exp4-multihorizon}
\begin{tabular}{lcccc}
\toprule
Training regime
&
$h=1$
&
$h=2$
&
$h=4$
&
$h=8$
\\
\midrule
HMM sequence + filter distill
&
$\mathbf{0.5271\pm0.0027}$
&
$\mathbf{0.7983\pm0.0045}$
&
$\mathbf{1.0939\pm0.0054}$
&
$1.3035\pm0.0047$
\\
Hybrid HMM + latent
&
$0.5295\pm0.0030$
&
$0.7996\pm0.0048$
&
$1.0943\pm0.0057$
&
$\mathbf{1.3029\pm0.0051}$
\\
JEPA latent objective
&
$0.5437\pm0.0019$
&
$0.8106\pm0.0040$
&
$1.1004\pm0.0056$
&
$1.3049\pm0.0049$
\\
\bottomrule
\end{tabular}
\end{table}

Both HMM-style and hybrid training improve over JEPA-only at every evaluated horizon. The differences are largest at shorter horizons and narrow by $h=8$, where repeated application of the transition matrix increasingly mixes the predictive state distribution.

\paragraph{Filtering agreement.}
Filtering KL is defined in \cref{app:experimental-metrics}. The exact reference is the filtering distribution implied by the probabilistic model associated with each regime. For HMM+filter and hybrid training, this is the exact filter of their jointly trained Gaussian HMM. For JEPA-only, it is the filter obtained after fitting the post-hoc Gaussian observation model to the learned JEPA states.

The resulting values are
\begin{equation*}
0.000060\pm0.000007
\end{equation*}
for HMM+filter,
\begin{equation*}
0.004355\pm0.001015
\end{equation*}
for the hybrid, and
\begin{equation*}
0.025460\pm0.006423
\end{equation*}
for JEPA-only.

These quantities do not compare every model with one common external filtering oracle. Moreover, filtering KL is explicitly optimized for HMM+filter and hybrid training, so for those regimes it is a \emph{training-aligned role diagnostic}. For JEPA-only it is instead a post-hoc diagnostic of how closely the learned context representation happens to agree with the filtering distribution induced by its fitted observation model.

\paragraph{Seed-wise consistency.}
The aggregate improvement of hybrid training over JEPA-only is not produced by one favorable seed. For each of the five paired runs, hybrid training improves over JEPA-only in ARI, NMI, observation-sequence NLL, transition error, filtering KL, and true-state prediction NLL at every evaluated horizon $h\in\{1,2,4,8\}$. We report this pattern descriptively and do not infer formal statistical significance from five seeds.

\begin{figure}[t]
\centering
\includegraphics[width=0.70\linewidth]{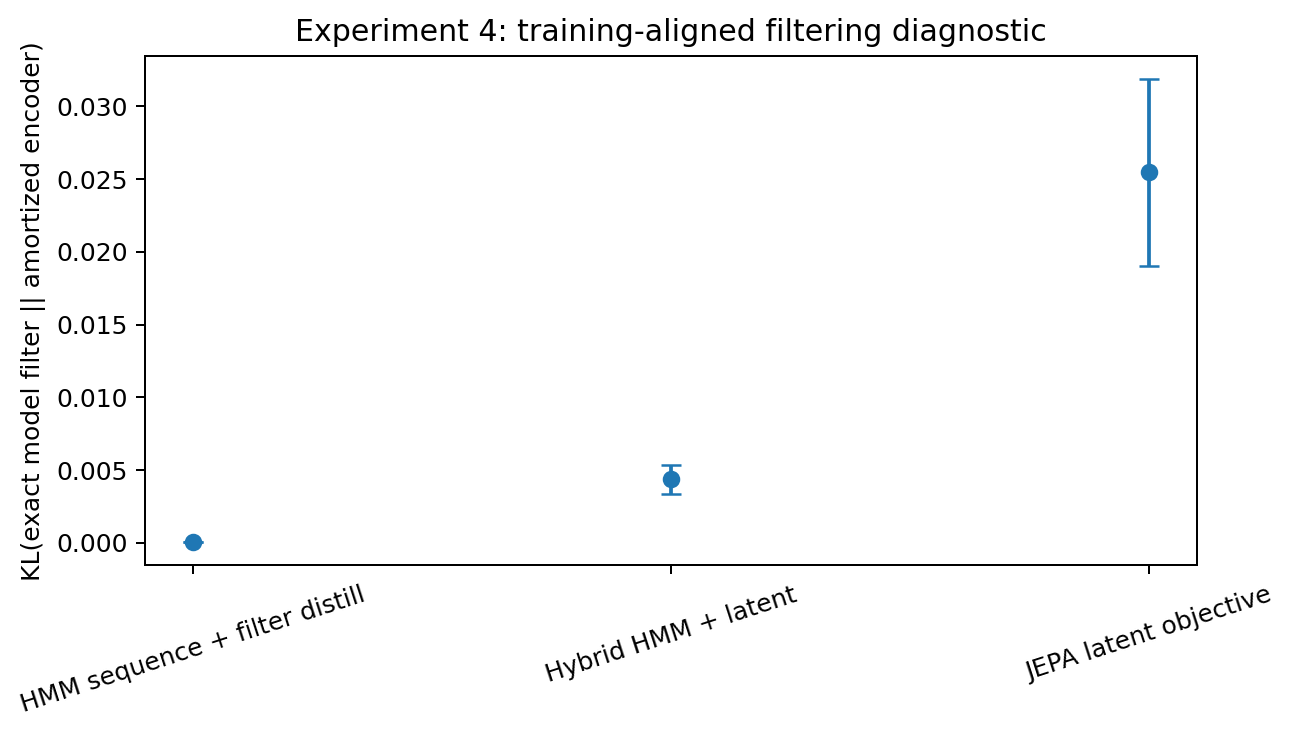}
\caption{Filtering agreement across the three Experiment~4 training regimes. The quantity is $\KL(q_t^{\mathrm{exact}}\|q_t^{\mathrm{enc}})$ averaged over test time points. HMM+filter and hybrid training explicitly optimize filtering alignment, so their values should be interpreted as training-aligned role diagnostics. JEPA-only is evaluated post hoc using the filtering distribution induced by its fitted observation model.}
\label{fig:app-exp4-filtering}
\end{figure}

\paragraph{Supplementary filtering diagnostic.}
\Cref{fig:app-exp4-filtering} reports filtering agreement separately from the two generative-model metrics emphasized in the main text.

\subsection{Reproducibility and computation}
\label{app:exp-reproducibility}

The Python code which implements these 4 experiments can be found at this Github repo: \text{https://github.com/YongchaoHuang/HMM-JEPA}.

Five fixed seeds
\begin{equation*}
\{0,1,2,3,4\}.
\end{equation*}
were used. 
The implementation uses single-precision PyTorch tensors,
\begin{equation*}
\texttt{torch.float32},
\end{equation*}
and automatically selects CUDA when available, otherwise falling back to CPU.

\paragraph{Experiments 1--3.}
The principal paper-mode settings are
\begin{equation*}
N_{\mathrm{train}}=400,
\qquad
N_{\mathrm{test}}=160,
\qquad
T=80,
\qquad
d_{\mathrm{hidden}}=48,
\qquad
B=64.
\end{equation*}
The main optimization budgets are
\begin{equation*}
180\ \text{MCJEPA epochs},
\qquad
100\ \text{MCJEPA warm-start steps},
\qquad
350\ \text{Gaussian-HMM steps},
\end{equation*}
and
\begin{equation*}
2200\ \text{updates per nonzero Experiment~3 compression stage}.
\end{equation*}
Experiment~3 uses five independent perturbation seeds for the learned information-bottleneck continuation and five independently generated long sequences for the residual diagnostic.

\paragraph{Experiment 4.}
The standalone Experiment~4 script uses
\begin{equation*}
N_{\mathrm{train}}=400,
\qquad
N_{\mathrm{test}}=160,
\qquad
T=80,
\qquad
d_{\mathrm{hidden}}=48.
\end{equation*}

The JEPA-only baseline uses $180$ epochs with batch size $64$, learning rate $3\times10^{-3}$, $100$ K-means warm-start updates, and EMA coefficient $0.995$. The HMM+filter and hybrid regimes each use $600$ full-data joint updates with Adam learning rate
\begin{equation*}
10^{-2}.
\end{equation*}
Their common objective coefficients are
\begin{equation*}
\lambda_{\mathrm{seq}}=1.0,
\qquad
\lambda_{\mathrm{latent}}=1.0,
\qquad
\lambda_{\mathrm{filter}}=0.5,
\end{equation*}
where $\lambda_{\mathrm{latent}}$ applies only to the hybrid, together with
\begin{equation*}
\lambda_{\mathrm{occ}}=0.30,
\qquad
\lambda_{\mathrm{ent}}=0.02.
\end{equation*}
The hybrid EMA coefficient is
\begin{equation*}
\tau=0.995.
\end{equation*}

The JEPA-only model is initialized with top-level seed $s$, while the independently trained HMM+filter and hybrid regimes use deterministic seed offsets associated with $s$ so that each run remains reproducible while avoiding accidental reuse of identical parameter initialization streams.

Because the JEPA-only objective is naturally optimized with sequence mini-batches whereas the differentiable HMM sequence objective is evaluated on the full training collection in the HMM+filter and hybrid implementations, Experiment~4 should be interpreted as an \emph{objective-behavior diagnostic}, not as a compute-matched optimization-efficiency benchmark. The latent-state family, data realization, context-encoder family, state-use regularization, and evaluation protocol are controlled across regimes.

\paragraph{Randomness and numerical reproducibility.}
Randomness is seeded for Python's \texttt{random} module, NumPy, PyTorch, and all available CUDA devices. The data generators use fixed seed offsets so that training data, test data, collapse diagnostics, Experiment~2 filtering datasets, Experiment~3 continuation runs, residual-diagnostic sequences, and Experiment~4 model initializations can be reproduced independently from the top-level seed.

We do not enforce PyTorch deterministic-algorithm mode, so exact bitwise reproducibility across different CUDA libraries or hardware is not guaranteed. The scripts do not record the specific accelerator model or wall-clock runtime, and package versions are not pinned in the experimental source; we therefore do not report hardware-specific timing claims. The implementation depends on NumPy, Pandas, PyTorch, scikit-learn, SciPy, and Matplotlib.

\section*{Disclaimer}

This work was developed with assistance from ChatGPT \citep{openai2026gpt} in idea development, technical formulation, writing, experimental design, and coding. The central idea, i.e. the correspondence between probabilistic temporal JEPA and hidden Markov models, was originally and independently proposed by the author, while ChatGPT contributed to its subsequent development. The presentation of this work, e.g. appearance of experimental results, is therefore different from previous work. The author estimates the overall contributions split as approximately 60\%:40\% between the author and ChatGPT. At the time of writing, the author does not expect an AI system to independently discover this research direction and refine it without substantial and careful human input, guidance, examination, correction and refinement. The work therefore reflects a hybrid mode of human--AI research collaboration, in which the human researcher provides the originating insight, direction, judgement, and verification, while the AI assists with elaboration and execution. All mathematical statements, technical claims, experimental procedures, results, and contents in main texts were manually reviewed and verified by the author, who takes full responsibility for the final work. Nevertheless, errors or inaccuracies may remain, and readers are encouraged to interpret the claims and results with appropriate caution.

\end{document}